%% file: main.tex
\documentclass[]{sensenova}

\usepackage{hyperref}
\usepackage{cleveref}
\usepackage{verbatim}

\usepackage{wrapfig}
\usepackage{graphicx}
\usepackage{floatrow}
\usepackage{placeins}
\usepackage{subcaption}
\usepackage{listings}
\usepackage{algorithm}
\usepackage{subcaption}
\usepackage{dsfont}

\usepackage{mmstyles}

\usepackage[toc,page,header]{appendix}

\usepackage{xspace}
\usepackage[utf8]{inputenc}  
\usepackage[T1]{fontenc}     
\usepackage{CJKutf8} 

\usepackage[misc]{ifsym}

\newcommand{\numbench}{eight\xspace}

\usepackage{CJKutf8} 
\usepackage[colorinlistoftodos,prependcaption,textsize=small]{todonotes}

\usepackage{fontawesome5}

\usepackage{soul} 
\definecolor{HLGreen}{rgb}{0.78,0.95,0.78}
\definecolor{HLRed}{rgb}{0.9725,0.8431,0.8549}   
\definecolor{benchblue}{RGB}{58, 122, 252}
\newcommand{\hlg}[1]{{\sethlcolor{HLGreen}\hl{#1}}}
\newcommand{\hlr}[1]{{\sethlcolor{HLRed}\hl{#1}}}
\newcommand{\colornum}[1]{%
  \footnotesize
  \ifdim #1 pt < 0pt
    \textcolor{sensepurple}{#1}%
  \else
    \textcolor{benchblue}{#1}%
  \fi
}

\usepackage{cuted} 

\definecolor{bestcolor}{RGB}{219, 208, 237}
\definecolor{secondcolor}{RGB}{241, 237, 248}
\definecolor{thirdcolor}{RGB}{211, 222, 190}
\definecolor{line-blue}{RGB}{243, 248, 252}

\newcommand{\name}{EASI\xspace}
\newcommand{\up}{\textcolor{green}{$\uparrow$}}
\newcommand{\down}{\textcolor{red}{$\downarrow$}}
\newcommand{\single}[1]{\multicolumn{2}{c}{#1}}
\usepackage{adjustbox}
\usepackage{makecell}
\usepackage{booktabs}
\newcommand{\pair}[2]{#1 & #2}
\newcommand{\subheads}{\scriptsize Standardized & \scriptsize Official}

\newlength{\ModelW}
\newlength{\NonModelW}
\newcolumntype{M}{>{\centering\arraybackslash}p{\NonModelW}}
\usepackage{graphicx}
\usepackage{subcaption}
\graphicspath{{assets/figures/}}

\usepackage{array}      
\usepackage{caption}
\usepackage{subcaption} 

\usepackage{makecell}

\usepackage{booktabs,tabularx,array}
\newcolumntype{Y}{>{\centering\arraybackslash}X}
\newcolumntype{L}{>{\raggedright\arraybackslash}X}

\title{
Holistic Evaluation of Multimodal LLMs on Spatial Intelligence
}

\author{
Zhongang Cai$^{*,1}$, Yubo Wang$^{*,1}$, Qingping Sun$^{*,1}$, Ruisi Wang$^{*,1}$, Chenyang Gu$^{*,1}$, Wanqi Yin$^{*,1}$, \\
Zhiqian Lin$^{*,1}$, Zhitao Yang$^{*,1}$, Chen Wei$^{*,1}$, 
Oscar Qian$^{*,1,2}$, Hui En Pang$^{*,2}$, Xuanke Shi$^{1}$, \\ Kewang Deng$^{1}$, Xiaoyang Han$^{1}$, Zukai Chen$^{1}$, Jiaqi Li$^{1}$, Xiangyu Fan$^{1}$, Hanming Deng$^{1}$, \\
Lewei Lu$^{1}$, Bo Li$^{2}$, Ziwei Liu$^{\textrm{\Letter}, 2}$, Quan Wang$^{\textrm{\Letter}, 1}$, Dahua Lin$^{\textrm{\Letter},1}$, Lei Yang$^{*,\textrm{\Letter},1}$

\parbox{\textwidth}{\centering\small
    $*$ Core Contributors,
    $\textrm{\Letter}$ Corresponding Authors, \\
    $^1$SenseTime Research,
    $^2$Nanyang Technological University
}}

\input{sec/0_abstract} 

\begin{document}
\maketitle
   
\input{sec/1_intro}
\input{sec/2_bench}
\input{sec/3_results}

\input{sec/4_case_study}

\input{sec/5_conclusion}

\clearpage

\bibliographystyle{plainnat}
\bibliography{main}

\clearpage


\beginappendix
\input{sec/6_suppl}

\end{document}

%% file: sec/0_abstract.tex
\abstract{
Multimodal models have achieved remarkable progress in recent years. Nevertheless, they continue to exhibit notable limitations in spatial understanding and reasoning, the very capability that anchors artificial general intelligence in the physical world. 
With the recent release of GPT-5, allegedly the most powerful AI model to date, it is timely to examine where the leading models (GPT, Gemini, Grok, Seed, Qwen, and Intern) stand on the path toward spatial intelligence (SI). 
We thus propose \textbf{\name} for holistic \underline{\textbf{E}}valuation of multimod\underline{\textbf{A}}l LLMs on \underline{\textbf{S}}patial \underline{\textbf{I}}ntelligence. \name conceptualizes a comprehensive taxonomy of spatial tasks that unifies existing benchmarks and a growing collection of newly curated ones, enabling systematic evaluation of state-of-the-art proprietary and open-source models. In this report, we conduct the study across \numbench key benchmarks, at a cost exceeding \textit{ten billion} total tokens.  
Our empirical study then reveals that (1) GPT-5 demonstrates unprecedented strength in SI, yet (2) still falls short of human performance significantly across a broad spectrum of SI-tasks. Moreover, we (3) show that SI-tasks expose greater model capability deficiency than non-SI tasks, to the extent that (4) proprietary models do not exhibit a decisive advantage when facing the most difficult ones. 
In addition, we conduct a qualitative evaluation across a diverse set of scenarios that are intuitive for humans, yet fail the most advanced multimodal models.
\name is an ongoing community effort: we have open-sourced the \name codebase that provides a one-stop and reproducible solution for SI assessment with standardized interfaces, integrated protocols and prompts that significantly reduce the friction of configuring and running multiple benchmarks; we have also launched an accompanying \name leaderboard to provide a continually updated snapshot of model performance across the full SI spectrum, accelerating collective progress toward robust SI.

\checkdata[Codebase]{\url{https://github.com/EvolvingLMMs-Lab/EASI/}}
\checkdata[Leaderboard]{\url{https://huggingface.co/spaces/lmms-lab-si/EASI-Leaderboard}}

}

%% file: sec/1_intro.tex
\section{Introduction}
\label{sec:intro}

\begin{figure}[t]
\centering
\includegraphics[width=\textwidth]{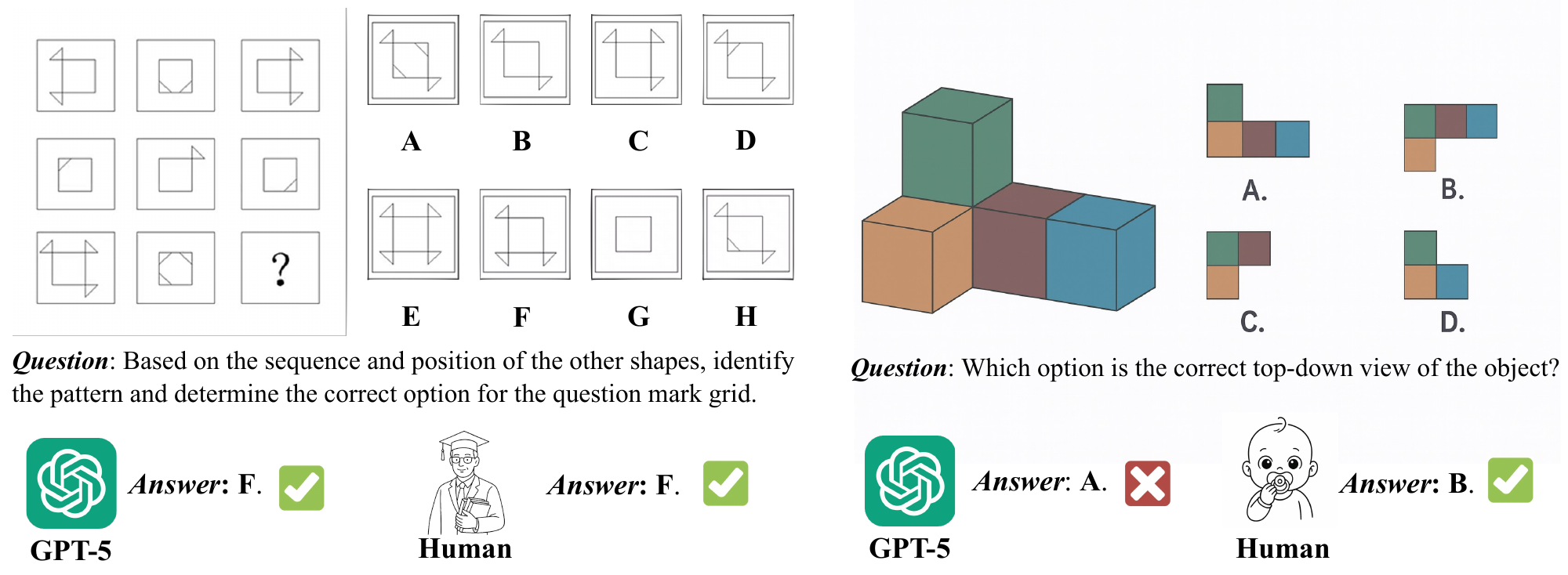}
\vspace{-6mm}
\captionof{figure}{While GPT-5~\cite{openai_gpt5_systemcard} excels at solving complex non-spatial problems (left) that are considered challenging for humans, it surprisingly struggles with some of the most basic spatial intelligence tasks (right), which even a human child can comprehend effortlessly. 
GPT-5's detailed reasoning process for this case can be found at~\cref{appendix:case_study}.
}
\label{fig:teaser}
\end{figure}

Spatial understanding and reasoning~\cite{cheng2024spatialrgpt,gong2025space,rizvi2025spare,wang2025spatial457,fu2024blink,cai2024spatialbot,szymanska2024space3d,du2024embspatial,tong2024cambrian,ma20243dsrbench,li2025zebra,deng2025bagel,li2024proximity,kim2024openvla,zitkovich2023rt,feng2025towards} constitute a critical yet underexplored dimension of intelligence, one that is indispensable for general embodied agents as it takes spatial intelligence to fully operate in, adapt to, or interact with the physical world.
Despite the impressive advancements in multimodal large language models (MLLMs)~\cite{liu2023llava,liu2024llavanext,Qwen-VL,chen2024internvl,zhu2025internvl3,wang2025internvl3, hurst2024gpt,team2023gemini,chen2024spatialvlm,wang2025ross3d, huang2025mllms,zheng2025learning,xu2025multi,li2024llava,li2025otter,yang2025egolife,wu2025visual}, it has become evident that even the most advanced MLLMs often fail at spatial tasks that are trivially easy for humans as shown in~\cref{fig:teaser}. Recent work~\cite{li2024core} has shown that spatial intelligence (SI) is a fundamentally distinct skill, arguably one of the last underexplored frontiers, compared to the multimodal capabilities measured by mainstream benchmarks~\cite{liu2024mmbench, chen2024we, zhang2021mme, li2024seed, yu2023mm, yue2024mmmu, lu2023mathvista, lu2022learn, guan2024hallusionbench, mishra2019ocr,masry2022chartqa, zhang2024bench, sun2024parrot, hou2025tdbench, sun2024pathmmu, cheng2025video, zhu2025gobench, yan2025mmcr, li2024mmsci, chen2024mega, fu2024ocrbench, ma2024mmlongbench}.
The recent release of GPT-5~\cite{openai_gpt5_systemcard} has intensified interest in evaluating progress along this dimension of intelligence: \textit{Have leading MLLMs achieved spatial intelligence?} To answer this question, we introduce \textbf{\name} (pronounced as "easy"), a \textit{holistic} framework for \underline{\textbf{E}}valuation of multimod\underline{\textbf{A}}l LLMs on \underline{\textbf{S}}patial \underline{\textbf{I}}ntelligence. 

Formally, \name aims to enable a comprehensive evaluation of spatial intelligence through systematic testing across a broad suite of benchmarks under a clearly defined protocol.
We begin by examining existing benchmarks to facilitate collective evaluation and meaningful cross-benchmark ranking. Notably, most spatial intelligence benchmarks~\cite{ji2025visualtrans,song2025siri,zhou2025roborefer,yang2025thinking,wang2025site,yang2025mmsi,jia2025omnispatial,yin2025spatial,li2025unfolding,li2024core,wang2025spatialviz,deng2025internspatial,zhou2025vlm4d,he2025egoexobench,wu2025spatialscore,kong2025lrr,yu2025far} have been introduced only in recent months, highlighting the rapidly growing interest in this domain.
Since each benchmark targets distinct facets of spatial intelligence and adopts its own taxonomy, we consolidate them into six fundamental capability categories (\cref{fig:concept}). For each subtask in each benchmark, we annotate the main fundamental capability needed to address the problem in~\cref{appendix:detailed_bench_results}.
Next, we study evaluation protocols, as benchmark outcomes can be highly sensitive to factors beyond intrinsic model capability. Specifically, we analyze prompts, metrics, and scoring strategies to ensure accurate, consistent, and fair comparison across benchmarks.
Although \name is an ongoing effort to continuously enrich the collection of benchmarks, in this technical report, we include \numbench representative benchmarks that together provide comprehensive coverage:
VSI-Bench~\cite{yang2025thinking},
SITE~\cite{wang2025site},
MMSI~\cite{yang2025mmsi},
OmniSpatial~\cite{jia2025omnispatial},
MindCube~\cite{yin2025spatial},
STARE~\cite{li2025unfolding},
CoreCognition~\cite{li2024core}, and
SpatialViz~\cite{wang2025spatialviz}.
This list is readily extensible as new benchmarks emerge.

We then present a comprehensive evaluation of recent high-performing MLLM families, such as
GPT-5~\cite{openai_gpt5_systemcard}, 
Gemini-2.5-pro~\cite{team2023gemini}, 
Grok-4~\cite{grok4_xai_2025}, 
Seed-1.6~\cite{seed2025seed1_5vl}, 
Qwen2.5-VL~\cite{Qwen-VL}, and
InternVL3~\cite{zhu2025internvl3}
on the \numbench key spatial intelligence benchmarks, many of which have not been thoroughly examined before.
Our empirical findings reveal that 
(1) GPT-5 establishes itself as the new state of the art in spatial intelligence, surpassing strong baselines such as Gemini-2.5-pro~\cite{team2023gemini} and InternVL3~\cite{zhu2025internvl3}, and even reaching human-level performance on tasks requiring \textit{Metric Measurement (MM)} and \textit{Spatial Relations (SR)}.
(2) Despite this progress, a considerable performance gap remains between the strongest models and humans on most benchmarks, particularly in \textit{Mental Reconstruction (MR)}, \textit{Perspective-Taking (PT)}, \textit{Deformation and Assembly (DA)}, and \textit{Comprehensive Reasoning (CR)}, highlighting substantial room for improvement.
(3) Overall, even state-of-the-art MLLMs perform significantly worse on fundamental spatial intelligence tasks than on non-spatial tasks, underscoring the need for a shift in research focus.
(4) Proprietary models show no decisive advantage over their open-source counterparts on the most challenging spatial intelligence tasks, suggesting that further advances can be effectively driven by building on open-source foundations.

In addition, we conduct a case study on representative failure cases drawn both from the selected benchmarks and from in-the-wild examples, providing deeper insights into the strengths and limitations of GPT-5 and other state-of-the-art models.
The qualitative findings are highly consistent with the quantitative results: while \textit{Metric Measurement (MM)} and \textit{Spatial Relations (SR)} show relatively strong performance, the remaining four capabilities exhibit substantial room for improvement.
Several noteworthy observations emerge from this analysis:
(1) Although \textit{Spatial Relations (SR)} performs well overall, the model still exhibits certain blind spots. Most notably, insufficient handling of perspective effects.
(2) GPT-5 shows significant improvements in \textit{Mental Reconstruction (MR)} compared with previous MLLMs, successfully solving test tasks for the first time. However, it continues to fail on problems that are trivial for humans (\eg, \cref{fig:teaser} right).
(3) \textit{Perspective-Taking (PT)}, \textit{Deformation and Assembly (DA)}, and \textit{Comprehensive Reasoning (CR)} remain particularly challenging due to their reliance on integrated capabilities and multi-stage reasoning. Analysis of the reasoning trajectories suggests that a lack of fundamental spatial representations prevents models from arriving at correct conclusions, even when the overall problem-solving strategy is sound.

We hope that \name will serve as a foundation for advancing future research on spatial intelligence in MLLMs. Beyond benchmarking current progress, our work delineates the fundamental capabilities that constitute spatial intelligence, examines evaluation protocols, and surfaces the unique challenges inherent to this domain. To support the community, we release the \name codebase, a unified evaluation suite that consolidates tasks, protocols, prompts, and reproducible pipelines, together with the \name leaderboard, which provides a continually updated snapshot of model performance across the SI spectrum. Collectively, these resources establish a common ground for comparing future models, guiding methodological innovation, and fostering cumulative progress toward robust spatial intelligence.

%% file: sec/2_bench.tex
\section{Evaluation Benchmarks}
\label{sec:bench}

In this section, we elaborate on the key concepts underlying \name: a taxonomy of six fundamental capabilities (\cref{sec:6_capabilities}) that unify the core aspects of existing benchmarks, along with detailed evaluation protocols designed to ensure fair cross-benchmark comparison (~\cref{sec:bench:eval_protocols}). We also provide a brief overview of the benchmarks used in this technical report. It is important to note that \name is not confined to a specific set of benchmarks; it can be readily extended to additional ones in the future.

\subsection{Six Fundamental Capabilities}
\label{sec:6_capabilities}

Existing benchmarks focus on distinct aspects of spatial intelligence and often adopt varying taxonomies to characterize cognitive and reasoning abilities. To accommodate all benchmarks within a unified framework, we distill six fundamental capabilities from existing benchmarks with spatial intelligence components~\cite{cheng2024spatialrgpt,gong2025space,rizvi2025spare,wang2025spatial457,fu2024blink,cai2024spatialbot,szymanska2024space3d,du2024embspatial,tong2024cambrian,ma20243dsrbench,ji2025visualtrans,song2025siri,zhou2025roborefer,yang2025thinking,wang2025site,yang2025mmsi,jia2025omnispatial,yin2025spatial,li2025unfolding,li2024core,wang2025spatialviz,deng2025internspatial,zhou2025vlm4d, fu2024ocrbench, ma2024mmlongbench, he2025egoexobench,wu2025spatialscore,kong2025lrr}, as conceptually illustrated in~\cref{fig:concept}. Refer to the tables in~\cref{appendix:detailed_bench_results} for the categorization of the benchmark tasks according to the six fundamental capabilities.

\begin{figure}[h]
  \centering
  \vspace{-2mm}
  \includegraphics[width=\textwidth]{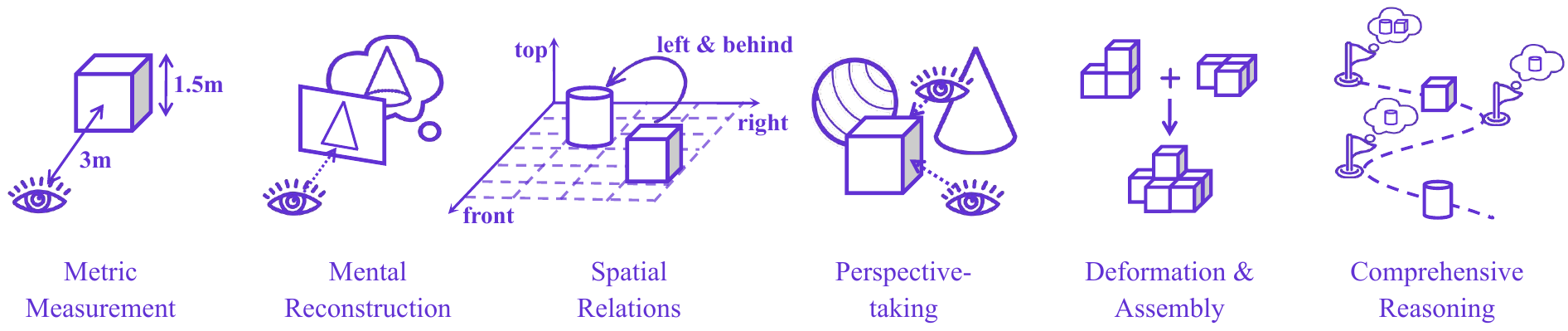}
  \vspace{-6mm}
  \caption{
    Six Fundamental Capabilities of Spatial Intelligence.
  }
  \label{fig:concept}
\end{figure}

\sensepurple{Metric Measurement (MM).} Inferring 3D dimensions (such as metric depth or lengths) from 2D observations is inherently ambiguous without additional information such as camera intrinsics. Hence, the ability to make a reasonable estimate reflects the understanding of the physical scale and typical object sizes. 

\sensepurple{Mental Reconstruction (MR).} This category assesses a model’s fine-grained geometric understanding of an object from one or more constrained viewpoints, requiring it to infer the complete 3D structure from limited 2D observations and sometimes perform virtual manipulation. While alternative viewpoints may be used to test this capability, MR differs from perspective-taking in that it involves constructing a detailed mental representation of the object, as in mental rotation tasks. Such a skill empowers real-world engineering applications, including interpreting or producing three-view drawings, and is arguably aligned with research areas such as single-view 3D object reconstruction.

\sensepurple{Spatial Relations (SR).} This capability concerns understanding the relative positions and orientations of multiple objects within the camera view. Such tasks can be seen as building upon Metric Measurement (MM) and Mental Reconstruction (MR). Typical applications include describing an object’s location relative to nearby objects. While SR does not require imagining a viewpoint transformation, it often involves conceptualizing and applying a virtual coordinate system to support the reasoning process.

\sensepurple{Perspective-taking (PT).} This ability involves reasoning across distinct viewpoints (\eg, aligning ego-centric and exo-centric perspectives). PT could subsume three components: (i) MR-like construction of a mental 3D representation of the scene, (ii) SR-like reasoning over multiple objects at the scene level, and (iii) explicit reasoning under changing camera viewpoints. A related research domain is cross-view correspondence matching. Notably, a PT problem does not necessarily involve multiple images: imagining viewpoint changes from a single image also falls within this category.

\sensepurple{Deformation and Assembly (DA).} While the preceding capabilities typically assume shape consistency, many spatial reasoning tasks go beyond this assumption. DA focuses on understanding and reasoning about deformations or structural changes. Examples include knot tying, interpreting box unfolding diagrams, and assembling multiple parts. This capability is essential for the embodied AI, where manipulation requires reasoning over such structural transformations.

\sensepurple{Comprehensive Reasoning (CR).} This category of tasks requires the coordinated use of various spatial capabilities in conjunction with extended memory and multi-stage reasoning. Examples include navigation in large, dynamic environments, and tackling spatial reasoning challenges such as long-horizon puzzle solving or mentally simulating complex physical interactions.

\subsection{Evaluation Protocols}
\label{sec:bench:eval_protocols}

\input{tables/metric_prompt}

With rapid development in spatial intelligence research, variations in \textit{system prompts} and \textit{metrics} complicate cross-benchmark evaluations. In~\cref{tab:metric_prompt}, we summarize some of the key differences in the selected benchmarks.
In this technical report, we focus on Official Protocol that uses Official Prompt \& Official Metric for direct comparison with existing baselines on the benchmarks using their official settings. In the Appendix, we also discuss about \name Protocol that has a unified setting to enable comparison between and across benchmarks. We clarify that \name Protocol is consistent across all benchmarks, whereas the Official Protocol is a collection of different evaluation settings on different benchmarks.
We also discuss additional evaluation details, such as
\textit{answer-matching methods}, and \textit{evaluation strategy} in this section.

\sensepurple{System Prompts.}
\label{sec:bench:sys_prompts}
Recent studies have shown that system prompts have a substantial impact on model performance, evaluation efficiency, and answer matching~\cite{yang2025mmsi,jia2025omnispatial,li2025unfolding}. Following the categorization in OmniSpatial~\cite{jia2025omnispatial}, we classify system prompts into three types: (1) \textit{Direct QA}, which instructs the model to answer directly without triggering chain-of-thought (CoT) reasoning; (2) \textit{Zero-shot CoT}, which prompts the model to “think step by step”; and (3) \textit{Manual CoT}, which provides structured guidance for the reasoning process.
To maintain alignment with prior work, we evaluate each benchmark using its native system prompt, referred to as the Official Prompt, enabling our results to be directly compared against existing baselines reported in their respective papers.

\sensepurple{Metrics.}
\label{sec:bench:metrics}
We report scores under two metric configurations.
\textbf{Official Metric.} For direct comparability with prior work, we adopt each benchmark’s native metric exactly as defined in its original paper. Per-benchmark metrics are summarized in \cref{tab:metric_prompt}.
Note that SITE~\cite{wang2025site} uses \textbf{\textit{Chance-Adjusted Accuracy (CAA)}} to account for the impact of random guesses on varying numbers of options in multiple-choice questions (MCQ); VSI-Bench~\cite{yang2025thinking} uses \textbf{\textit{Mean Relative Accuracy (MRA)}} for numerical-answer (NA) questions. Formal definitions appear in~\cref{appendix:evaluation_metrics}.

\sensepurple{Answer-Matching Methods.}
Variations in answer-matching methods introduce inconsistencies due to under-extraction and incorrect extraction. Following best practices from VLMEvalKit~\cite{duan2024vlmevalkit} and LMMS-Eval~\cite{lmms_eval2024,zhang2024lmmsevalrealitycheckevaluation}, we employ a two-step matching process:
\textbf{1) Initial Rule-Based Matching:} Extract answers enclosed within the ``$<$answer$><$/answer$>$'' tags, as required by our system prompt.
\textbf{2) Extended Rule-Based Matching:} If the first step fails, we draw on SpatialViz~\cite{wang2025spatialviz} to extract answers using additional patterns such as "$<$answer$>$'', "Answer:'', "Final answer'', and similar formats.
%
%
If both steps fail, the response is considered incorrect.

\sensepurple{Circular Evaluation.}
In addition, to reduce option-position bias (\ie, model performance fluctuates due to changing order of the options), circular evaluation strategies have been proposed: each multiple-choice question with $k$ possible answers is presented $k$ times, with the answer options rotated each time. Scores are computed in two variants:
\textbf{1) Soft-circular scoring:} aligned with CoreCognition~\cite{li2024core}, we measure the proportion of correct selections across all rotations, questions with rotated options are essentially considered equally weighted new questions.
\textbf{2) Hard-circular scoring:} following MMBench~\cite{liu2024mmbench}, this is a very conservative strategy that a question is only considered correctly answered if the right answers are selected for \textit{all} its variants with rotated options. 
However, considering the testing time and cost increase $k$-fold for circular evaluation, we investigate the differences between evaluation strategies in~\cref{appendix:circular_results} and discover that the \textit{standard (Non-circular)} evaluation strategy yields broadly consistent ranking results as that of circular strategies. 
Hence, we primarily adopt the \textit{standard (Non-circular)} strategy, except that we use Soft-circular protocol for CoreCognition results (\cref{tab:corecognition_ss} and \cref{appendix:bench:corecognition}) to ensure a fair comparison with the original paper.

\subsection{Benchmark Statistics}
\input{tables/bench}

To comprehensively evaluate model performance in spatial intelligence, we assess them on \numbench key benchmarks.
We summarize the key aspects of these benchmarks in~\cref{tab:bench}. We highlight that these benchmarks are released very recently, indicating the increasing research attention on spatial intelligence. 
In particular, MindCube~\cite{yin2025spatial} contains 21K questions, significantly exceeding other benchmarks. However, the three subsets of MindCube (among, around, rotation) are imbalanced, with the ``among'' subset containing 18K questions. Therefore, we adopt 
MindCube-Tiny for testing, which includes 1,050 QA pairs with a balanced distribution (among:around:rotation = 600:250:200) and 428 unique images.
Across all \numbench benchmarks, each \textit{standard evaluation (non-circular)} was evaluated on approximately \textbf{31K} images, \textbf{4.5K videos}, and \textbf{24K QA} in total.

%% file: tables/metric_prompt.tex
\begin{table}[h]
    \centering\small
    \begin{tabular}{llll}
    \toprule
        \textbf{Benchmark} & 
        \textbf{Official Metric} &
        \textbf{Official System Prompt} &
        \textbf{Output Format} \\
    \midrule
        VSI-Bench~\cite{yang2025thinking} & MRA, Acc & \textbf{Direct QA} & Single Letter  \\
        SITE~\cite{wang2025site} & CAA & \textbf{Direct QA} & Single Letter  \\
        MMSI~\cite{yang2025mmsi} & Acc & \textbf{Direct QA}, Zero-shot CoT & Single Letter \\
        OmniSpatial~\cite{jia2025omnispatial} & Acc & Direct QA, Zero-shot CoT, \textbf{Manual CoT} & Single Letter \\
        MindCube~\cite{yin2025spatial} & Acc & \textbf{Direct QA} & Template \\
        STARE~\cite{li2025unfolding} & Acc, F1 & Direct QA, \textbf{Zero-shot CoT} & Template \\
        CoreCognition~\cite{li2024core} & Acc & \textbf{Direct QA} & No Template \\ 
        SpatialViz~\cite{wang2025spatialviz} & Acc & Direct QA, \textbf{CoT} & Template \\
        
    \bottomrule
    \end{tabular}


\caption{\textbf{Overview of official evaluation configurations for the \numbench selected benchmarks}. Definitions of all metrics are provided in~\cref{appendix:evaluation_metrics}. 
For system prompt, we follow the definition of OmniSpatial~\cite{jia2025omnispatial}. SpatialViz's CoT explicitly requests the model to output its reasoning process. \textbf{For clarity, the prompt type actually used as the \textit{Official Prompt} is highlighted in boldface.}
For output formats: \emph{Single Letter} requires only the option label (\eg, A–D); \emph{No Template} means returning an answer with no extra requirements; \emph{Template} requires wrapping the answer in a specified pattern (\eg, \texttt{<answer>\ldots</answer>}).}

\label{tab:metric_prompt}
\end{table}

%% file: tables/bench.tex
\begin{table}[h]
    \centering
    \resizebox{\textwidth}{!}{
    \begin{tabular}{lcccccccccccccc}
    \toprule
        \multirow{2}{*}{\textbf{Benchmark}} & 
        \multirow{2}{*}{\textbf{YY/MM}} &  
        \multirow{2}{*}{\textbf{\#Image}} &
        \multirow{2}{*}{\textbf{\#Video}} &
        \multirow{2}{*}{\textbf{\#QA}} &
        \multirow{2}{*}{\textbf{CoT}} &
        \multicolumn{3}{c}{\textbf{Anno. Method}} &
        \multicolumn{6}{c}{\textbf{Fundamental Capabilities}} \\
        \cmidrule(lr){7-9} \cmidrule(lr){10-15}
         & & & & & & 
        \faPenNib & \faRobot & \faFile &
        \textbf{MM} & \textbf{MR} & \textbf{SR} & 
        \textbf{PT} & \textbf{DA} & \textbf{CR} \\        
    \midrule
        VSI-Bench~\cite{yang2025thinking} &24/12 &- &288 &5K &- & 
            \checkmark & - & \checkmark & 
            \checkmark & - & \checkmark & \checkmark & - & \checkmark \\
        SITE~\cite{wang2025site} &25/05  &13.2K  &3.8K  &8.1K &- & 
            \checkmark & \checkmark & - &
            \checkmark & - & \checkmark & \checkmark & - & \checkmark \\
        MMSI~\cite{yang2025mmsi} &25/05  &2K  &-  &1K  & \checkmark & 
            \checkmark & - & - &
            \checkmark & \checkmark & - & \checkmark & - & \checkmark \\
        OmniSpatial~\cite{jia2025omnispatial} &25/06  &1.3K  &-  &1.5K  &- &
            \checkmark & - & - &
            \checkmark & - & - & \checkmark & - & \checkmark  \\
        MindCube~\cite{yin2025spatial} &25/06  &3.2K  &-  &21.1K  &- &
            - & - & \checkmark & - & 
            - & - & \checkmark & - & -\\
        STARE~\cite{li2025unfolding} &25/06  & 10.3K  &-  &4K  &- &
            - & - & \checkmark & - & 
            - & - & \checkmark & \checkmark & \checkmark \\ 
        CoreCognition~\cite{li2024core} &25/06  & 1.5K  & 217  &1.5K  &- &
            \checkmark & - & \checkmark & 
            - & - & \checkmark & \checkmark & - & - \\ 
        SpatialViz~\cite{wang2025spatialviz} &25/07  &1.2K  &-  &1.2K  &- &
            \checkmark & - & \checkmark &
            - & \checkmark & \checkmark & - & \checkmark & \checkmark \\ %
        
    \bottomrule
    \end{tabular}}
    \caption{\textbf{Overview of the key aspects of the \numbench selected benchmarks}. YY/MM: the year and the month of publication or preprint release. CoT: whether to have chain-of-thought labels. Anno. Method: annotation method (\faPenNib: manually annotated, \faRobot: curated with LLM, \faFile: generated with templates). Fundamental Capabilities: see~\cref{sec:6_capabilities} for definitions.}
    \label{tab:bench}
\end{table}

%% file: sec/3_results.tex
\section{Results}
\label{sec:results}

\input{tables/oo_tables/main_results_oo}


We summarize the results of the leading proprietary and open-source models in ~\cref{tab:main_results_ss} (\name Protocol for cross-benchmark comparison) and ~\cref{tab:main_results_oo} (Official Protocol that aligns with the original benchmarks). Refer to~\cref{sec:bench:eval_protocols} for details of the evaluation protocols; \textit{the \name protocol is used here if not specified}. Our key findings include
\footnote{We include additional analysis on token consumptions (\cref{sec:token_compare}) and think modes (\cref{sec:gpt5_thinking_mode}) in the Appendix.}:

\sensepurple{GPT-5 sets a new state of the art in spatial intelligence.} 
As shown in~\cref{tab:main_results_ss}, where average accuracy is computed across the \numbench benchmarks, GPT-5 surpasses all competing models and ranks first overall, followed by Gemini-2.5-pro. Among open-source models, InternVL3 outperforms Qwen2.5-VL, emerging as the strongest performer in this category. Notably, GPT-5 consistently achieves the best results across benchmarks, with the sole exception of MindCube.
We further highlight that the SoTA models show remarkable performance on two fundamental spatial intelligence tasks: \textbf{Metric Measurement (MM)} and \textbf{Spatial Relations (SR)}. For example, GPT-5 demonstrates superior ability in MM tasks (\eg, object or room size estimation) in VSI-Bench (\cref{appendix:bench:vsi}), and excels at SR tasks (\eg, 3D information understanding, spatial relationship reasoning) in SITE (\cref{appendix:bench:site}), outperforming humans. These results indicate that state-of-the-art MLLMs have begun to acquire more basic spatial intelligence capabilities, particularly in making informed dimensional estimations and performing straightforward spatial deductions between objects from a single view.
Despite this progress, we urge caution in interpretation. The human performance baseline on SITE is noticeably lower than other benchmarks, and more challenging variants of MM and SR tasks, such as Attribute (Measurement) in MMSI (\cref{appendix:bench:mmsi}) and Block Moving in SpatialViz (\cref{appendix:bench:spatialviz}), remain insufficiently addressed by current models.

\sensepurple{Despite these advances, a substantial gap remains between model and human spatial intelligence.}
Although leading MLLMs (\eg, GPT-5) represent a significant leap forward, their performance still lags far behind human-level performance in core aspects of spatial reasoning, revealing research opportunities. Notable gaps, often exceeding 30 percentage points, persist across the following fundamental capabilities:
\textbf{(1) Perspective-Taking (PT)} remains the most fundamental and prevalent challenge in spatial intelligence, requiring reasoning from multiple viewpoints. We observe large performance discrepancies between humans and the best model in a total of 18 PT-related subtasks. Prominent examples include: MMSI's (\cref{tab:mmsi_ss}) eight PT subtasks show gaps as high as 80 points, and MindCube (\cref{tab:mindcube_ss}) with all its three subtasks fall under PT and an average gap of around 50 points. 
\textbf{(2) Comprehensive Reasoning (CR)} demands multi-step inference and sustained spatial memory. Across eight CR subtasks, significant performance gaps remain. OmniSpatial (\cref{tab:omnispatial_ss}) exemplifies this difficulty, with three subtasks showing disparities of up to 50 points. Similarly, VSI's (~\cref{tab:vsi_ss}) CR tasks (Route Planning and Appearance Order) show gaps up to 68 points.
\textbf{(3) Deformation and Analysis (DA)} encompasses reasoning over complex shape transformations and part-based compositionality. SpatialViz (\cref{tab:spatialviz_ss}) serves as a key benchmark, where five DA subtasks exhibit performance gaps ranging from over 30 to more than 50 points.
\textbf{(4) Mental Reconstruction (MR)} requires reconstructing 3D structures from limited observations. Alarmingly, GPT-5 performs \textit{worse than random guessing} on two MR subtasks (scoring below 0 on the CAA metric): the Attribute (Appearance) task in MMSI (\cref{tab:mmsi_ss}) and the 3D Mental Rotation task in SpatialViz (\cref{tab:spatialviz_ss}).

\sensepurple{Spatial intelligence (SI) tasks pose significantly greater challenges than non-SI tasks.}
A clear illustration of this difficulty is provided by MMSI (\cref{tab:mmsi_ss}), a comprehensive benchmark composed entirely of SI-related subtasks, where GPT-5 still scores more than 76 percentage points below human performance on average. 
In contrast, models have already achieved human-level accuracy on a handful of non-SI tasks in CoreCognition (\cref{tab:corecognition_ss}), including Boundary, Perceptual Constancy, Conservation, and the entire Formal Operation category.
Further comparisons between analogous tasks highlight the increased complexity of spatial reasoning. In SpatialViz (\cref{tab:spatialviz_ss}), models perform close to, even surpassing, humans on \textit{all} non-SI tasks, yet struggle with SI tasks under the same categories. For example, the best model performance on the 3D Mental Rotation task is roughly 46 percentage points lower than on the 2D Mental Rotation task.
Similarly, the best model perform about 20 percentage points worse on the 3D Transformation task compared to the 2D Transformation task in STARE (\cref{tab:stare_ss}).

\sensepurple{Proprietary models do not hold a decisive advantage over open-source models on difficult SI tasks.} 
While proprietary models generally outperform open-source counterparts, this advantage narrows considerably when it comes to the most challenging spatial intelligence tasks. 
For instance, in subtasks where model performance remains more than 60 percentage points below human levels, the performance gap between open- and closed-source models is typically around or less than 15 points in MMSI (\cref{tab:mmsi_ss}). 
Moreover, the strongest open-source model can even match or surpass the leading proprietary model in specific categories, such as Geometric Reasoning and Hypothetical Reasoning, in OmniSpatial (with official setting in \cref{tab:omnispatial_oo}).
This parity on the hardest tasks presents a timely opportunity for the research community to drive advances by building on open-source models.

%% file: tables/oo_tables/main_results_oo.tex
\begin{table*}[t]
\centering

\setlength{\tabcolsep}{1.2pt}
\resizebox{1.0\linewidth}{!}{

\begin{tabular}{lcccccccc} 
\toprule
\textbf{Models} &
\textbf{VSI~\cite{yang2025thinking}} &
\textbf{SITE~\cite{wang2025site}} &
\textbf{MMSI~\cite{yang2025mmsi}} &
\textbf{OmniSpatial~\cite{jia2025omnispatial}} &
\textbf{MindCube$^*$~\cite{yin2025spatial}} &
\textbf{STARE~\cite{li2025unfolding}} &
\textbf{CoreCognition~\cite{li2024core}} &
\textbf{SpatialViz~\cite{wang2025spatialviz}} \\

\midrule
\rowcolor{line-blue}\textbf{Metric} & MRA, Acc & CAA & Acc & Acc & Acc & Acc, F1 & Acc & Acc \\

\midrule
\rowcolor{line-blue}\textbf{Random Choice} & 34.00 & 0.0 & 25.00 & 24.98 & 32.35 & 34.80 & 33.93 & 25.08 \\
\midrule
\rowcolor{line-blue}\textbf{Proprietary Models} & & & & & & & & \\

Seed-1.6-2025-06-15~\cite{seed2025seed1_5vl} 
& 49.91 
& 54.61 
& \cellcolor{secondcolor}{38.30}
& 49.32 
& 48.75 
& 46.06
& 77.17
& 34.58
\\

Gemini-2.5-pro-2025-06~\cite{team2023gemini} 
& \cellcolor{secondcolor}{53.57} 
& \cellcolor{secondcolor}{57.06} 
& 38.00 
& 55.38 
& \cellcolor{secondcolor}{57.60}
& 49.14
& 76.70
& 42.71
\\

Grok-4-2025-07-09~\cite{grok4_xai_2025}
& 47.92
& 47.01
& 37.80
& 46.84
& \cellcolor{bestcolor}{\textbf{63.56}}
& 26.90
& \cellcolor{secondcolor}{79.27}
& 19.40$^{\dagger}$
\\

GPT-5-nano-2025-08-07~\cite{openai_gpt5_systemcard}
& 43.22
& 35.81
& 28.90
& 47.81
& 41.48
& 46.05
& 67.92
& 35.59
\\

GPT-5-mini-2025-08-07~\cite{openai_gpt5_systemcard}
& 48.67
& 52.47
& 34.10
& \cellcolor{secondcolor}{55.52}
& 56.69 
& \cellcolor{secondcolor}{52.51}
& 77.77
& \cellcolor{secondcolor}{44.66}
\\

GPT-5-2025-08-07~\cite{openai_gpt5_systemcard}
& \cellcolor{bestcolor}{\textbf{55.03}}
& \cellcolor{bestcolor}{\textbf{61.88}}
& \cellcolor{bestcolor}{\textbf{41.80}} 
& \cellcolor{bestcolor}{\textbf{59.90}}
& 56.30 
& \cellcolor{bestcolor}{\textbf{54.59}}
& \cellcolor{bestcolor}{\textbf{84.37}}
& \cellcolor{bestcolor}{\textbf{51.27}}
\\

\midrule

\rowcolor{line-blue}\textbf{Open-source Models} & & & & & & & & \\


Qwen2.5-VL-3B-Instruct~\cite{Qwen-VL} & 27.00 & 33.14 & 28.60 & 42.47 & 37.60 & 37.83 & 60.19 & 21.86\\

Qwen2.5-VL-7B-Instruct~\cite{Qwen-VL} & 32.30 & 37.64 & 26.80 & 39.07 & 36.05 & 35.03 & 62.16 & 26.78\\

Qwen2.5-VL-72B-Instruct~\cite{Qwen-VL} & 35.77 & \cellcolor{secondcolor}{47.41} & \cellcolor{bestcolor}{\textbf{32.50}} & \cellcolor{secondcolor}{47.81} & 42.40 & 38.37 & 69.22 & \cellcolor{bestcolor}{\textbf{32.54}}\\

InternVL3-8B~\cite{zhu2025internvl3} & 42.14 & 41.15 & 28.00 & 46.25 & 41.54 & \cellcolor{secondcolor}{41.36} & 60.92 & 30.00\\

InternVL3-78B~\cite{zhu2025internvl3} & 47.55 & \cellcolor{bestcolor}{\textbf{52.72}} & 30.50 & \cellcolor{bestcolor}{\textbf{50.95}} & \cellcolor{bestcolor}{\textbf{49.52}} & \cellcolor{bestcolor}{\textbf{42.00}} & \cellcolor{bestcolor}{\textbf{71.16}} & \cellcolor{secondcolor}{31.10}\\

InternVL3.5-8B~\cite{wang2025internvl3} & \cellcolor{secondcolor}{56.05} & 43.79 & 27.30 & 46.71 & \cellcolor{secondcolor}{42.50} & 40.18 & 66.40 & 23.98\\

Qwen3-8B-Instruct~\cite{yang2025qwen3technicalreport} & \cellcolor{bestcolor}{\textbf{57.90}} & 45.83 & \cellcolor{secondcolor}{31.10} & 45.73 & 29.42 & 39.76 & \cellcolor{secondcolor}{69.67} & 17.54$^{\dagger}$ \\

\midrule
\rowcolor{line-blue}\textbf{Human Evaluation} & & & & & & & & \\
\footnotesize $\Delta$(Best Model,Human) & 
\colornum{-21.3} &
\colornum{-5.62} &
\colornum{-55.40} &
\colornum{-32.73} &
\colornum{-30.99} &
\colornum{-42.06} &
\colornum{-2.61} &
\colornum{-31.19} \\

\textbf{Human} & \textbf{79.2}& \textbf{67.5}  & \textbf{97.2}& \textbf{92.63}& \textbf{94.55} & \textbf{96.50} & \textbf{86.98} & \textbf{82.46} \\

\bottomrule
\end{tabular}
}
\vspace{-2pt}
\caption{\textbf{Evaluation on \numbench recent spatial benchmarks (Official Protocol).} Each metric reported in the table is consistent with the definitions in the original papers (see \cref{appendix:evaluation_metrics}). In addition, we strictly follow the original prompt specified by each benchmark, either from the paper or the released code. Metrics across different columns are not directly comparable if they differ. 
MindCube$^*$ denotes MindCube-Tiny. VSI random choice here is chance level(Frequency). 
$^{\dagger}$ indicates cases where generations were truncated due to overlong chains of thought, yielding no final answer; such instances are counted as incorrect, which depresses the score.
\textbf{\raisebox{0pt}[0pt][0pt]{\colorbox{bestcolor}{Dark purple}}} highlights the best result and \raisebox{0pt}[0pt][0pt]{\colorbox{secondcolor}{light purple}} indicates the second-best result within Proprietary and Open-source models, respectively. 
Detailed results on each benchmark can be found in~\cref{appendix:detailed_bench_results}. }
\label{tab:main_results_oo}
\end{table*}

%% file: sec/4_case_study.tex
\section{Case Study}
\label{sec:case_study}

\begin{figure}[t!]
  \centering
  \includegraphics[width=\textwidth]{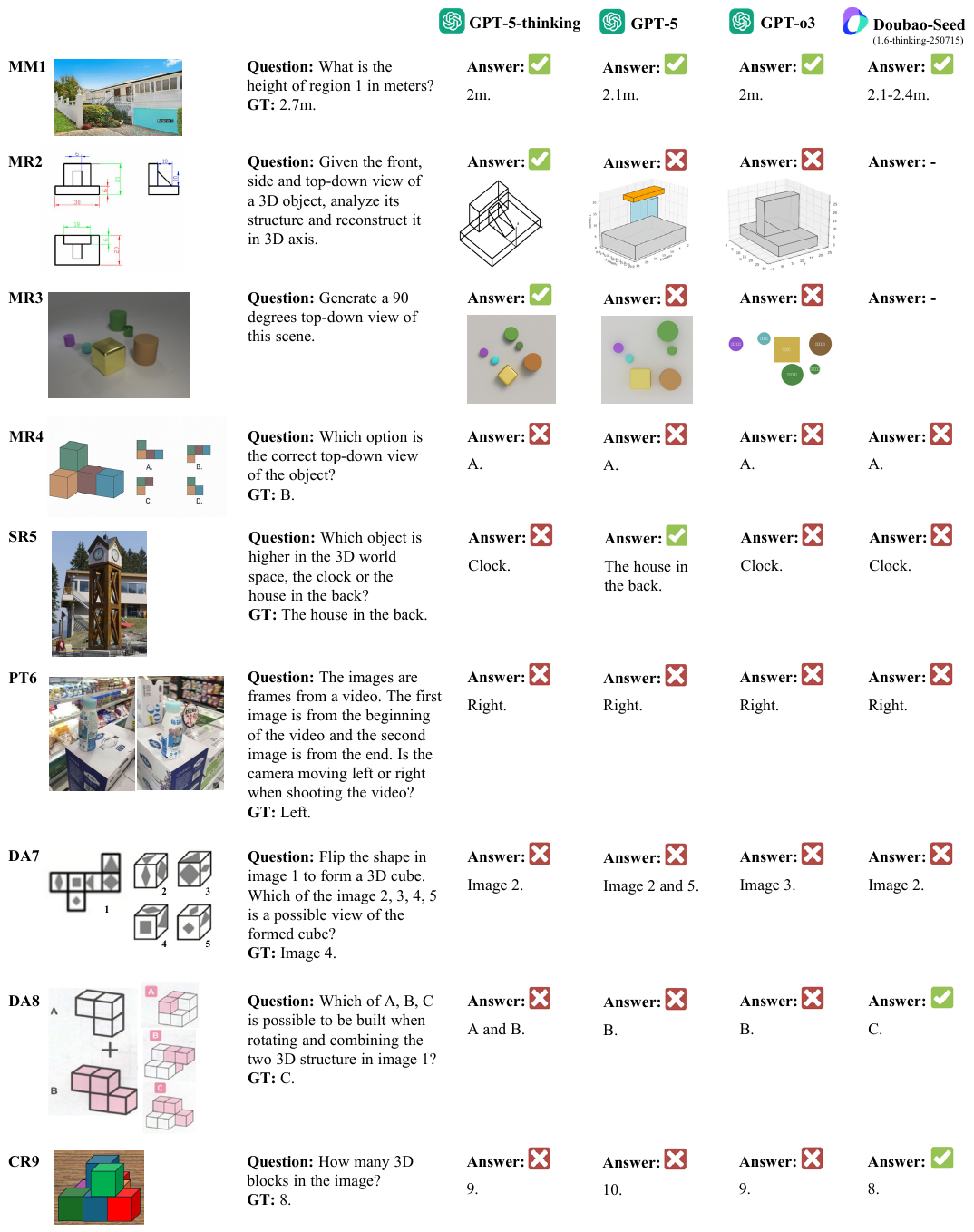}
  \captionof{figure}{\textbf{Case Study.} We compare the performance of GPT-5 with thinking capability (GPT-5-thinking), the standard GPT-5 model, the previous strong thinking model GPT-o3~\cite{gpt_o3}, and another leading reasoning model, Doubao-Seed-1.6-thinking~\cite{seed2025seed1_5vl}.
  While GPT-5-thinking exhibits notable improvements over its predecessors, it remains far from conquering the full spectrum of spatial intelligence.
  For MR2 and MR3, Doubao-Seed-1.6-thinking is exempted from visual comparisons because it cannot generate images.
  Note in this comparison, the web-based services are used.
  The reasoning output and more examples can be found in~\cref{appendix:case_study}.}
  \label{fig:compare}
\end{figure}

In~\cref{fig:compare}, we conducted a qualitative evaluation focusing on GPT-5~\cite{openai_gpt5_systemcard} and its improvements over its predecessor, GPT-o3~\cite{gpt_o3}. More detailed thinking processes are included in~\cref{appendix:case_study}. Note that instead of using API (like that in all quantitative analyses), we use the website platform for all case studies (\eg, \textit{GPT-5} and \textit{GPT-5-thinking}).
Our findings are largely consistent with previous analysis: GPT-5 excels in some tasks (\eg, MM), but it remains far from achieving human-level spatial intelligence in general. Our evaluation highlights several key findings by capabilities:

\sensepurple{Metric Measurement (MM).}
GPT-5 performs reliably on basic real-world images, as illustrated in~\cref{fig:compare} (MM1), suggesting that it possesses fundamental knowledge of object dimensions in everyday contexts. We observe that GPT-5 typically follows a principled reasoning process: it first identifies the objects present in the scene, then draws upon its extensive repository of common-sense knowledge about their typical sizes, and finally makes a dimensional estimation based on that information. It is important to note that MM is inherently an ambiguous task without camera intrinsics, and therefore a relatively large margin of error is generally considered acceptable in such estimations.

\sensepurple{Mental Reconstruction (MR).}
This category shows mixed results. On the positive side, GPT-5 demonstrates, for the first time, strong capabilities in reconstructing objects from multiple views (\cref{fig:compare}, MR2). Moreover, it significantly outperforms o3~\cite{gpt_o3} in novel view generation, particularly when the thinking mode is activated, resulting in correct top-down views (\cref{fig:compare} MR3). However, we also observed that it is highly sensitive to prompts, with only specific prompts capable of eliciting accurate view generation. Furthermore, MR4 reveals a surprising limitation (also shown in~\cref{fig:teaser}): a task that is trivially easy for a human child fails across all state-of-the-art MLLMs, underscoring the persistent challenges of spatial understanding in this domain.

\sensepurple{Spatial Relations (SR).}
SR is generally well-addressed. However, there are still cases that may confuse the models. As shown in~\cref{fig:compare} SR5, the scene becomes more complex with multiple objects and visual illusions due to the perspective effect. In such cases, GPT-5 fails to correctly infer the true relative sizes of objects, showing no substantial improvement over o3~\cite{gpt_o3}, revealing a lack of robust understanding of spatial relationships between objects and their impact on the apparent physical scale, underscoring a key limitation in GPT-5’s spatial reasoning capabilities.

\sensepurple{Perspective-taking (PT).}
GPT-5 struggles to reason between changing camera viewpoints, especially when the view overlap is relatively minimal (\cref{fig:compare} PT6), which is difficult for all state-of-the-art models.
From its thinking process~\cref{appendix:case_study}, we observed that GPT-5 attempts to establish visual correspondence across different perspectives.
However, it misinterprets the camera's rotation, suggesting that it has not developed a solid ability for perspective transformation.
We emphasize that PT is a core component of spatial intelligence in the physical 3D world that is difficult to emerge from mainstream multimodal tasks~\cite{li2024core}.

\sensepurple{Deformation and Assembly (DA).}
This remains a critical weakness. GPT-5 fails in tasks requiring mental folding or reasoning about structural transformations, such as folding a 2D net into a 3D cube (\cref{fig:compare} DA7) and assembly of objects (\cref{fig:compare} DA8), highlighting its limitations in reasoning beyond rigid shapes, which may be attributed to a lack of effective training data that leads to this fundamental knowledge deficiency.

\sensepurple{Comprehensive Reasoning (CR).}
CR consists of multi-stage spatial reasoning tasks that require extended memory and logical deduction.
A representative task of such reasoning is counting partially occluded objects (\cref{fig:compare}, CR9). Surprisingly, GPT-5 struggles even with this seemingly simple task: while it reliably recognizes the visible blocks, it fails to infer the presence of hidden ones through spatial reasoning. This suggests that the model lacks a fundamental understanding of underlying physical principles beyond what is directly plausible in the pixel space. The detailed thinking process is showcased in~\cref{appendix:case_study}.

%% file: sec/5_conclusion.tex
\section{Conclusion}
\label{sec:conclusion}

In this technical report, we demonstrate that spatial intelligence remains a critical challenge for large multimodal models through the \name framework: we propose a set of fundamental capabilities to unify existing spatial intelligence benchmarks and study evaluation protocols for a robust analysis of model performance. While the leading MLLMs demonstrate exceptional performance and set a new state of the art in spatial intelligence, there remain key areas in which even the most advanced models severely fall short of human performance, particularly \textit{Mental Reconstruction (MR)}, \textit{Perspective-Taking (PT)}, \textit{Deformation and Assembly (DA)}, and \textit{Comprehensive Reasoning (CR)}. Closing these gaps will be essential for advancing multimodal models toward robust spatial reasoning, a crucial step on the path toward embodied general intelligence. To accelerate progress, we release the \name evaluation toolkit and an accompanying \name leaderboard. \name provides a one-stop, unified environment for spatial intelligence assessment: standardized interfaces, integrated official protocols and prompts, and reproducible pipelines that substantially reduce the overhead of configuring and running disparate benchmarks, thereby enabling holistic and consistent evaluation across models.

%% file: sec/6_suppl.tex

\section{Discussions on the Evaluation Protocol}

\subsection{Evaluation Prompt Comparison}
\label{appendix:eval_prompt}

\begin{figure*}[h]
\centering
\vspace{-2mm}
\includegraphics[width=1.0\linewidth]{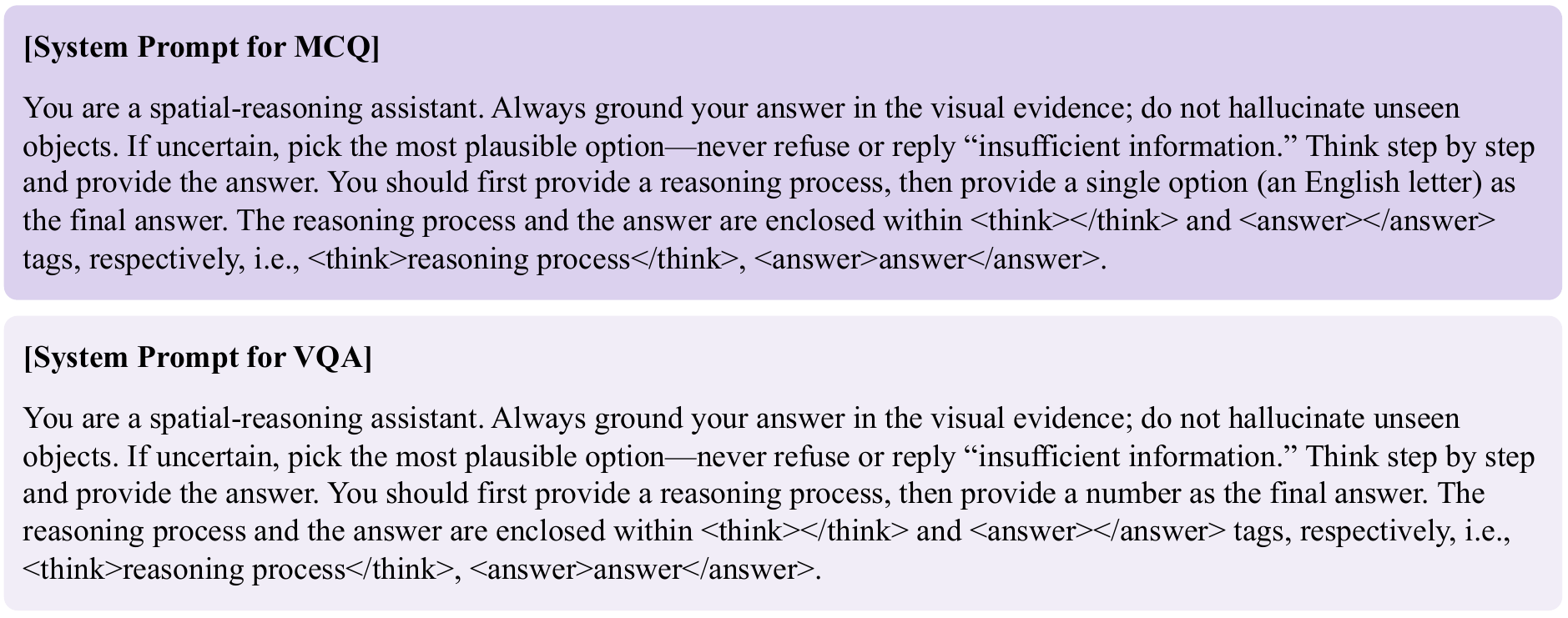}
\vspace{-4mm}
\caption{\textbf{\name Prompts} for cross-benchmark comparison. Note only results reported with \name Protocol uses \name Prompts.}
\label{fig:system_prompt}
\end{figure*}

Different benchmarks use different system prompts, introducing an additional variable that may affect model performance. To facilitate cross-benchmark comparability, we experiment with a unified system prompt (\cref{fig:system_prompt}), referred to as the \textbf{\name Prompts}. Specifically, building on observations from OmniSpatial~\cite{jia2025omnispatial} that CoT prompting generally outperforms Direct QA, we adopt the \textit{zero-shot CoT} approach as our default. Additionally, to improve answer matching accuracy, we incorporate structured answer templates inspired by SpatialViz~\cite{wang2025spatialviz}, requiring the model to enclose its responses within predefined tags.

\input{tables/two_prompt/main_results_oo_so}
In~\cref{tab:main_results_oo_so}, we compare baseline performances using our \textbf{\name} Prompts against the \textbf{Official} Prompts provided for individual benchmarks.
Focusing on the five benchmarks whose \textbf{Official} prompts follow a Direct-QA format, we find that incorporating chain-of-thought prompting (\textbf{\name}) does not yield a consistent performance trend; instead, its effectiveness is highly dependent on both the dataset and the model. 
For open-source models, we observe that the \textbf{Official} setting performs better on \textit{SITE} and \textit{MindCube}, likely due to closer alignment between the native prompt templates and the benchmarks’ answer-matching criteria. 
Outside of these cases, the performance gap between \textbf{\name} and \textbf{Official} prompts is mixed: while explicit reasoning sometimes improves results—especially on tasks requiring multi-step spatial reasoning, these gains are neither uniform across datasets nor consistent across model families.

\subsection{Evaluation Metrics}
\label{appendix:evaluation_metrics}


The \textbf{\textit{CAA}} is computed as follows:
\begin{equation}
\label{eq:caa}
CAA = \left( \sum_{i=1}^N X_i - \sum_{i=1}^N \frac{1}{n_i} \right) \Big/ \left( N - \sum_{i=1}^N \frac{1}{n_i} \right).
\end{equation}
Here, $N$ denotes the total number of questions and $n_i$ represents the number of options for the $i$-th question.Let $X_i=\mathds{1}\{\hat{y}_i=y_i\}$, where $\mathds{1}(\cdot)$ is the indicator function.\textbf{CAA = 1} indicates all questions were answered correctly, \textbf{CAA = 0} matches random guessing, and \textbf{CAA < 0} is worse than random.

In VSI-Bench~\cite{yang2025thinking}, for Numerical Answer questions, we follow the original paper and report the results using \textbf{\textit{MRA}}:
\begin{equation}
\label{eq:mra}
MRA = \frac{1}{10} \sum_{\theta \in \mathcal{C}} \mathds{1} \left( \frac{| \hat{y} - y |}{y} < 1 - \theta \right),
\end{equation}
where $y$ and $\hat{y}$ represent the ground truth and prediction, respectively, while $\theta$ denotes the confidence threshold. A prediction is considered correct only if the relative error rate $| \hat{y} - y |/y$ is below $1 - \theta$. To ensure more reliable evaluation, \textbf{\textit{MRA}} averages the scores across 10 thresholds, where $\cC = \{0.5, 0.55, ..., 0.95\}$.

And in detailed benchmark results reported in~\cref{appendix:detailed_bench_results}, we report other mertics include \textbf{\textit{Accuracy(Acc)}} and \textbf{\textit{F1 score(F1)}}.
The \textbf{\textit{F1}} score is given by:
\begin{equation}
\label{eq:f1}
F_1 = \frac{2 \cdot \mathrm{Precision} \cdot \mathrm{Recall}}
{\mathrm{Precision} + \mathrm{Recall}},
\end{equation}
\begin{equation}
\mathrm{Precision}=\frac{\mathrm{TP}}{\mathrm{TP}+\mathrm{FP}},\qquad
\mathrm{Recall}=\frac{\mathrm{TP}}{\mathrm{TP}+\mathrm{FN}},
\end{equation}

where TP, FP, and FN denote the number of true positives, false positives, and false negatives.

For \textbf{\textit{Acc}}, let $y_i$ and $\hat{y}_i$ denote the ground-truth and predicted labels for the $i$-th question. The Accuracy is defined as:
\begin{equation}
\label{eq:acc}
Acc = \frac{1}{N} \sum_{i=1}^{N} \mathds{1}\left( \hat{y}_i = y_i \right),
\end{equation}

Let $r$ denote the expected accuracy of random guessing on this set:
\begin{equation}
r=\frac{1}{N}\sum_{i=1}^N \frac{1}{n_i}.
\end{equation}
Since $\mathrm{Acc}=\frac{1}{N}\sum_{i=1}^N X_i$, the definition of CAA in \eqref{eq:caa} gives:
\begin{equation}
\mathrm{CAA}
= \frac{\sum_i X_i - \sum_i \tfrac{1}{n_i}}{N - \sum_i \tfrac{1}{n_i}}
= \frac{\tfrac{1}{N}\sum_i X_i - \tfrac{1}{N}\sum_i \tfrac{1}{n_i}}{1-\tfrac{1}{N}\sum_i \tfrac{1}{n_i}}
= \frac{\mathrm{Acc} - r}{1 - r}.
\end{equation}
Hence, for any two models evaluated on the \emph{same} benchmark (so $r$ is fixed), their difference $\Delta \mathrm{CAA}$: 
\begin{equation}
\Delta \mathrm{CAA} \;=\; \frac{\Delta \mathrm{Acc}}{1-r}
\qquad\Longleftrightarrow\qquad
\Delta \mathrm{Acc} \;=\; (1-r)\,\Delta \mathrm{CAA}.
\end{equation}

\noindent\textbf{Implication.}
Because $0<r<1$, the factor $\frac{1}{1-r}>1$, so \textbf{\textit{CAA amplifies score differences relative to Acc}} on the same dataset.
The amplification is stronger when $r$ is larger (\eg, binary questions with $r\approx 0.5$), and weaker when $r$ is smaller (many-option questions).

\subsection{Circular Strategies}
\label{appendix:circular_results}
\input{tables/circular}

We compare model performance under three evaluation protocols: \textbf{Non-circular}, \textbf{Soft-circular}, and \textbf{Hard-circular}, as mentioned in Section~\cref{sec:bench:eval_protocols} to ensure the robustness of our findings.
%
%
As shown in~\cref{tab:circular}, for a given model, a large drop from Non-circular to Soft-circular or Hard-circular indicates that part of its accuracy in the Non-circular setting may come from successful random guesses in MCQ tasks. In particular, the Hard-circular metric, which requires all rotated variants of a question to be answered correctly, serves as a stricter measure of true task competence and more reliably discriminates among model capabilities. 
Occasionally, Soft-circular scores exceed Non-circular scores because easier questions with more options are effectively repeated more times, inflating the Soft-circular average. 
More importantly, across Non-circular and Hard-circular evaluation modes, model rankings are broadly consistent, indicating that reporting only Non-circular results suffices for fair comparison while substantially reducing evaluation time and computational cost.

\input{tables/ss_tables/main_results_ss}

\section{Cross-benchmark Comparison}

As summarized in~\cref{tab:metric_prompt} the Official Protocol (Official Prompt \& Official Metric) is the focus of this technical report: results computed strictly under each benchmark’s native protocol, using the original paper’s system prompt and metric definitions. 
In~\cref{tab:main_results_ss}, however, we investigate the standardized \name Protocol to support cross-benchmark analysis: allowing the average score to be computed in a more meaningful way with a unified metric, and allowing a fair comparison between different benchmarks due to unified prompts. Specifically, we re-evaluate baseline models using 
\name Prompt (\cref{appendix:eval_prompt}) and \name Metric: for multiple-choice questions (MCQ), we use \textbf{\textit{Chance-Adjusted Accuracy (CAA)}} to account for the impact of random guesses on varying numbers of options (SITE~\cite{wang2025site}); for numerical-answer (NA) questions, we use \textbf{\textit{Mean Relative Accuracy (MRA)}} following VSI-Bench~\cite{yang2025thinking}. Formal definitions of the metrics can be found at~\cref{appendix:evaluation_metrics}. 

\section{Detailed Results by Benchmark}
\label{appendix:detailed_bench_results}

In this section, we provide detailed results for all \numbench benchmarks. For each benchmark, we include two complementary tables with the Official and \name Protocol, respectively.

\subsection{VSI-Bench}
\label{appendix:bench:vsi}
\input{tables/oo_tables/vsi_oo}

\input{tables/ss_tables/vsi_ss}

GPT-5 ranks first or very close to the top across all evaluation metrics on VSI-Bench in~\cref{tab:vsi_ss} and~\cref{tab:vsi_oo}. 
On Metric Measurement (MM), GPT-5 effectively closes the human–model performance gap, and surpassing human performance in Object Size and Room Size. This advantage likely derives from robust geometric priors acquired through large-scale training, similar to humans’ reliance on heuristic assumptions about typical object sizes.
Nevertheless, across the remaining spatial intelligence capabilities such as Perspective-taking (PT) and Comprehensive Reasoning (CR), GPT-5 continues to underperform relative to humans, indicating that while its proficiency in basic geometric estimation is comparable to or exceeds human ability, it remains less adept at handling complex, dynamic, or transformation-intensive reasoning tasks.

\FloatBarrier

\subsection{SITE}
\label{appendix:bench:site}
\input{tables/oo_tables/site_oo}
\input{tables/ss_tables/site_ss}

From \Cref{tab:vsi_oo}, GPT-5 already shows strong performance under the \textbf{Official} setting. Under the unified \textbf{\name} setting (\Cref{tab:vsi_ss}), GPT-5 attains near state-of-the-art results across almost all SITE subsets and is the only model that demonstrates consistently strong performance on the multi-view \& cross-image reasoning (PT) category—especially for egocentric–exocentric view transitions. Yet, there is still a performance gap >30 percentage points.
That said, GPT-5 remains less reliable on other forms of subject-centric viewpoint transformation, such as inferring orientation or answering questions from a hypothetical position adjacent to a specified object.
Additionally, GPT-5 approached or exceeded human performance in 3D Information Understanding (3D Inf), Spatial Relationship Reasoning (Rel), and Motion Prediction and Navigation (Mov). However, we would like to refer to~\cref{tab:main_results_ss} and point out that SITE is the only dataset with human performance at $\sim$60, whereas others at >75 or even >90. 

\FloatBarrier

\subsection{MMSI}
\label{appendix:bench:mmsi}
\input{tables/oo_tables/mmsi_oo}
\input{tables/ss_tables/mmsi_ss}

Our results in ~\cref{tab:mmsi_ss} shows minimal differences among proprietary and open-source models (\eg, <15 points) when the overall performance far below (\eg, >60 points) human level. Existing models remain limited in their ability to handle viewpoint transformations, particularly tasks requiring them to be hypothetically positioned next to a specific object and reason from that object’s perspective, highlighting persistent weaknesses in Perspective-taking (PT). 
Moreover, it is observed that challenging Metric Measurement (MM) tasks such as Attribute (Measurement), can still be problematic for even the best models.
%
%
Notably, in ~\cref{tab:mmsi_ss}, GPT-5 attains a negative CAA score on the Appr. task, indicating performance worse than random guessing.

\FloatBarrier

\subsection{OmniSpatial}
\label{appendix:bench:omnispatial}
\input{tables/oo_tables/omni_oo}
\input{tables/ss_tables/omni_ss}

In \cref{tab:omnispatial_ss}, proprietary models generally outperform open-source counterparts, with typical gaps of about 10–15 points. Moreover, in \cref{tab:omnispatial_oo} where the official setting is used, we even observe that the best open-source models outperform or are exactly on par with the proprietary SoTA. These results reveal that proprietary models do not hold decisive advantages on challenging SI tasks such as those included in OmniSpatial.

Note that in \cref{tab:omnispatial_oo}, OmniSpatial’s \textbf{Official} setting uses a \textit{manual-CoT} prompt that explicitly instructs the model on how to perform spatial reasoning (see the original Omnispatial paper~\cite{jia2025omnispatial}). For comparison, \cref{tab:omnispatial_ss} reports results under our \textbf{\name} setting, which applies a unified zero-shot CoT system prompt with SpatialViz-style answer templates (full prompt in \cref{fig:system_prompt}).

As for performance by fundamental capabilities, the \textbf{CR} and \textbf{PT} tasks show the largest gaps to human performance and are also the lowest-scoring categories overall. Across baselines, the weakest subtasks concentrate in Complex Logic and Perspective Taking, with one notable exception: the \emph{Ego-Centric} subtask. Although labeled as PT in OmniSpatial, this subtask mainly requires analyzing 2D relative positions, counts, and related attributes from a single image; it does not necessitate the kind of 3D spatial reasoning we target, and thus we do not treat it as an SI subtask in our analyses.


\FloatBarrier

\subsection{MindCube}
\label{appendix:bench:mindcube}
\FloatBarrier
\input{tables/oo_tables/mindcube_oo}

\input{tables/ss_tables/mindcube_ss}

In~\cref{tab:mindcube_oo} and~\cref{tab:mindcube_ss}, we show that \texttt{Gemni-2.5-pro} performs the best, making MindCube the only benchmark that GPT-5 is not the state of the art. 
It is interesting to find out that amongst the three subtasks (all categorized as PT), proprietary models performs no significantly better than open-sourced counterparts on ``Among" and ``Around", but the ``Rotation" task exhibits a pronounced disparity between model families, with leading closed-source systems (e.g., GPT-5~\cite{openai_gpt5_systemcard}, Seed~\cite{seed2025seed1_5vl}, Gemini~\cite{team2023gemini}), Grok-4 achieving accuracy around 85–95 points. Despite the high performance, we point out that the ``Rotation" task involves a relatively simple camera transformation: the camera remains fixed in position while rotating in place, eliminating the need for mental translation of viewpoints. Consequently, the task reduces primarily to determining the angular differences between perspectives, which is restricted to discrete values of 90° (left/right) and 180°.

Across both \cref{tab:mindcube_oo} and \cref{tab:mindcube_ss}, \texttt{Qwen2.5-VL-7B-Instruct} underperforms its \texttt{3B} counterpart on MindCube; a similar pattern is also observable in MindCube paper’s result. This inversion with respect to parameter count suggests that larger scale does not automatically translate into stronger spatial robustness on this suite. We thus caution against using model size as a proxy for spatial reasoning capability.

\FloatBarrier

\subsection{STARE}
\label{appendix:bench:stare}
\input{tables/oo_tables/stare_oo}

\input{tables/ss_tables/stare_ss}

We present results on STARE~\cite{li2025unfolding} in~\cref{tab:stare_oo} and~\cref{tab:stare_ss}. Proprietary models exhibit a pronounced advantage across all tasks, with an average gap of approximately 10-15 points compared to open-source models. 

Notably, under the \textbf{\name} setting, GPT-5 exhibits a strong ability to exploit visual-simulation (VSim) images, whereas most other models tend to underperform in this regard. In contrast, under the \textbf{Official} setting, VSim benefits GPT-5 only on \textit{Cube Net} and \textit{Tangram}, while its performance on \textit{2D Trans} and \textit{3D Trans} declines when VSim is provided.

More broadly, on \textit{2D Trans}, \textit{3D Trans}, \textit{Cube Net}, and \textit{Tangram}, we observe consistent improvements for GPT-5 under the \textbf{\name} setting when VSim images are available. These images externalize intermediate states and reduce the search space, enabling more reliable multi-step reasoning. Pairing VSim with deeper reasoning (e.g., higher \verb|effort|) yields further gains, suggesting a synergy between explicit visual cues and chain-of-thought reasoning, in line with recent observations~\cite{yang2025mmsi}.

Furthermore, model performance on SI tasks remains consistently lower than on non-SI tasks, consistent with our observations from \cref{tab:omnispatial_oo} and ~\cref{tab:omnispatial_ss}.

\FloatBarrier

\subsection{CoreCognition}
\label{appendix:bench:corecognition}
\input{tables/oo_tables/core_oo}

\input{tables/ss_tables/core_ss}



From~\cref{tab:corecognition_oo} and ~\cref{tab:corecognition_ss}, we observe that GPT-5 substantially outperforms all other proprietary and open-source models, yet still remains below human performance on Perspective-taking (PT) sub-task. Moreover, the leading MLLMs have demonstrated surprising capabilities in non-SI tasks, outperforming humans in Boundary, Perceptual Constancy, Conservation, and all sub-tasks under Formal Operation.
%
%

\FloatBarrier

\subsection{SpatialViz}
\label{appendix:bench:spatialviz}
\input{tables/oo_tables/spatialviz_oo}

\input{tables/ss_tables/spatialviz_ss}
From~\cref{tab:spatialviz_oo} and ~\cref{tab:spatialviz_ss}, we observe that proprietary models generally outperform open-source models. On non-SI tasks such as Arrow Moving (AM) and Cube Counting (CC), model performance approaches even exceeds human-level accuracy; however, on spatial tasks, all models still lag substantially behind humans, with Paper Folding (PF), exemplifying the category of Deformation and Assembly (DA), showing the largest gap and underscoring its difficulty as a spatial intelligence challenge.
Additionally, comparing 2D Rotation (2DR) and 3D Rotation (3DR), we reveal that despite their similar concepts, mental rotation in 3D requires strong SI, and is thus much more challenging. GPT-5 even performs worse than random guessing under the \name setting on 3DR, and its performance on 3DR is >60 percentage points worse than its 2D counterpart under the official setting.
Moreover, Box Moving, a challenging variant of the SR task, is not well addressed by even the strongest model. We also observe GPT-5 performs worse than random guessing (<0 CAA score in~\cref{tab:spatialviz_ss}) on some Mental Rotation (3D Rotation, or 3DR) and Metal Folding (Cube Reconstruction, CR) tasks.
We also note a robustness issue related to overlong CoT. SpatialViz contains 1{,}180 items; with \verb|max_completion_tokens| set to 16{,}384, all models produced complete answers except \textit{Grok-4}, which incurred truncation on 300+ items, leading to an artificially lower overall score for that model on this benchmark.

\FloatBarrier

\section{Token Consumption}
\label{sec:token_compare}

\begin{figure}[t]
    \centering
    \begin{subfigure}[t]{0.48\textwidth}
        \centering
        \includegraphics[width=\textwidth]{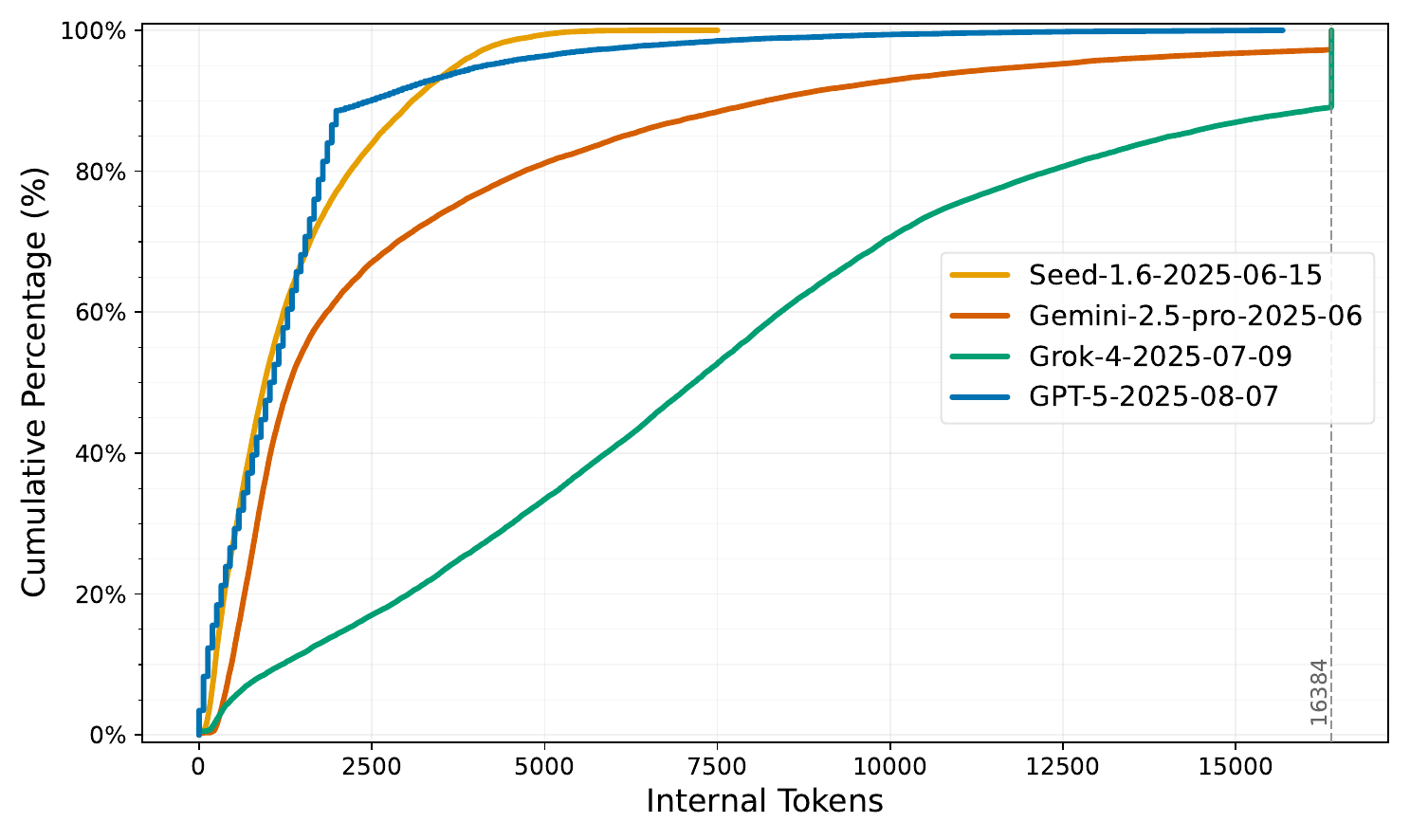}
        \label{fig:reasoning}
    \end{subfigure}
    \hfill
    \begin{subfigure}[t]{0.48\textwidth}
        \centering
        \includegraphics[width=\textwidth]{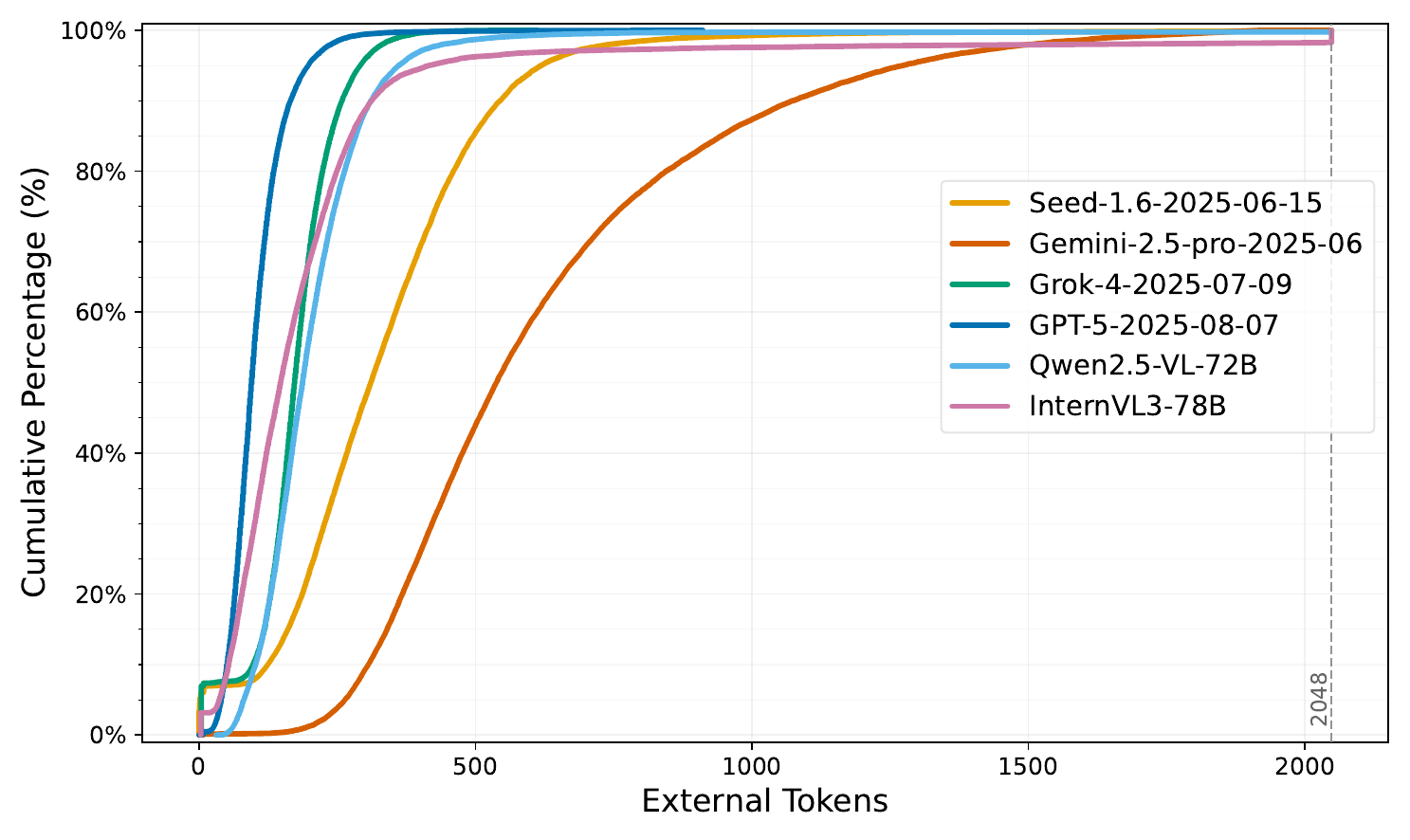}
        \label{fig:pred}
    \end{subfigure}
    \vspace{-6mm}
    \caption{Cumulative distribution of token consumptions. The horizontal axis represents the token usage, whereas the vertical axis represents the cumulative percentage of questions. Left: internal reasoning tokens (not applicable to open-source models); Right: externalized reasoning tokens.}
    \label{fig:token-overview}
\end{figure}

\input{tables/token}

Reasoning has been recognized as a central component of model understanding capabilities since the discovery of CoT~\cite{wei2022chain}. However, proprietary models often conceal their full reasoning trajectory, exposing only the final outputs to users.
To distinguish between these two forms of information, 
To distinguish these, we define tokens used in the hidden reasoning process as ``internal tokens" and user-visible outputs as ``external tokens''.
Note that the selected open-source models do not have such an internal process, thus consist solely of external tokens.

Understanding reasoning patterns is critical for configuring inference parameters in spatial intelligence tasks.
For instance, we find that the default setting of \verb|max_tokens| = 2{,}048 
is insufficient for GPT-5 to complete complex spatial reasoning, particularly on the SpatialViz benchmark~\cite{wang2025spatialviz}, where it leads to frequent failures from truncated outputs in over 50\% of cases.
To ensure reliable evaluation, we standardize \verb|max_tokens| \textbf{=16{,}384 for proprietary models}, while \verb|max_tokens| \textbf{= 2{,}048 suffices for open-source models}.

We observe substantial variation in reasoning behavior among leading MLLMs when tackling spatial intelligence tasks.
~\cref{fig:token-overview} shows the cumulative distribution of token consumption across all benchmark responses.
The results reveal pronounced hidden-stage reasoning token usage in Grok-4 and Gemini-2.5-pro, with Grok-4 in particular frequently reaching the 16{,}384-token cap, suggesting a non-negligible risk of output truncation.
Conversely, GPT-5 and Seed-1.6 exhibit more constrained internal reasoning processes. Notably, GPT-5 displays a discrete pattern in its reasoning token consumption, characterized by stepwise increments before a turning point around 2{,}048 tokens, beyond which the reasoning length grows substantially for approximately 10\% of queries.
Regarding external tokens (user-accessible outputs), Gemini-2.5-pro consistently produces longer responses than all other models, with Seed-1.6 being the second most verbose. By contrast, GPT-5, Grok-4, and the two open-source models tend to generate more concise outputs. 
Interestingly, Grok-4, Seed-1.6, and InternVL3 produce approximately 2.5–7.5\% of responses that are extremely short ($\sim$ 0 tokens), whereas Gemini-2.5-pro rarely outputs responses shorter than 100 tokens.
For certain tasks (\textit{e.g.}, SITE, STARE), Seed-1.6 occasionally outputs single-character responses, while Grok-4 may produce no external tokens due to maxing out its internal token limit.

\paragraph{Exposure ratio.}
In~\cref{tab:token}, we introduce the \emph{exposure ratio}, defined as
\[
\text{Exposure Ratio}=\frac{\text{Mean External Tokens}}{\text{Mean Internal Tokens}}
\]
This metric captures the tendency between externalized and internal reasoning. A higher ratio indicates that a model externalizes more of its reasoning in the user-facing output, while a lower ratio suggests heavier reliance on hidden internal reasoning.
Leading MLLMs exhibit distinct strategies: Seed-1.6 and Gemini-2.5-pro show relatively high exposure ratios, externalizing more reasoning steps. In contrast, GPT-5 and Grok-4 produce concise outputs while keeping the majority of their reasoning internal.
Moreover, we highlight that \boldmath$p_{95}$ values are practically important when determining safe token caps, since 95\% of responses will not exceed this length.

Joint analysis of Figure~\ref{fig:token-overview} and Table~\ref{tab:token} highlights GPT-5 as the most efficient model, with minimal internal token use, compact outputs, and state-of-the-art performance across benchmarks.
Gemini-2.5-pro, in contrast, externalizes longer chains of thought and exhibits a heavy-tailed distribution in internal token consumption.
Grok-4 represents the opposite pattern, relying extensively on internal reasoning (very low exposure ratio) and therefore facing a substantial risk of truncation under fixed token budgets.

\section{Thinking Modes of GPT-5}
\label{sec:gpt5_thinking_mode}

GPT-5 allows control over its thinking modes over four reasoning efforts: \textit{Minimal}, \textit{Low}, \textit{Medium}, and \textit{High}. To investigate the accuracy–cost trade-off, we compare GPT-5’s the impact of the \verb|reasoning_effort| parameter in~\cref{tab:thinking_mode} on SpatialViz dataset, which typically incurs extremely long reasoning processes from the leading MLLMs. Particularly, we construct \textbf{SpatialViz-Tiny} by sampling one-tenth of the instances from each task in the original dataset (118 questions in total), with \verb|max_completion_tokens| set to 16{,}384.

As shown in~\cref{tab:thinking_mode}, reasoning tokens and runtime increase significantly with the thinking mode level, and accuracy improves from \textit{Minimal} to \textit{Medium}, indicating expected benefits from reasoning. 
However, we observe that the marginal benefit is diminishing. The accuracy even drops to 52.54\% in \textit{High} mode, which we find out is because 28 of the 118 questions time out ($>$15 min) or hit token length limit, and are therefore counted as incorrect. Excluding these cases, \textit{High} mode reaches 68.89\%, the best raw accuracy but it is noteworthy that the remaining questions are typically less difficult. 
In practice, while \textit{High} mode could perform the best, its substantially higher time and compute costs (leading to the risk of overlong reasoning that triggers times-out or truncation) must be weighed carefully; \textit{Medium}, which is also the default, often offers a more balanced accuracy–cost trade-off and is used in this technical report.
\input{tables/thinking_mode}

\section{Elaborated Results in the Case Study}
\label{appendix:case_study}

In this section, we showcase some typical test cases and GPT-5-thinking's response. Correct reasoning is marked \hlg{green}, whereas problematic reasoning is marked \hlr{red}.

\par\vspace{0.6em}
\noindent\rule{\textwidth}{0.9pt}
\noindent
\begin{minipage}[t]{0.2\linewidth}\vspace{0pt}
    \includegraphics[width=\linewidth,keepaspectratio]{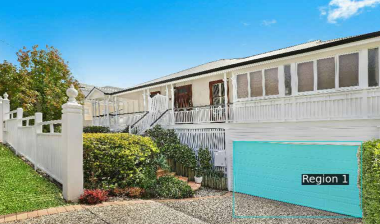}
\end{minipage}\hspace{1em}%
\begin{minipage}[t]{0.78\linewidth}\vspace{0pt}%
  \textbf{Metric Measurement (MM)}
  
  \medskip
  \textbf{Prompts:} 
What is the height of region 1 in meters?

  \textbf{GT:} 2.7m.
  
  \medskip
  \textbf{Model Responses:} 
I'll identify "Region 1" using the bounding box shown in the image, and then estimate its height through pixel-to-meter conversion, based on the dataset's expected measurements.

I’m not sure if there's a scale involved or if it’s more straightforward. Without numeric labels, \hlg{it seems "Region 1" just refers to a bounding box, likely the garage door area}. If I had to guess, \hlg{the height could be around 2 meters, which is typical for garage doors}, though I’m not fully certain.

The task might be from a dataset used in visual language models, like those evaluating layout and measurements. The region height could potentially be 2 meters, especially since it appears to be a garage door — a common dimension in Australia. I’m leaning towards this answer, assuming no scale is required.

  \textbf{Answer:} 2m.
\end{minipage}
\par\vspace{0.6em}
\noindent\rule{\textwidth}{0.9pt}
\begin{minipage}[t]{0.2\linewidth}\vspace{0pt}
    \includegraphics[width=\linewidth,keepaspectratio]{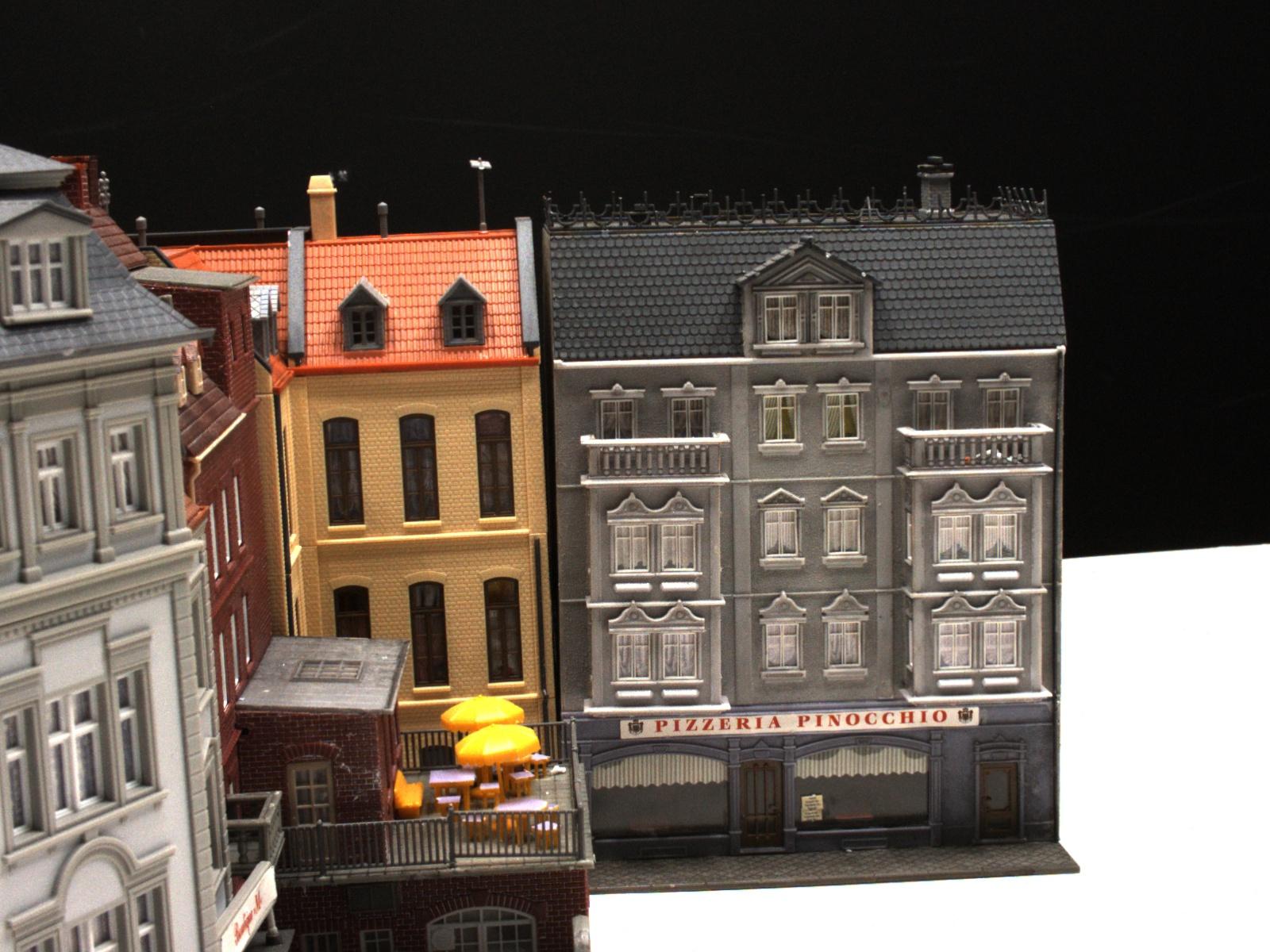}
    \includegraphics[width=\linewidth,keepaspectratio]{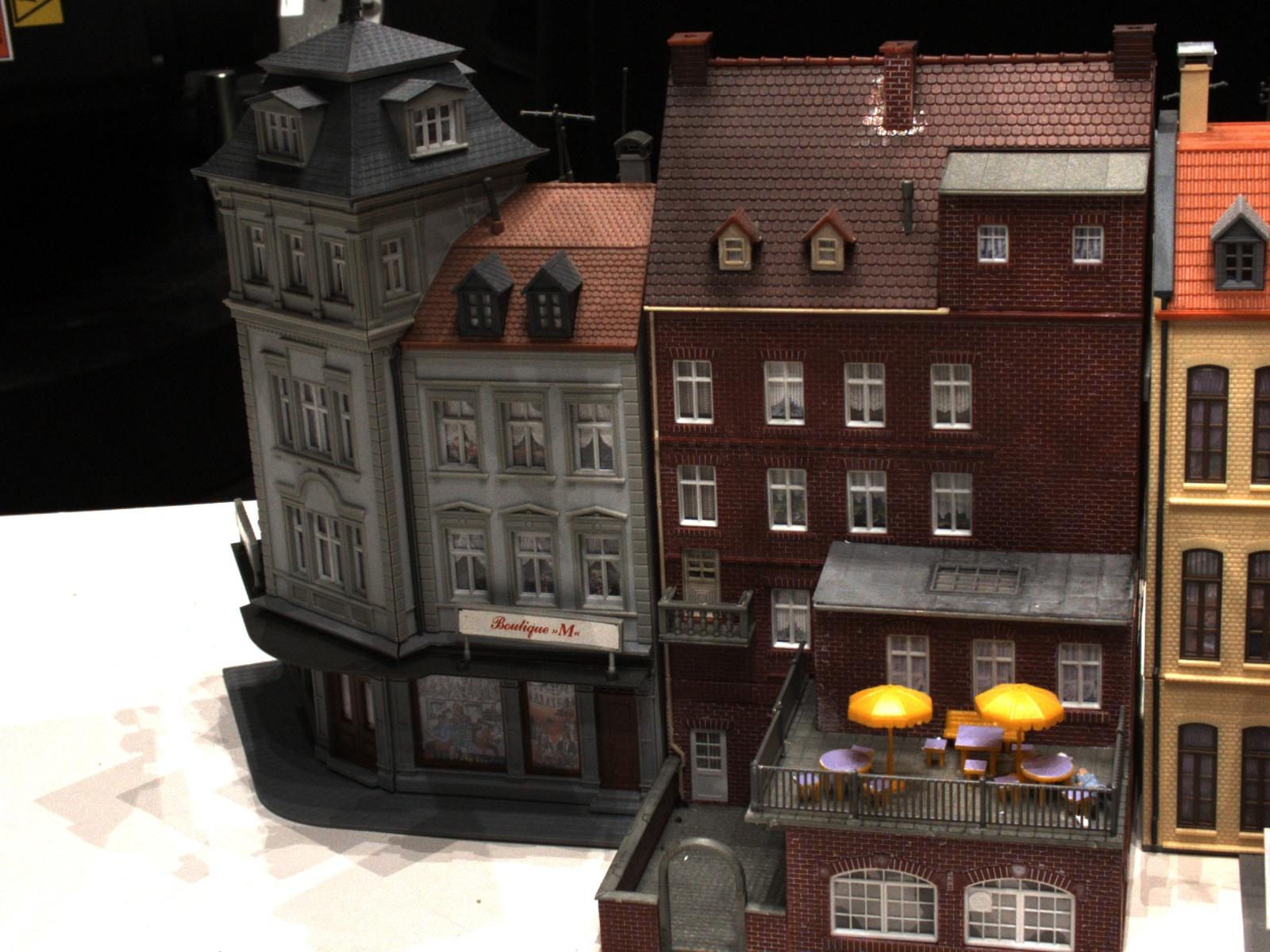}
\end{minipage}\hspace{1em}%
\begin{minipage}[t]{0.78\linewidth}\vspace{0pt}%
  \textbf{Metric Measurement (MM)}
  
  \medskip
  \textbf{Prompts:} 
In Figure 1, comparing the brown door at the bottom right corner with the door of the house with a brown roof and gray walls in Figure 2, which door is taller? Options: A: The door in Figure 2, B: The door in Figure 1, C: Sometimes the one in Figure 1 is taller, sometimes the one in Figure 2 is taller, D: They are the same height

  \textbf{GT:} A
  
  \medskip
  \textbf{Model Responses:} 
I need to examine two images and compare the height of doors. In the first image, there's a brown door located at the bottom right corner of the gray building \hlg{labeled PIZZERIA PINOCCHIO}. In the second image, the building with a brown roof and gray walls seems to be the one with \hlg{a brown shingled roof and light gray walls}. Now, I’ll focus on comparing these details.
\hlg{Fig2's door seems taller}, though there could be scale considerations. But since the question asks for one, I'll go with Fig2.

  \textbf{Answer:} A
\end{minipage}
\par\vspace{0.6em}
\noindent\rule{\textwidth}{0.9pt}
\begin{minipage}[t]{0.2\linewidth}\vspace{0pt}
  \includegraphics[width=\linewidth,keepaspectratio]{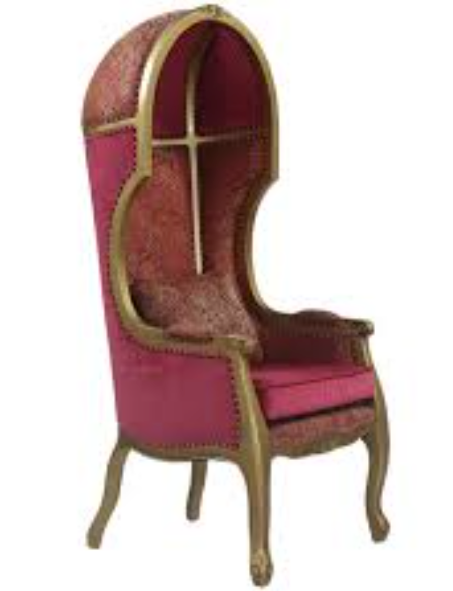}
\end{minipage}\hspace{1em}%
\begin{minipage}[t]{0.78\linewidth}\vspace{0pt}%
  \textbf{Mental Reconstruction (MR)}
  
  \medskip
  \textbf{Prompts:} generate a realistic front view of this chair.

  \textbf{GT:} -
  
  \medskip
  \textbf{Model Responses:} 
The user asked for a realistic front view, so I'll set the neutral gray studio background. The content needs to stay non-sensitive, and I need to follow the style guidelines—avoiding extra text since the tool output will be blank. I'll go ahead and generate the image without asking for further clarification unless specifically needed.

This isn't an editing task—it’s about generating the image from a scene description. Including the image ID is optional, given the tool doesn’t seem to require it. \hlg{I’ll focus on describing the chair features in detail to ensure it captures everything accurately in the generated image.}

  \textbf{Answer:} \includegraphics[width=0.2\linewidth,clip,trim=60 40 60 60]{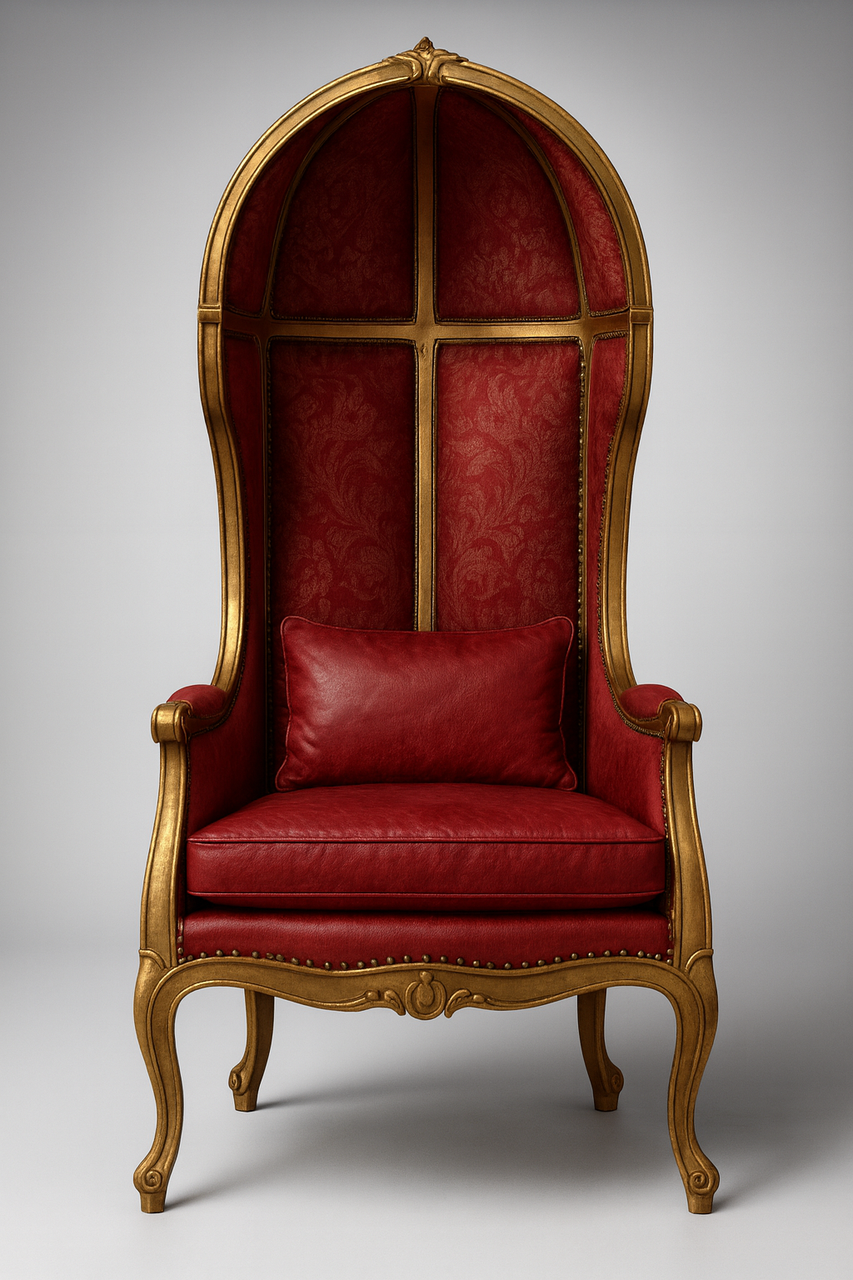}
\end{minipage}
\par\vspace{0.6em}
\noindent\rule{\textwidth}{0.9pt}
\noindent
\begin{minipage}[t]{0.2\linewidth}\vspace{0pt}
  \includegraphics[width=\linewidth,keepaspectratio]{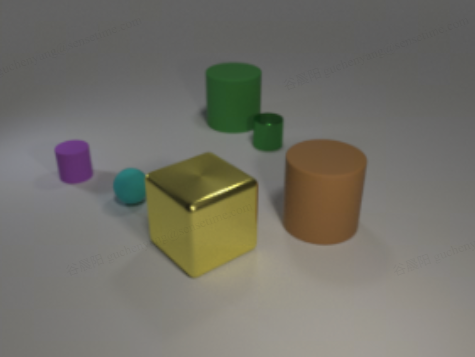}
\end{minipage}\hspace{1em}%
\begin{minipage}[t]{0.78\linewidth}\vspace{0pt}%
  \textbf{Mental Reconstruction (MR)}
  
  \medskip
  \textbf{Prompts:} Generate a 90 degrees top-down view of this scene.  

  \textbf{GT:} -
  \medskip

  \textbf{Model Responses:} 

I need to ensure the objects are positioned correctly in the top-view image. From the user’s image, there’s a gold cube slightly left of center, a brown cylinder to the right, and a big green cylinder at the back. There's also a small green cylinder in front of the big one. \hlg{I’ll carefully note the purple cylinder and a small cyan sphere near the left.}

I'll create a detailed prompt for the image: \hlg{a top-down view of a small tabletop scene.} It’ll include objects like a gold rounded cube, brown and green cylinders, and a teal sphere. I’ll specify soft studio lighting, a smooth gray floor, and realistic materials with ray traced reflections and shadows. I'll specify the \hlg{camera position directly above} for a top view. I'll also \hlg{detail the scale and relative positioning of the objects}, as seen in the image.

\textbf{Answer:} \includegraphics[width=0.4\linewidth,keepaspectratio]{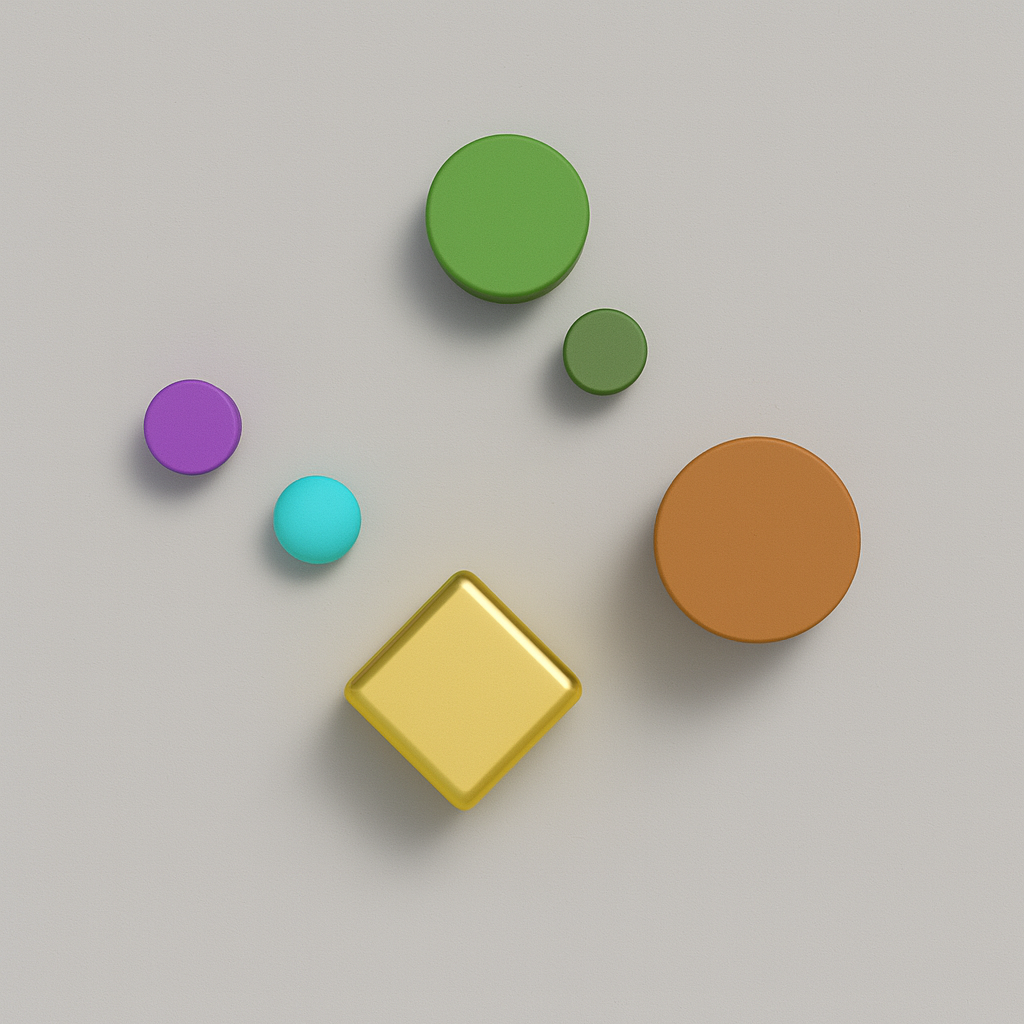}
\end{minipage}
\par\vspace{0.6em}
\noindent\rule{\textwidth}{0.9pt}
\begin{minipage}[t]{0.2\linewidth}\vspace{0pt}
  \includegraphics[width=\linewidth,keepaspectratio]{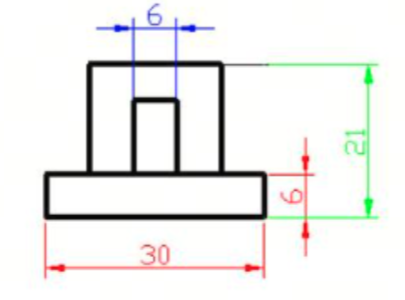}
  \includegraphics[width=0.7\linewidth,keepaspectratio]{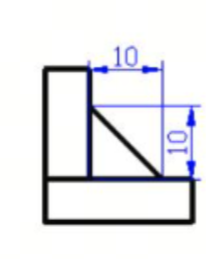}
  \includegraphics[width=\linewidth,keepaspectratio]{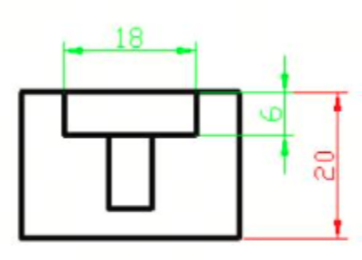}
\end{minipage}\hspace{1em}%
\begin{minipage}[t]{0.78\linewidth}\vspace{0pt}%
  \textbf{Mental Reconstruction (MR)}
  
  \medskip
  \textbf{Prompts:} Given the front, side and top-down view of a 3D object, analyze its structure and reconstruct it in 3D axis.

  \textbf{GT:} -

  \medskip
  \textbf{Model Responses:} 
I’m analyzing the first image: the front view shows a base with \hlg{a width of 30 and a thickness of 6. The 21 indicates the total height} from the base to the top. There’s also a 6 indicating base thickness. Above the base is a U-shaped bracket with side walls of 15 height, and a central inner pillar that’s smaller but with a horizontal dimension of 6. \hlg{I'll need to check the top view for clarification}.

In the top view, I see that the width is 18, and there’s a recess near the top. The depth is 20, shown in red, and the thickness of the recessed shelf is 6 vertically. The shape resembles a "T"-slot, which aligns with the front view’s central pillar. \hlg{The overall dimensions match the front view, where width is likely 30 and depth is 20.}

The object seems \hlg{symmetrical with a base of 30x20x6.} On top, there's \hlr{a rectangular frame in a U-shape, surrounding a central pillar. The side view shows a triangular gusset with a 45-degree angle. This results in varying heights across different views}, but overall, it fits a block with a central T-shaped ridge and sloped sides.

\hlg{The structure should include a base with dimensions 30x20x6, a crossbar (18x6x15), and a central pillar (6x?x10).} The front view indicates a horizontal line at 21 along the width 18, with a lower 16 at the central region. \hlr{The side ramp slopes from 21 to 6 based on the defined measurements.}

The ramp is a triangular prism with a base at \hlg{z=6 (y=4) and sloping up to z=16 at y=14, with a width of 6 units.} The structure, including the base, ramp, and crossbar, aligns with the top view’s 'T' shape and the front view’s height relationships.

  \textbf{Answer:} \includegraphics[width=0.2\linewidth,clip]{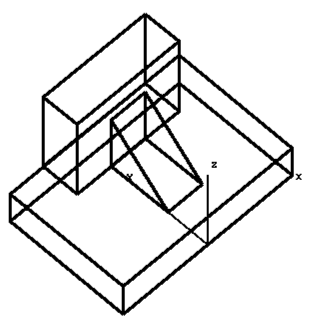}
\end{minipage}
\par\vspace{0.6em}
\noindent\rule{\textwidth}{0.9pt}
\begin{minipage}[t]{0.2\linewidth}\vspace{0pt}
  \includegraphics[width=\linewidth,keepaspectratio]{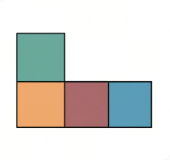}
\end{minipage}\hspace{1em}%
\begin{minipage}[t]{0.78\linewidth}\vspace{0pt}%
  \textbf{Mental Reconstruction (MR)}
  
  \medskip
  \textbf{Prompts:} Draw a 3D shape whose top view looks like this, and composed of 3D cubes.

  \textbf{GT:} -
  
  \medskip

  \textbf{Model Responses:} 
The user requested a 3D object with a top view resembling an L-tetromino, made out of cubes. It’s a 4-square footprint, with varied possible heights. \hlr{The simplest format is an isometric view with a single cube stacked.} I'll create the requested 3D shape and generate the image without needing extra edits or references.

  \textbf{Answer:}\raisebox{-0.4\height}{\includegraphics[height=6em]{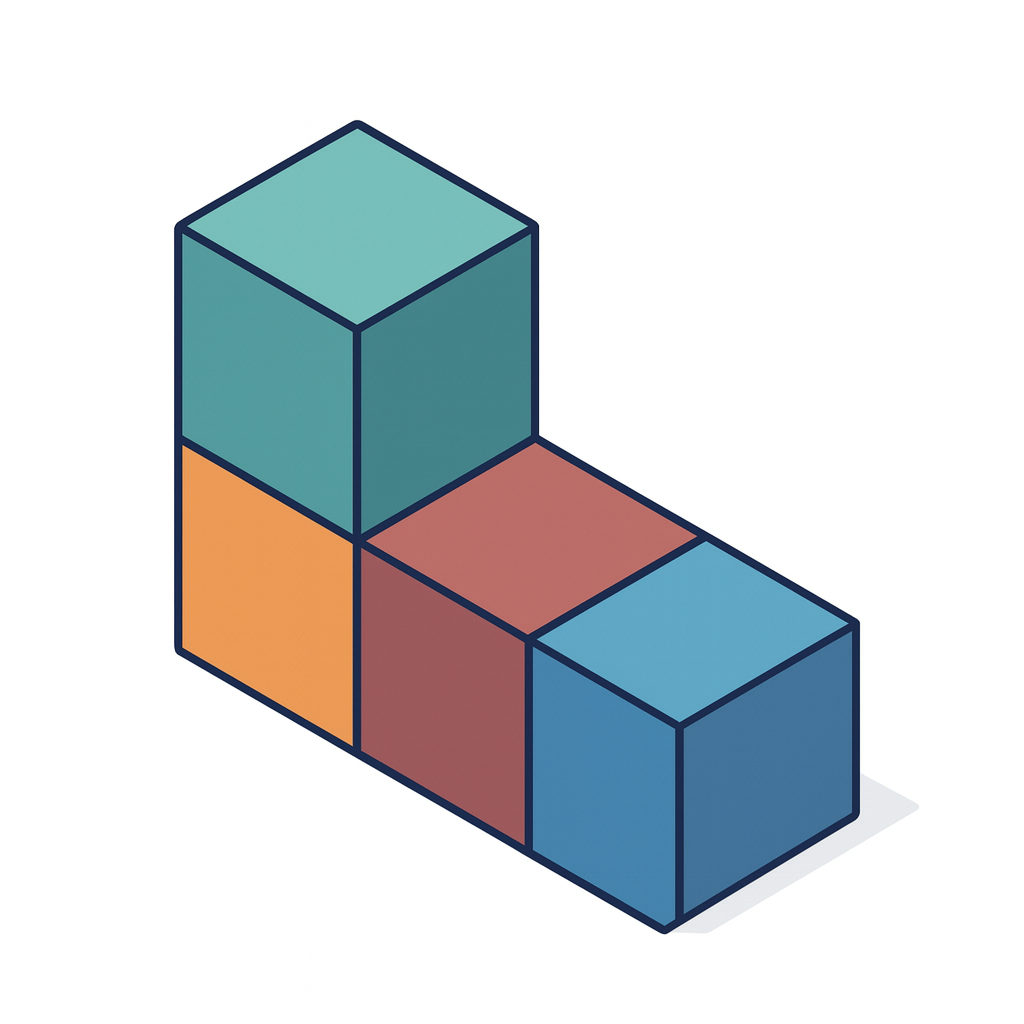}}
\end{minipage}
\par\vspace{0.6em}
\noindent\rule{\textwidth}{0.9pt}
\begin{minipage}[t]{0.2\linewidth}\vspace{0pt}
  \includegraphics[width=\linewidth,keepaspectratio]{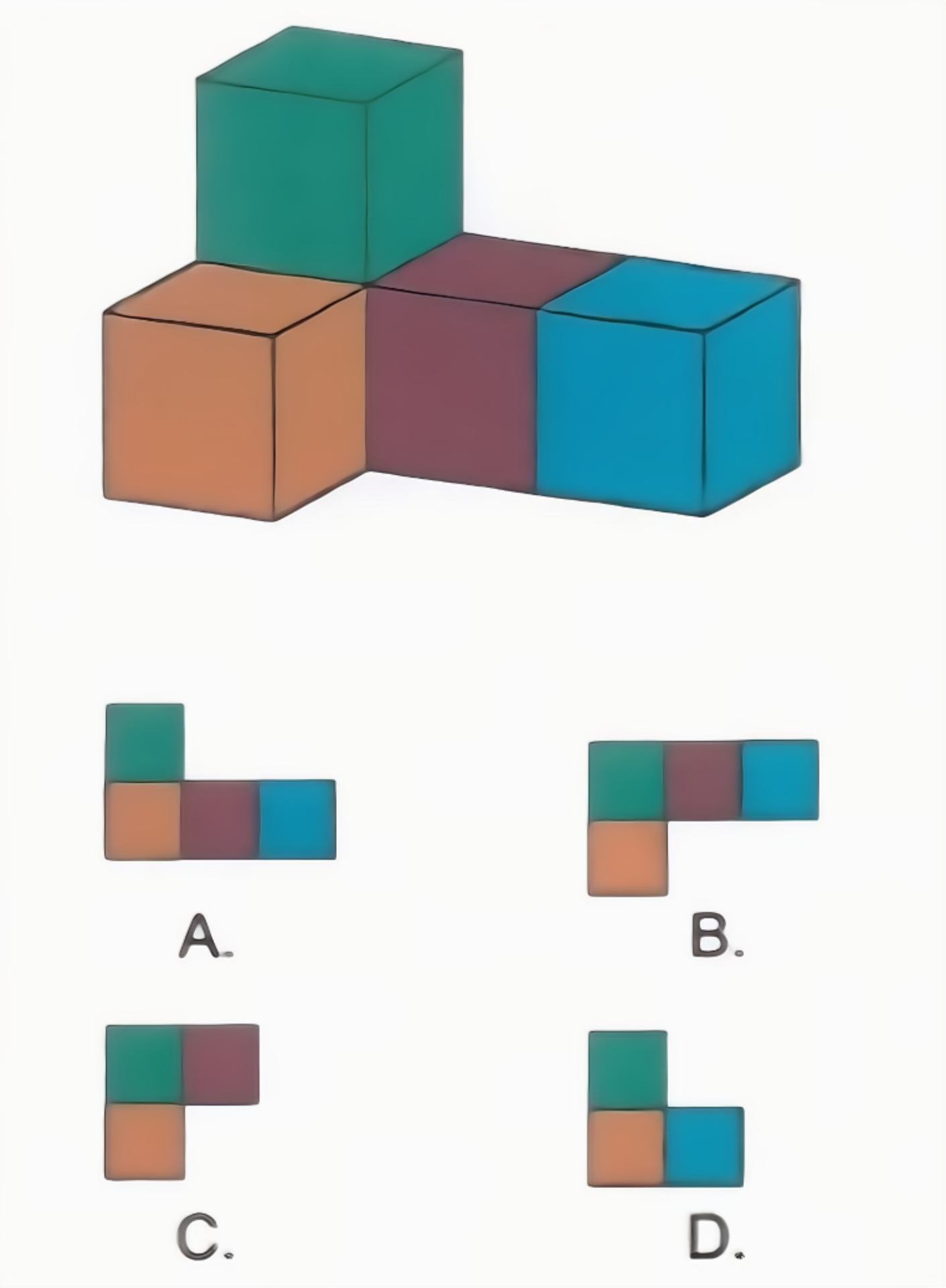}
\end{minipage}\hspace{1em}%
\begin{minipage}[t]{0.78\linewidth}\vspace{0pt}%
  \textbf{Mental Reconstruction (MR)}
  
  \medskip
  \textbf{Prompts:} which option is the correct top-down view of the above object?

  \textbf{GT:} B

  \medskip
  \textbf{Model Responses:} 
\raisebox{-0.2\height}{\includegraphics[height=3.0em]{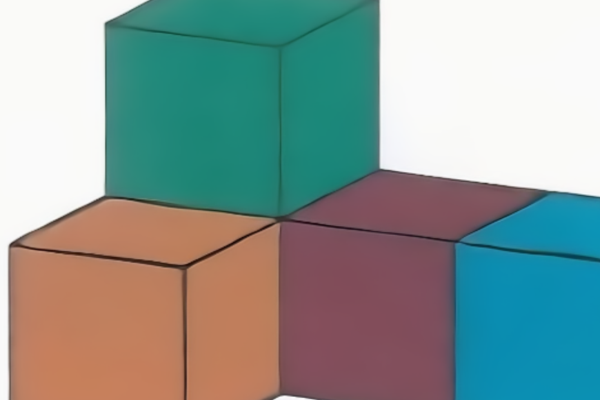}} 
\raisebox{-0.2\height}{\includegraphics[height=3.0em]{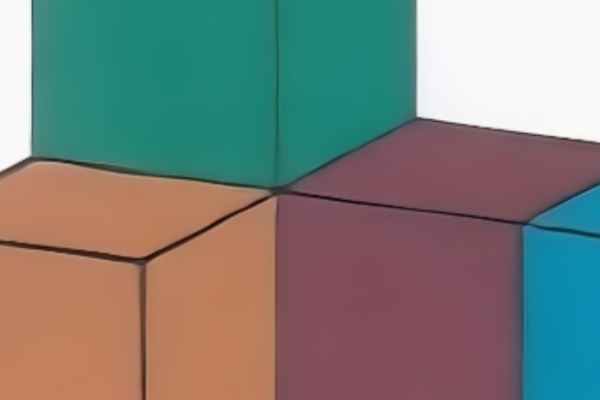}} 
Looking at the 3D arrangement of cubes, it seems to form \hlg{an L-shape on the bottom plane} — with 4 cubes. The leftmost front cube is orange, the middle one is purple, and the rightmost is cyan. \hlr{The green cube seems to sit above the purple, slightly back and left from it.}

Option A shows an L shape with four squares: green above orange, followed by orange, purple, and cyan in a horizontal line. Option B also forms an L shape, but with green at the top-left corner. Option C only has three squares (green, purple, and orange). Option D shows three squares as well, with green placed above orange in the left column, and cyan at the bottom right.

Only Option A and B show four squares. The 3D structure seems to have \hlg{four visible cubes}, plus one stacked on top. The green cube appears stacked above, while the base includes three visible cubes: orange, purple, and cyan. It's possible there's a hidden base cube under green, but this depends on visibility.

The bottom row in Option B suggests that the cubes are placed in a straight row: orange, purple, and cyan. For the green cube, its position above the leftmost cube (likely orange) is questioned by the visibility of each cube's top face. \hlr{The green cube appears to sit behind orange and purple at a junction, but wouldn't fully obstruct orange's top face.}

The green cube isn't stacked on any of the base cubes since we can see the tops of purple and orange. \hlr{The green is positioned behind these cubes, elevated, making it appear higher due to perspective. It likely sits on the ground, aligning with the top row's cubes, but it's behind them geometrically.}

Option A correctly shows green above orange. Option B, however, places purple on top next to green, which seems like a misplacement when compared to the first.

\textbf{Answer:}\par
A — three cubes in a row (orange–purple–blue) with one green cube behind the orange.
\end{minipage}
\par\vspace{0.6em}
\noindent\rule{\textwidth}{0.9pt}
\begin{minipage}[t]{0.2\linewidth}\vspace{0pt}
  \includegraphics[width=0.9\linewidth,keepaspectratio]{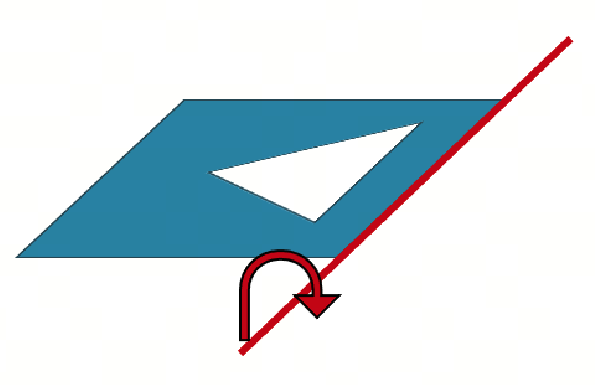}
  \includegraphics[width=0.9\linewidth,keepaspectratio]{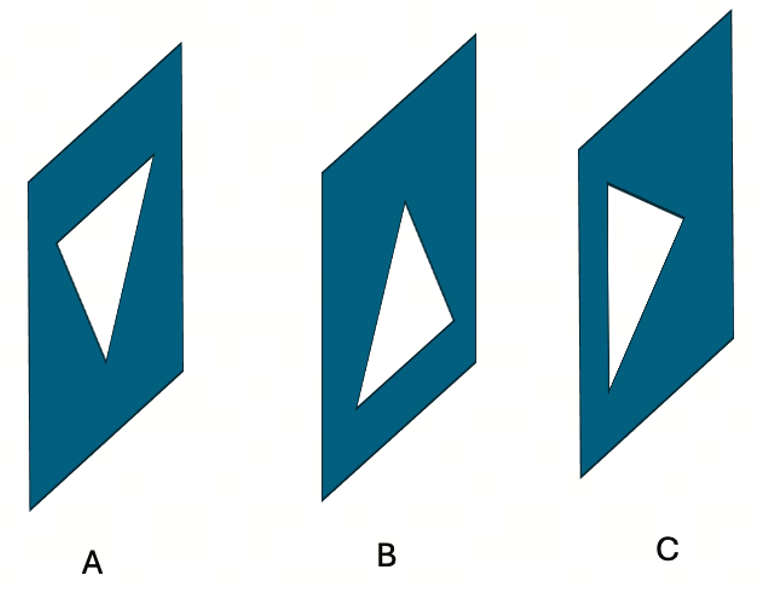}
\end{minipage}\hspace{1em}%
\begin{minipage}[t]{0.78\linewidth}\vspace{0pt}%
  \textbf{Mental Reconstruction (MR)}
  
  \medskip
  \textbf{Prompts:} Rotate the leftmost image along the red line by the direction of the arrow. if rotate 90 degrees, which of A B C will it be like? Consider the shape as a thin layer in 3D space, and the rotation is in 3D space.

  \textbf{GT:} B
  
  \medskip
  \textbf{Model Responses:} 
The red line is along the plate's lower-right corner,\hlg{with an arrow indicating a clockwise rotation.} This suggests \hlr{the left side of the plate moves downward, away from the viewer.} \hlg{Initially flat, the plate rotates to stand vertically}, with the triangular hole’s orientation adjusting accordingly. Now, I'll focus on matching it with the options.

\hlg{The correct option is the one where the triangle is not mirrored across the pivot edge.}

After rotation, \hlr{the pivot edge becomes the front vertical edge}, but the triangle’s orientation relative to the pivot stays the same. 

In option C, the triangular base is vertical, and its apex points toward the right, which seems like a mirror of the original triangle's layout. The original apex pointed up and to the right, while in C, it points similarly to the right edge. Given the pivot seems to align with the right edge in vertical configurations, C appears plausible.

  \textbf{Answer:} C
\end{minipage}
\par\vspace{0.6em}
\noindent\rule{\textwidth}{0.9pt}
\begin{minipage}[t]{0.2\linewidth}\vspace{0pt}
    \includegraphics[width=0.7\linewidth,keepaspectratio]{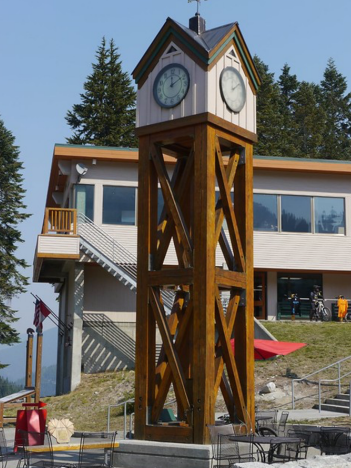}
\end{minipage}\hspace{1em}%
\begin{minipage}[t]{0.78\linewidth}\vspace{0pt}%
  \textbf{Spatial Relations (SR)}
  
  \medskip
  \textbf{Prompts:} 
Which object is higher in the 3D world space, the clock or the house in the back?

  \textbf{GT:} The house in the back.
  
  \medskip
  \textbf{Model Responses:} 
The question is asking which object is higher in physical elevation, the clock tower or the house behind it, not just perceived height. The clock tower seems taller and \hlr{extends above the house roof}, even factoring in the potential slope of the ground. So, the clock tower is likely the higher object in 3D world space.

  \textbf{Answer:} The clock.
\end{minipage}
\par\vspace{0.6em}
\noindent\rule{\textwidth}{0.9pt}
\begin{minipage}[t]{0.2\linewidth}\vspace{0pt}
    \includegraphics[width=0.9\linewidth,keepaspectratio]{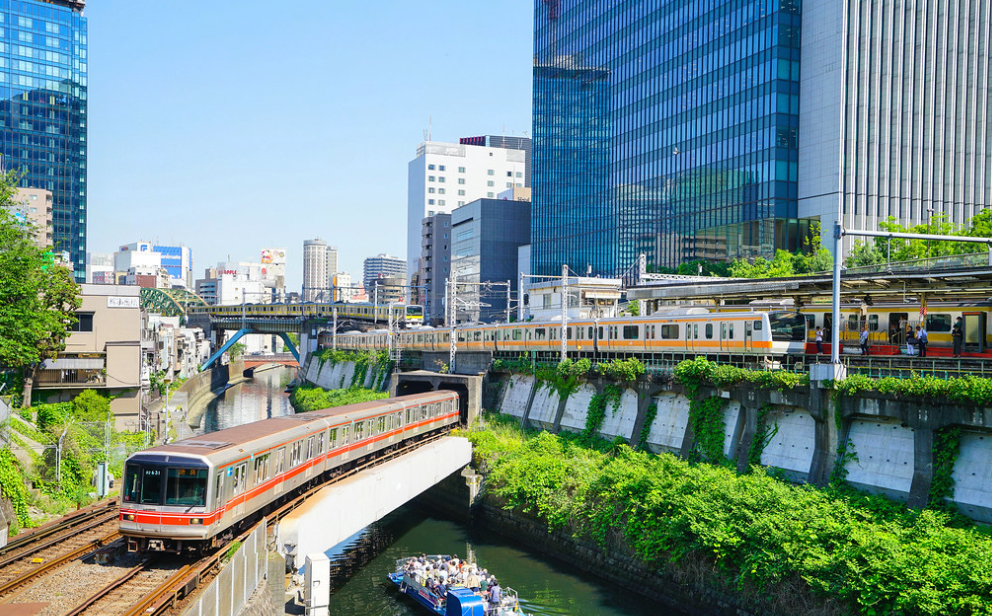}
\end{minipage}\hspace{1em}%
\begin{minipage}[t]{0.78\linewidth}\vspace{0pt}%
  \textbf{Spatial Relations (SR)}
  
  \medskip
  \textbf{Prompts:} 
Consider the real-world 3D locations and orientations. Is the blue bridge parallel or perpendicular to the train with orange stripes?

  \textbf{GT:} Perpendicular.
  
  \medskip
  \textbf{Model Responses:} 
In considering the 3D world orientation, not the 2D perspective, the blue bridge crosses above the canal perpendicular to the train, which runs along the canal. While the bridge may appear diagonal in the image, its orientation is most likely perpendicular to the train. There might be some nuance, but overall, it's best categorized as perpendicular.

The blue bridge, which crosses the canal, has a triangular support that runs perpendicular to the orange-striped train track. \hlg{Though the bridge may appear to cross the track at an angle, it's still generally perpendicular. So, the final answer is that the blue bridge is perpendicular to the train.}

Initially, I thought the blue bridge crossed the canal, and the orange train ran along it, but it turns out the orange train also crosses the canal in the same direction. Hmm, \hlr{the train at the station may be oriented along the canal, but there seems to be some visual illusion. }

Looking at the full scene, \hlr{the orange-striped train does indeed curve from right to left, crossing the canal below the yellow train on the blue bridge. The two tracks are not perpendicular, but rather nearly parallel, both aligned left to right.} The blue bridge deck slopes slightly, but overall, they run along the same direction.

  \textbf{Answer:} Parallel.
\end{minipage}
\par\vspace{0.6em}
\noindent\rule{\textwidth}{0.9pt}
\begin{minipage}[t]{0.2\linewidth}\vspace{0pt}
\raisebox{-0.2\height}{\includegraphics[height=8.0em]{assets/case_studies/multiviewQ1_1.png}} 
\raisebox{-0.2\height}{\includegraphics[height=8.0em]{assets/case_studies/multiviewQ1_2.png}} 
\end{minipage}\hspace{1em}%
\begin{minipage}[t]{0.78\linewidth}\vspace{0pt}%
  \textbf{Perspective-taking (PT)}
  
  \medskip
  \textbf{Prompts:} The images are frames from a video. The first image is from the beginning of the video and the second image is from the end. Is the camera moving left or right when shooting the video?

  \textbf{GT:} Left.
  
  \medskip
  \textbf{Model Responses:} 
OK, so the camera seems to be moving around the bottle. In the first image, we see more of the label's \hlr{left side} and a small circle "72". In the second image, the front label
\begin{CJK}{UTF8}{gbsn}
("每益添")
\end{CJK}
faces more towards us. \hlr{This suggests the camera moved clockwise to the right, revealing more of the bottle's front.} Both shots show the same stack of boxes, but the perspective shifts.
\hlr{If the camera is moving right, the background would shift left relative to the bottle.} I’m comparing the handle slot's position relative to the bottle in each image. The first image places the handle on the right, while the second places it on the left, suggesting the camera moved. 

  \textbf{Answer:} Right.
\end{minipage}
\par\vspace{0.6em}
\noindent\rule{\textwidth}{0.9pt}

\begin{minipage}[t]{0.36\linewidth}\vspace{0pt}
\raisebox{-0.2\height}{\includegraphics[height=8.0em]{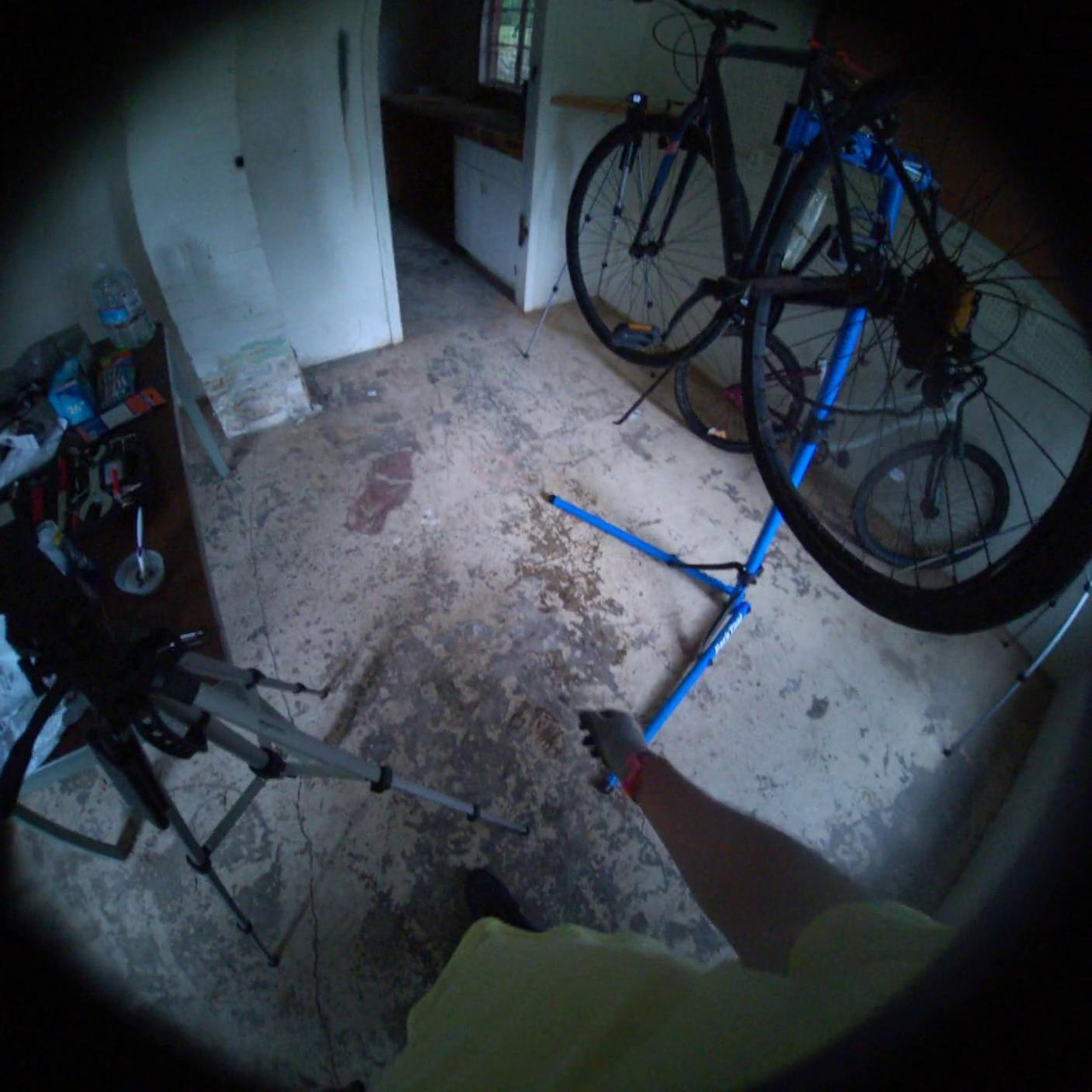}}
\raisebox{-0.2\height}{\includegraphics[height=8.0em]{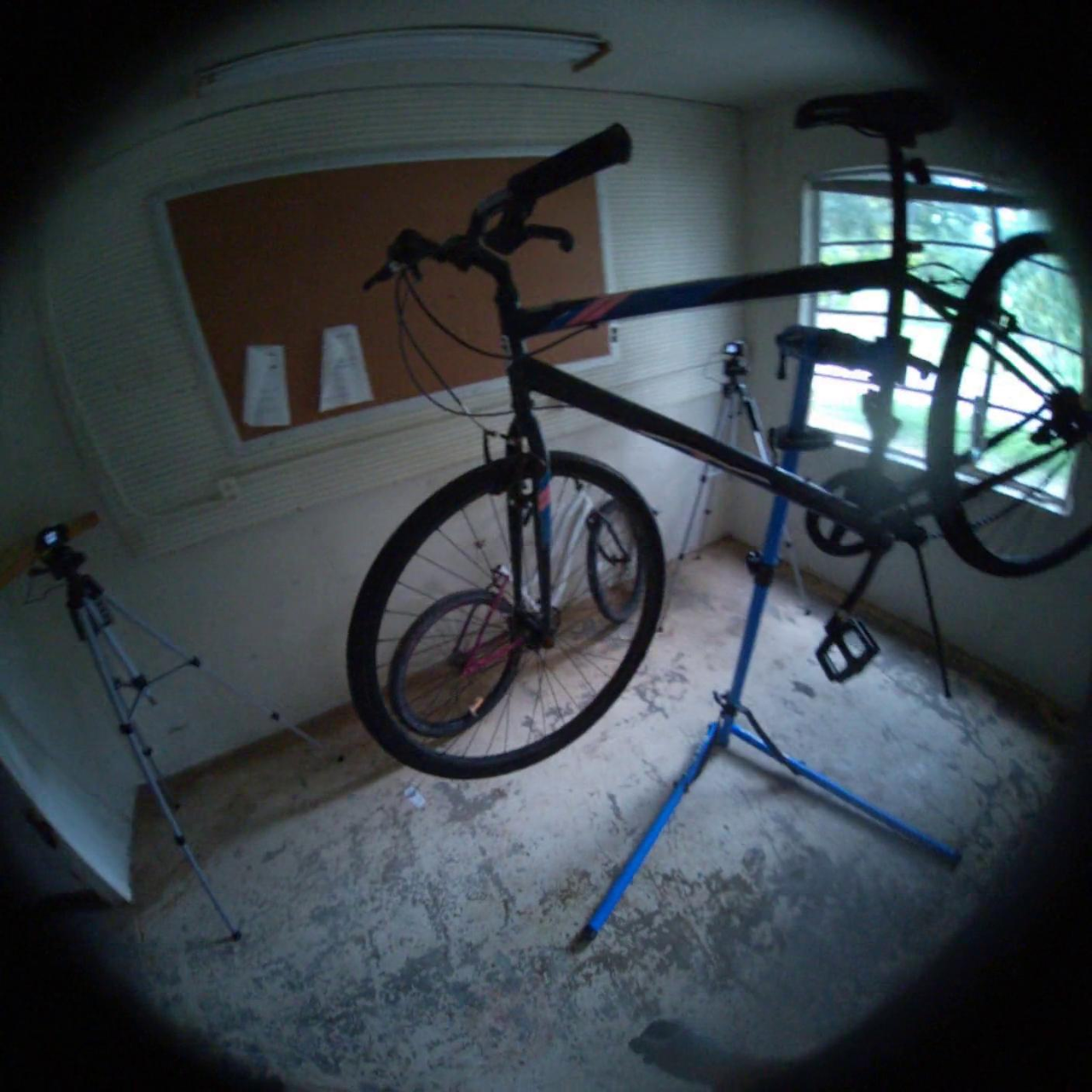}} 
\raisebox{-0.2\height}{\includegraphics[height=7.0em]{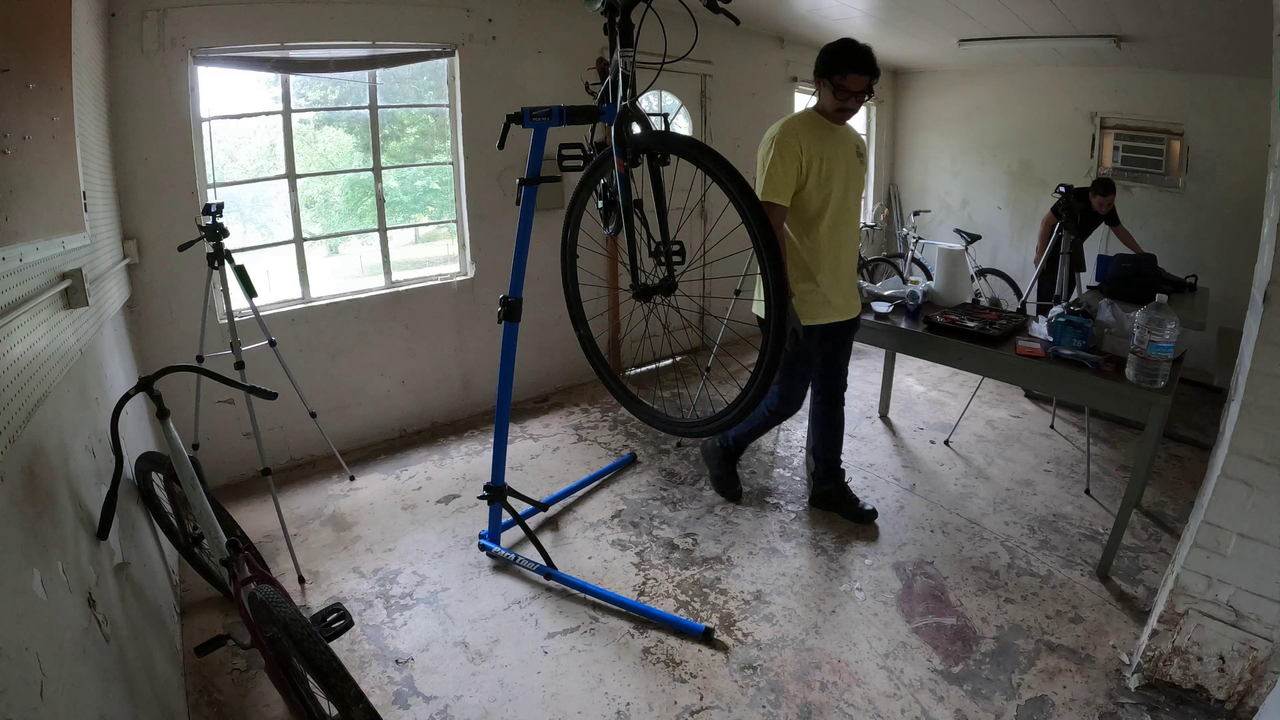}} 
\raisebox{-0.2\height}{\includegraphics[height=7.0em]{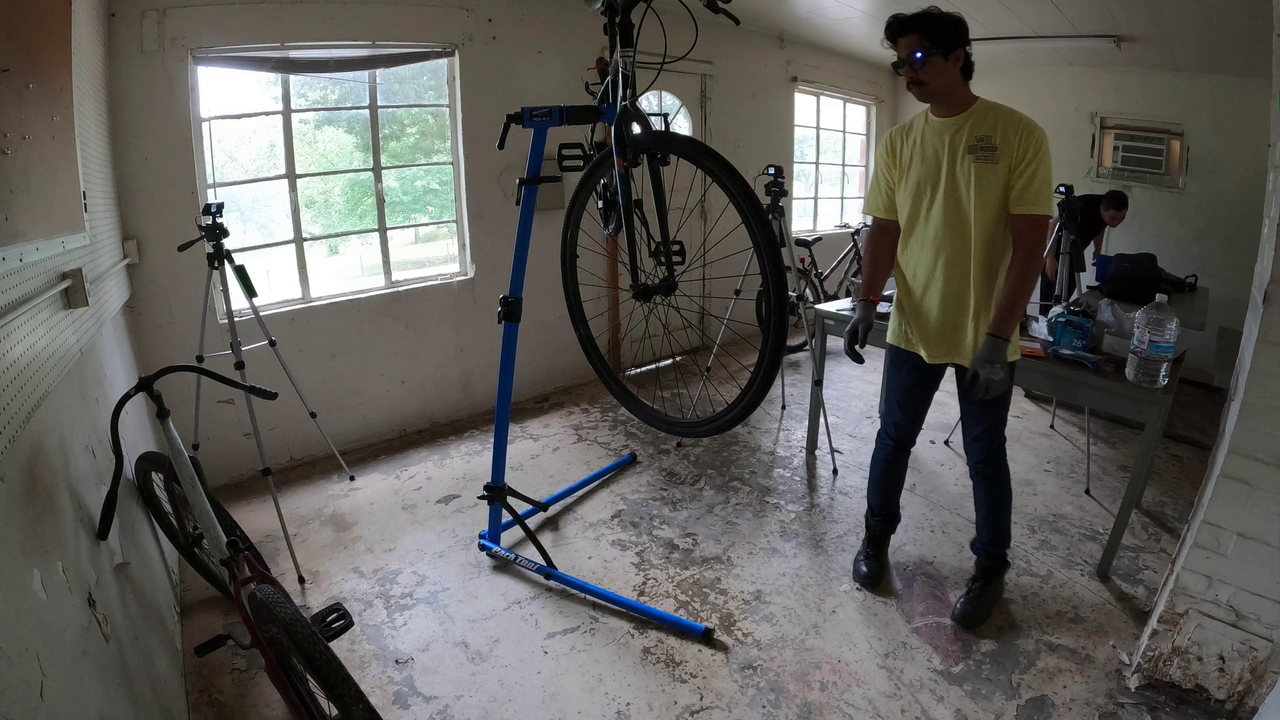}} 
\raisebox{-0.2\height}{\includegraphics[height=7.0em]{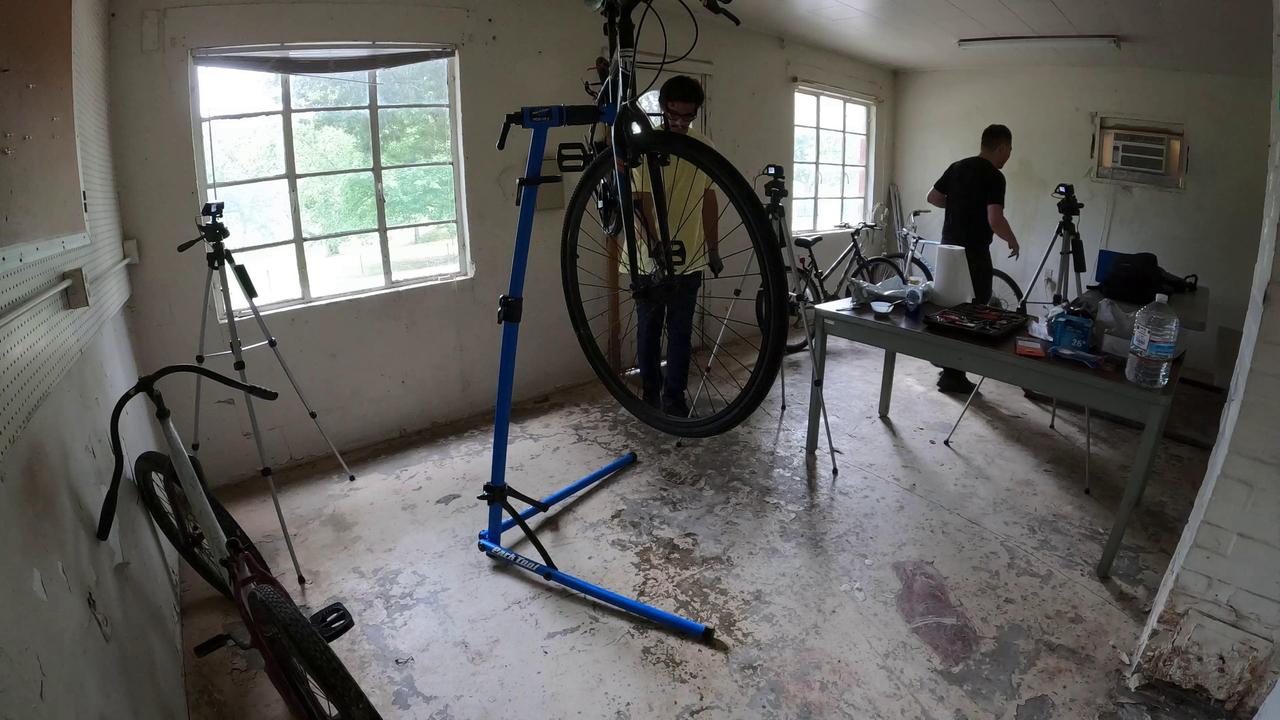}} 
\raisebox{-0.2\height}{\includegraphics[height=7.0em]{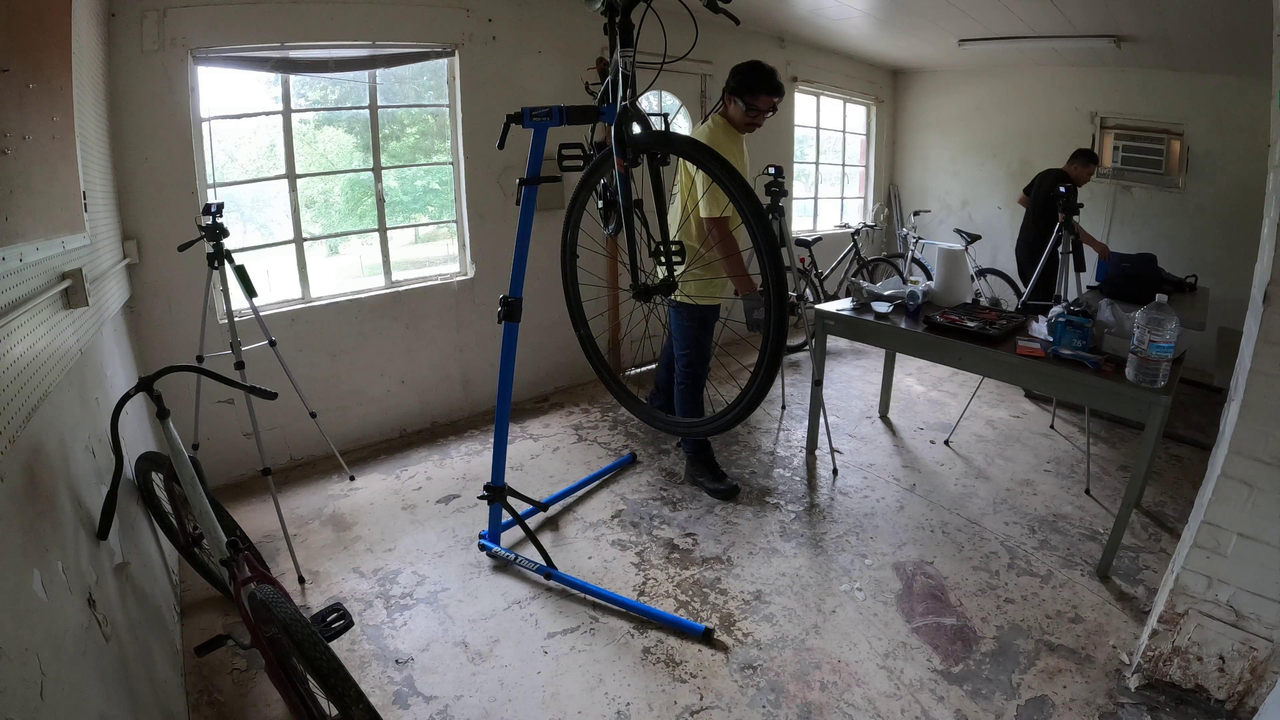}} 
\end{minipage}\hspace{1em}%
\begin{minipage}[t]{0.58\linewidth}\vspace{0pt}%
  \textbf{Perspective-taking (PT)}
  
  \medskip
  \textbf{Prompts:} You are given two first-person perspective images, representing the start and end frame of a video clip: <image><image>. And then there are four third-person perspective frames, each captured during the video clip but in random temporal order: \textless image \textgreater\textless image \textgreater\textless image\textgreater\textless image\textgreater. Analyze the images carefully based on scene details, lighting, and object positions, and determine which option gives the correct temporal order of these 4 third-person perspective frames.

  \textbf{GT:} 3 → 4 → 2 → 1.
  \medskip
  
  \textbf{Model Responses:} 
The first-person start frame shows the man reaching near the blue stand, possibly indicating his location. The end frame shows him facing the window with the bike still on the stand. The first-person frames help establish the timeline, implying that the third-person frames should be ordered based on his movement from entering the frame to adjusting the bike. 

He \hlr{starts on the far right of the room, turning toward the bike stand (frame 2), then walks left past the front wheel (frame 1)}. Next he stops at the stand and reaches down to adjust it (frame 4), and finally he’s positioned behind the front wheel inspecting/working on the bike (frame 3).
  
\textbf{Answer:} \par
2 → 1 → 4 → 3.
\end{minipage}
\par\vspace{0.6em}
\noindent\rule{\textwidth}{0.9pt}
\begin{minipage}[t]{0.36\linewidth}\vspace{0pt}
\includegraphics[width=.48\linewidth]{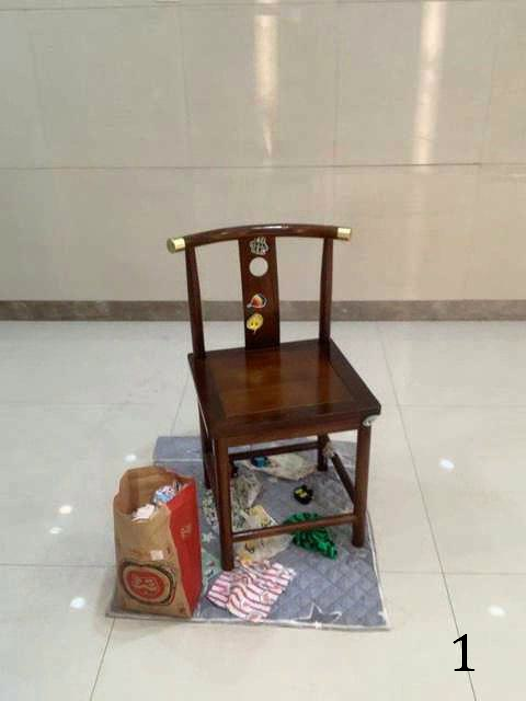}%
\includegraphics[width=.48\linewidth]{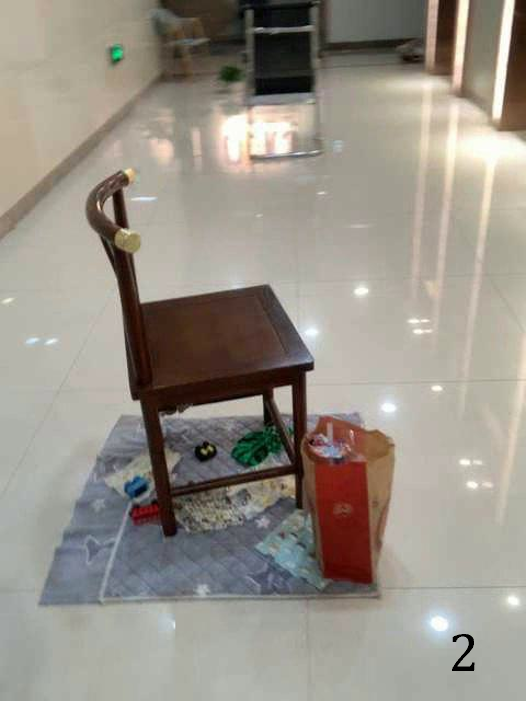}
\includegraphics[width=.48\linewidth]{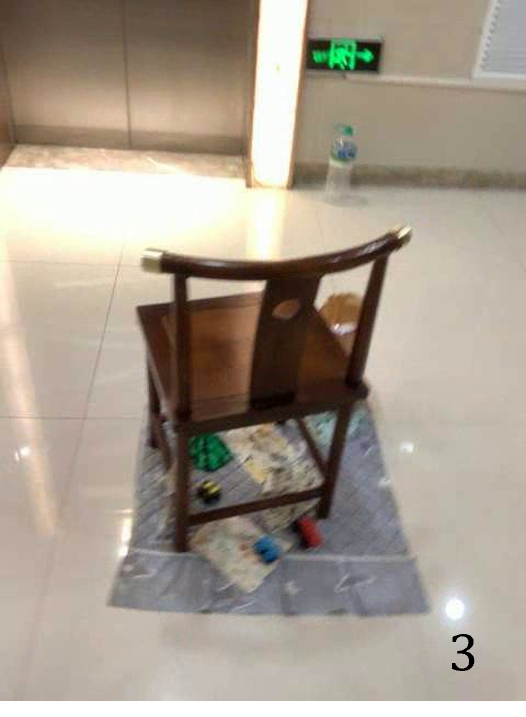}%
\includegraphics[width=.48\linewidth]{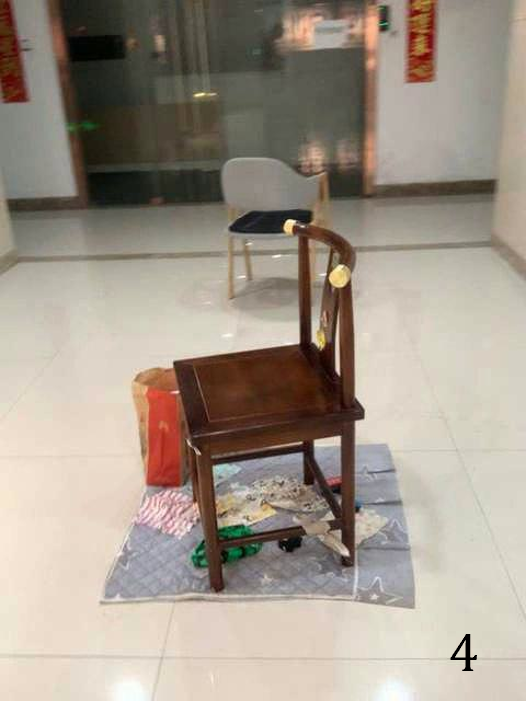}
\end{minipage}\hspace{1em}%
\begin{minipage}[t]{0.58\linewidth}\vspace{0pt}%
  \textbf{Perspective-taking (PT)}
  
  \medskip
  \textbf{Prompts:} Based on these four images (image 1, 2, 3, and 4) showing the red wooden chair from different viewpoints (front, left, back, and right), with each camera aligned with room walls and partially capturing the surroundings: If I am standing at the same spot and facing the same direction as shown in image 3, then I turn left and move forward, will I get closer to the corridor and black chair? 

  \textbf{GT:} Yes.
  
  \medskip
  
  \textbf{Model Responses:} 
\hlr{From image 3 (back view), turning left puts you facing the orientation of image 4 (right-side view)}. The corridor with the black chair is in image 2 (left-side view). So after turning left you’d head the opposite way, not toward the corridor/black chair.
  
  \textbf{Answer:} No.
\end{minipage}
\par\vspace{0.6em}
\noindent\rule{\textwidth}{0.9pt}
\noindent
\begin{minipage}[t]{0.2\linewidth}\vspace{0pt}
    \centering
  \includegraphics[width=0.8\linewidth,keepaspectratio]{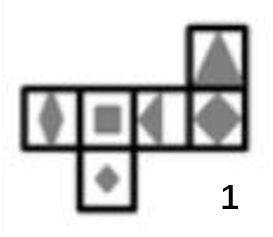}
  \includegraphics[width=0.4\linewidth,keepaspectratio]{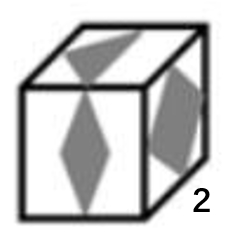}
  \includegraphics[width=0.4\linewidth,keepaspectratio]{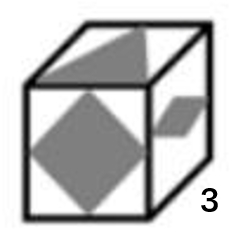}
  \includegraphics[width=0.4\linewidth,keepaspectratio]{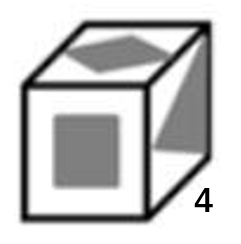}
  \includegraphics[width=0.4\linewidth,keepaspectratio]{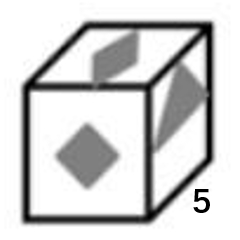}
\end{minipage}\hspace{1em}%
\begin{minipage}[t]{0.78\linewidth}\vspace{0pt}%
  \textbf{Deformation and Assembly (DA)}
  
  \medskip
  \textbf{Prompts:} Flip the shape in image 1 to form a 3D cube. Which of the image 2, 3, 4, 5 is a possible view of the formed cube?

  \textbf{GT:} Image 4.

  \medskip
  \textbf{Model Responses:} 
Reason: From the net, opposite faces are: triangle (top) v.s. small diamond (bottom),
big diamond v.s. small square, long diamond v.s. the split-triangle face.
Any valid cube view must show three faces that meet at one corner (no opposites touching).

Image 3 shows triangle next to the \hlr{small diamond (opposites)} → impossible.

Image 4 puts the small square next to the \hlr{big diamond (opposites)} → impossible.

Image 5 has the long diamond on top even though \hlr{the top is the triangle} → impossible.

Only Image 2 has \hlr{three mutually adjacent faces in a configuration consistent} with the net.

  \textbf{Answer:} Image 2.
\end{minipage}
\par\vspace{0.6em}
\noindent\rule{\textwidth}{0.9pt}
\begin{minipage}[t]{0.2\linewidth}\vspace{0pt}
  \includegraphics[width=0.4\linewidth,keepaspectratio]{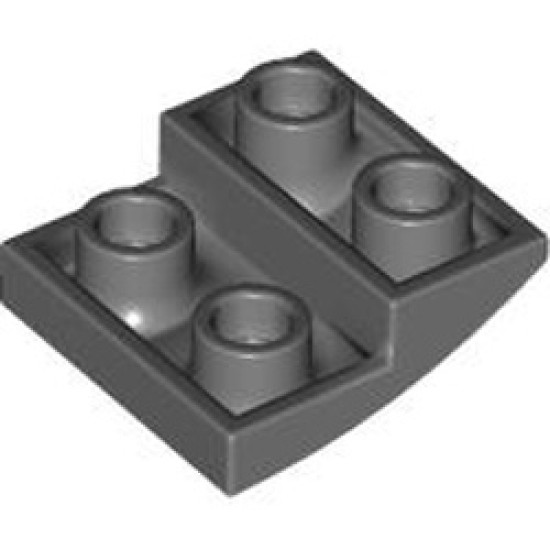}
  \includegraphics[width=0.4\linewidth,keepaspectratio]{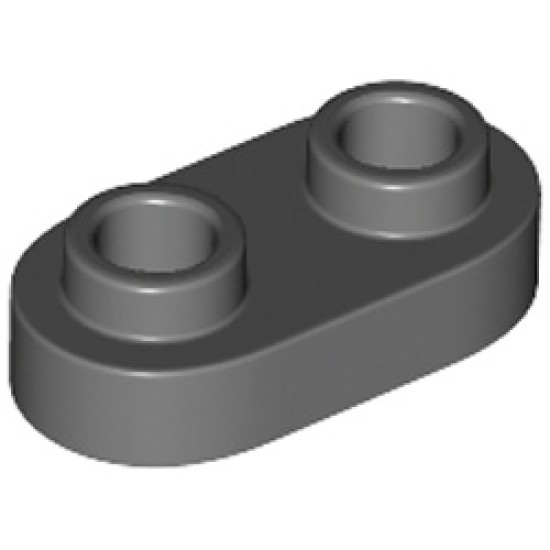}
  \includegraphics[width=0.8\linewidth,keepaspectratio]{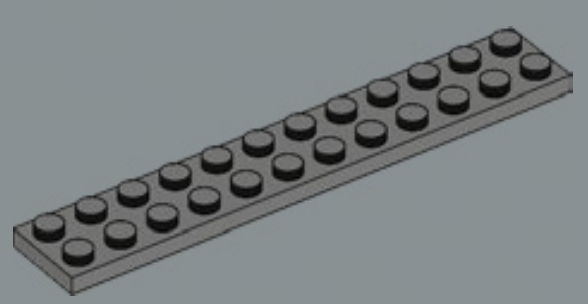}
  \includegraphics[width=0.8\linewidth,keepaspectratio]{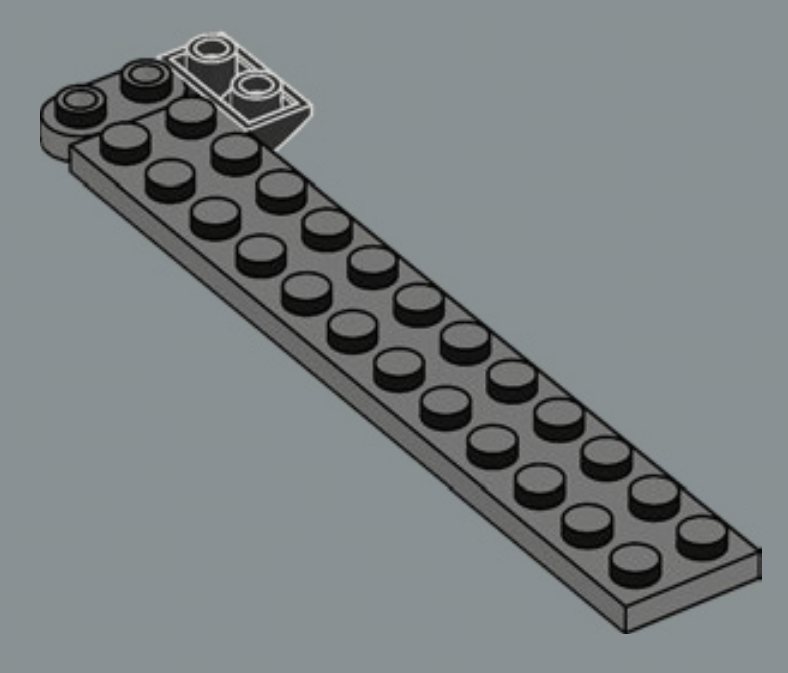}
\end{minipage}\hspace{1em}%
\begin{minipage}[t]{0.78\linewidth}\vspace{0pt}%
  \textbf{Deformation and Assembly (DA)}
  
  \medskip
  \textbf{Prompts:} How to use the first three to build the last structure? show a draft image.

  \textbf{GT:} -

  \medskip
  \textbf{Model Responses:} 
Here’s a quick draft showing how to place the pieces: the long 2×12 plate is the base, the rounded 1×2 caps the far-left end, and the \hlr{2×2 corner/step sits directly to its right.}

  \textbf{Answer:} \includegraphics[width=0.4\linewidth,keepaspectratio]{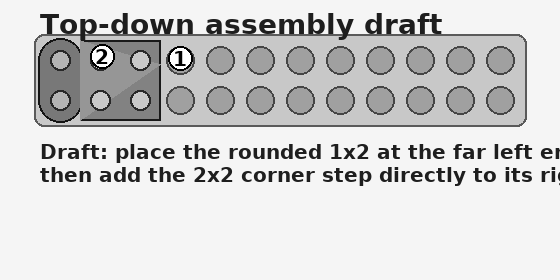}
\end{minipage}
\par\vspace{0.6em}
\noindent\rule{\textwidth}{0.9pt}

\begin{minipage}[t]{0.2\linewidth}\vspace{0pt}
  \includegraphics[width=\linewidth,keepaspectratio]{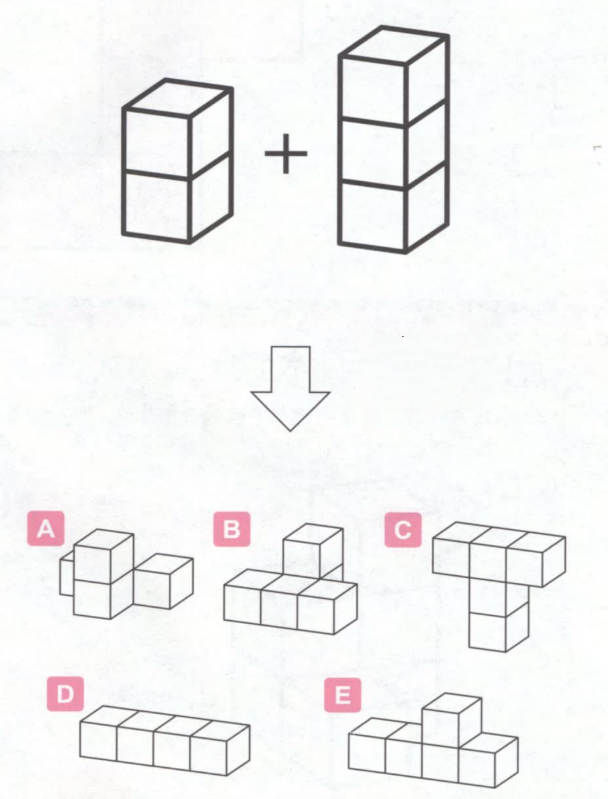}
\end{minipage}\hspace{1em}%
\begin{minipage}[t]{0.78\linewidth}\vspace{0pt}%
  \textbf{Deformation and Assembly (DA)}
  
  \medskip
  \textbf{Prompts:} For each of A to E, analyze if it can be built by stacking the two structures in the first image.

  \textbf{GT:} B, C

  \medskip
  \textbf{Model Responses:} 

We have 2 cubes + 3 cubes → total of 5 cubes must be used.

A — No. \hlg{Only 4 cubes shown.}

B — Yes. \hlg{Exactly 5 cubes.}

C — No. \hlr{Shows 6 cubes} (a row of 4 with a column of 3 sharing one).

D — No. \hlg{Only 4 cubes.}

E — Yes. \hlr{Exactly 5 cubes.}

\textbf{Answer:}\par
E
\end{minipage}
\par\vspace{0.6em}
\noindent\rule{\textwidth}{0.9pt}
\begin{minipage}[t]{0.2\linewidth}\vspace{0pt}
    \centering
  \includegraphics[width=0.6\linewidth,keepaspectratio]{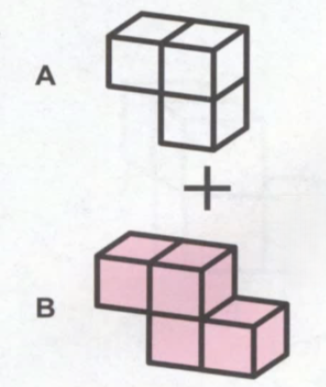}
  \includegraphics[width=0.8\linewidth,keepaspectratio]{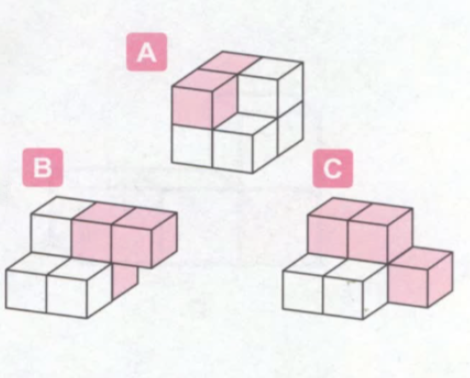}
\end{minipage}\hspace{1em}%
\begin{minipage}[t]{0.78\linewidth}\vspace{0pt}%
  \textbf{Deformation and Assembly (DA)}
  
  \medskip
  \textbf{Prompts:} Which of A B C is possible to be built when rotating and combining the two 3D structure in Image 1?

  \textbf{GT:} C
  
  \medskip
  \textbf{Model Responses:} 
A and B are possible; C is not.
Reason: \hlr{Each piece is an L-shaped tri-cube}. 

In options A and B, the pink L fits and the remaining white cubes form the second L.
In option C, after placing the pink L, the remaining white cubes can’t make another L (you’re left with only two adjacent whites with no spot for the third), so that shape can’t be built.

  \textbf{Answer:} A and B.
\end{minipage}
\par\vspace{0.6em}
\noindent\rule{\textwidth}{0.9pt}
\begin{minipage}[t]{0.2\linewidth}\vspace{0pt}
  \includegraphics[width=\linewidth,keepaspectratio]{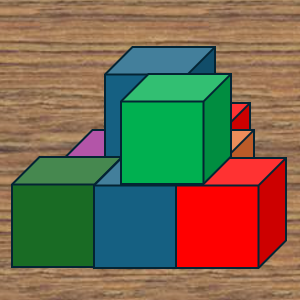}
\end{minipage}\hspace{1em}%
\begin{minipage}[t]{0.78\linewidth}\vspace{0pt}%
  \textbf{Comprehensive Reasoning (CR)}
  
  \medskip
  \textbf{Prompts:} How many 3D blocks in the image?

  \textbf{GT:} 8
  \medskip
  
  \textbf{Model Responses:} 

8 visible cubes, with some potential hidden ones for support. I estimate there are at least 10 cubes.

After double-checking, I’m sticking with 9 cubes. Bottom row has 3 (green, blue, red); back row shows 1 left (purple) and 2 on the right (orange, red); \hlr{there’s a tall blue stack behind the center that’s 2 cubes high (only the top is visible, one is hidden); plus the green cube on top = 3 + 3 + 2 + 1 = 9}

The front faces match too, so I think this count works.

\textbf{Answer:} 9
\end{minipage}
\par\vspace{0.6em}
\noindent\rule{\textwidth}{0.9pt}
\begin{minipage}[t]{0.2\linewidth}\vspace{0pt}
  \includegraphics[width=\linewidth,keepaspectratio]{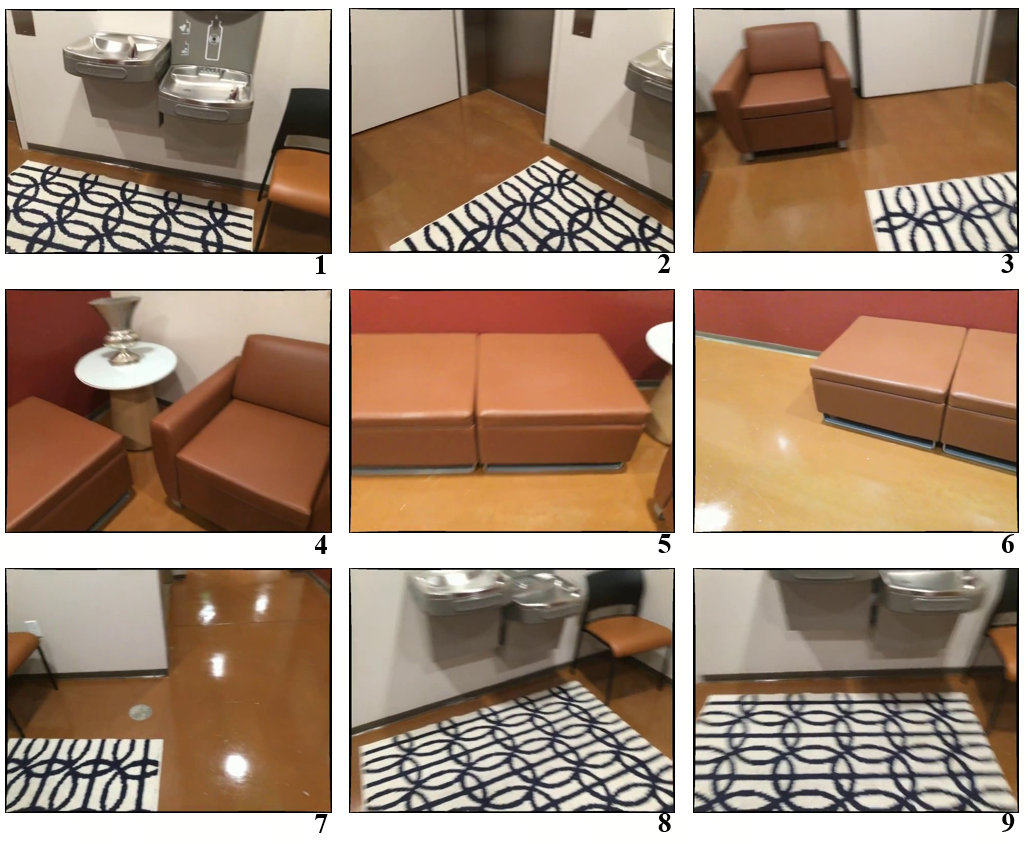}
\end{minipage}\hspace{1em}%
\begin{minipage}[t]{0.78\linewidth}\vspace{0pt}%
  \textbf{Comprehensive Reasoning (CR)}
  
  \medskip
  \textbf{Prompts:} You are a robot beginning at the door facing the brown single-seat sofa. You want to navigate to the chair next to the water fountain. 
  
You will perform the following actions (Note: for each [please fill in], choose either 'turn back,' 'turn left,' or 'turn right.'): 1. Go forward until the brown single-seat sofa. 2. [please fill in] 3. Go forward until passing by the brown two-seats sofa. 4. [please fill in] 5. Go forward until the chair. You have reached the final destination. 
[A. Turn Left, Turn Left, B. Turn Back, Turn Right, C. Turn Left, Turn Right, D. Turn Right, Turn Left]

  \textbf{GT:} A
  \medskip
  
  \textbf{Model Responses:} 

The robot starts facing a brown armchair, and I need to deduce the best turns to reach the chair near the water fountain. I'll review images 2, 8, and 9 for more clarity on the layout and confirm if the chair by the fountain is in these frames.

Frame 4 shows a brown armchair near a white wall, but it's not clear to me what the context is. Frame 5 depicts two single chairs with a table, and frame 6 shows two ottomans, possibly being part of a two-seat sofa. Frame 8, 9 all display water fountains and the chair location.

I need to figure out the robot's orientation based on the images. Frame 9 shows the water fountains, and the rug and furniture give clues. \hlg{After step 1, the robot faces a brown single-seat sofa. Step 2 requires a turn, and step 3 means passing by the two-seats sofa along the red wall. Deciding whether to turn left or right is key.}

I’m trying to figure out where everything is relative to each other. \hlg{I think the route from the armchair to the water fountain involves crossing a rug and passing by the two-seats sofa, which might be along the red wall.}

I’m considering the layout and analyzing different options for navigating from the door to the chair next to the water fountain. \hlr{Option D—turning right, passing by the two-seats along the red wall, then turning left toward the water fountain—seems to align well with the scene's features.} I’m leaning toward it as most probable.

I'm exploring the pathway to the water fountain, considering turns and placements. Based on the layout, \hlr{it seems that Option C—turn left to pass the two-seats, then turn right toward the water fountain—aligns well with the scene’s features.} It makes sense based on the furniture arrangement and directionality in the frames.

\textbf{Answer:} C. Turn Left, Turn Right
\end{minipage}
\par\vspace{0.6em}
\noindent\rule{\textwidth}{0.9pt}











%% file: tables/two_prompt/main_results_oo_so.tex
\begin{table*}[ht!]
\centering
\setlength{\tabcolsep}{1.2pt}
\resizebox{1.0\linewidth}{!}{

\begin{tabular}{l *{8}{cc}} 
\toprule
\multirow{2}{*}{\textbf{Models}} &
\multicolumn{2}{c}{\textbf{VSI}~\cite{yang2025thinking}} &
\multicolumn{2}{c}{\textbf{SITE}~\cite{wang2025site}} &
\multicolumn{2}{c}{\textbf{MMSI}~\cite{yang2025mmsi}} &
\multicolumn{2}{c}{\textbf{OmniSpatial}~\cite{jia2025omnispatial}} &
\multicolumn{2}{c}{\textbf{MindCube}$^*$~\cite{yin2025spatial}} &
\multicolumn{2}{c}{\textbf{STARE}~\cite{li2025unfolding}} &
\multicolumn{2}{c}{\textbf{CoreCognition}~\cite{li2024core}} &
\multicolumn{2}{c}{\textbf{SpatialViz}~\cite{wang2025spatialviz}} \\
& \subheads & \subheads & \subheads & \subheads & \subheads & \subheads & \subheads & \subheads \\
\cmidrule(lr){2-3}\cmidrule(lr){4-5}\cmidrule(lr){6-7}\cmidrule(lr){8-9}
\cmidrule(lr){10-11}\cmidrule(lr){12-13}\cmidrule(lr){14-15}\cmidrule(lr){16-17}

\rowcolor{line-blue}\textbf{Metric} &
\single{MRA, Acc} &
\single{CAA} &
\single{Acc} &
\single{Acc} &
\single{Acc} &
\single{Acc, F1} &
\single{Acc} &
\single{Acc} \\

\midrule

\rowcolor{line-blue}\textbf{Random Choice} &
\single{28.60} &  
\single{0.00}  &  
\single{25.00} &  
\single{24.98} &  
\single{32.35} &  
\single{34.80} &  
\single{37.70} &  
\single{25.08}    
\\

\midrule

\rowcolor{line-blue}\textbf{Proprietary Models} &&&&&&&&&&&&&&& \\

Seed-1.6-2025-06-15~\cite{seed2025seed1_5vl} &
\pair{45.24}{49.99\up} &
\pair{53.87}{54.61\up} &
\pair{38.30}{38.30} &
\pair{51.40}{49.32\down} &
\pair{50.67}{48.75\down} &
\pair{46.32}{46.06\down} &
\pair{74.54}{77.70\up} &
\pair{34.92}{34.58\down} \\

Gemini-2.5-pro-2025-06~\cite{team2023gemini} &
\pair{52.44}{53.57\up} &
\pair{56.43}{57.06\up} &
\pair{39.50}{38.00\down} &
\pair{56.03}{55.38\down} &
\pair{59.52}{57.60\down} &
\pair{49.03}{49.14\up} &
\pair{80.85}{76.70\down} &
\pair{47.54}{42.71\down} \\

Grok-4-2025-07-09~\cite{grok4_xai_2025} &
\pair{48.18}{47.92\down} &
\pair{49.16}{47.01\down} &
\pair{36.42}{37.80\up} &
\pair{48.47}{46.84\down} &
\pair{58.56}{63.56\up} &
\pair{29.79}{26.90\down} &
\pair{81.41}{80.93\down} &
\pair{20.02}{19.40\down} \\

GPT-5-nano-2025-08-07~\cite{openai_gpt5_systemcard} &
\pair{43.18}{43.22\up} &
\pair{37.67}{35.81\down} &
\pair{30.00}{28.90\down} &
\pair{52.05}{47.81\down} &
\pair{42.79}{41.48\down} &
\pair{45.39}{46.05\up} &
\pair{71.64}{68.80\down} &
\pair{34.92}{35.59\up} \\

GPT-5-mini-2025-08-07~\cite{openai_gpt5_systemcard} &
\pair{49.59}{48.67\down} &
\pair{53.39}{52.47\down} &
\pair{36.40}{34.10\down} &
\pair{56.36}{55.52\down} &
\pair{57.02}{56.69\down} &
\pair{51.91}{52.51\up} &
\pair{79.77}{78.42\down} &
\pair{46.19}{44.66\down} \\

GPT-5-2025-08-07~\cite{openai_gpt5_systemcard} &
\pair{52.82}{55.02\up} &
\pair{62.76}{61.88\down} &
\pair{40.10}{41.80\up} &
\pair{58.33}{59.90\up} &
\pair{56.73}{56.30\down} &
\pair{56.37}{54.59\down} &
\pair{85.16}{84.90\down} &
\pair{51.10}{51.27\up} \\

\midrule
\rowcolor{line-blue}\textbf{Open-source Models} &&&&&&&&&&&&&&& \\


Qwen2.5-VL-3B-Instruct~\cite{Qwen-VL} &
\pair{32.95}{27.00\down} &
\pair{28.19}{33.14\up} &
\pair{25.60}{28.60\up} &
\pair{40.70}{42.47\up} &
\pair{39.23}{37.60\down} &
\pair{35.94}{37.83\up} &
\pair{58.13}{59.20\up} &
\pair{22.46}{21.86\down} \\

Qwen2.5-VL-7B-Instruct~\cite{Qwen-VL} &
\pair{29.58}{32.30\up} &
\pair{31.96}{37.64\up} &
\pair{27.60}{26.80\down} &
\pair{39.53}{39.07\down} &
\pair{32.50}{36.05\up} &
\pair{40.19}{35.03\down} &
\pair{62.85}{61.36\down} &
\pair{27.71}{26.78\down} \\

Qwen2.5-VL-72B-Instruct~\cite{Qwen-VL} &
\pair{34.18}{35.77\up} &
\pair{44.06}{47.41\up} &
\pair{30.40}{32.50\up} &
\pair{48.01}{47.81\down} &
\pair{39.90}{42.40\up} &
\pair{41.32}{38.37\down} &
\pair{69.32}{69.90\up} &
\pair{29.83}{32.54\up} \\

InternVL3-8B~\cite{zhu2025internvl3} &
\pair{41.18}{42.14\up} &
\pair{38.74}{41.15\up} &
\pair{27.90}{28.00\up} &
\pair{46.38}{46.25\down} &
\pair{43.75}{41.54\down} &
\pair{40.95}{41.36\up} &
\pair{61.69}{57.88\down} &
\pair{29.58}{30.00\up} \\

InternVL3-78B~\cite{zhu2025internvl3} &
\pair{45.85}{47.55\up} &
\pair{49.08}{52.72\up} &
\pair{28.50}{30.50\up} &
\pair{51.40}{50.95\down} &
\pair{42.12}{49.52\up} &
\pair{42.64}{42.00\down} &
\pair{69.65}{69.24\down} &
\pair{32.12}{31.10\down} \\


Qwen3-8B-Instruct~\cite{yang2025qwen3technicalreport} &
\pair{39.68}{57.90\up} &
\pair{37.03}{45.83\up} &
\pair{27.60}{31.10\up} &
\pair{44.16}{45.73\up} &
\pair{38.56}{29.42\down} &
\pair{41.48}{39.76\down} &
\pair{69.08}{69.67\up} &
\pair{17.20}{17.54\up} \\

\midrule
\rowcolor{line-blue}\textbf{Human Evaluation} &&&&&&&&&&&&&&& \\
\textbf{Human} &
\single{\textbf{79.2}} &
\single{\textbf{67.5}} &
\single{\textbf{97.2}} &
\single{\textbf{92.63}} &
\single{\textbf{94.55}} &
\single{\textbf{96.65}} &
\single{\textbf{86.98}} &
\single{\textbf{82.46}} \\

\bottomrule
\end{tabular}
}

\vspace{-2pt}
\caption{\textbf{Evaluation on \numbench recent spatial benchmarks} \emph{(two prompt settings per column: \name / Official)}.
Each metric reported in the table is consistent with the definitions in the original papers.
MindCube$^*$ denotes MindCube-Tiny.
\textbf{\raisebox{0pt}[0pt][0pt]{\colorbox{bestcolor}{Dark purple}}} marks the best and
\raisebox{0pt}[0pt][0pt]{\colorbox{secondcolor}{light purple}} the second-best within Proprietary / Open-source blocks.
\emph{Prompts.} For the \textbf{\name} setting, we use the unified chain-of-thought prompt defined in \cref{fig:system_prompt}; for the \textbf{Official} setting, we follow each benchmark’s original prompt/inference protocol as specified by the respective papers (or their released code). The \textbf{Official} prompt format is also summarized in~\cref{sec:bench:eval_protocols}. \up~/~\down~indicate the model performs better / worse on \textbf{Official} prompt.
%
}

\label{tab:main_results_oo_so}
\end{table*}

%% file: tables/circular.tex
\begin{table*}[ht!]
\centering\small
\setlength{\tabcolsep}{2pt}

\begin{tabular}{ l *{3}{c} *{3}{c} *{3}{c}}
\toprule
\multirow{2}{*}{\textbf{Models}} &
\multicolumn{3}{c}{\textbf{SITE}} &
\multicolumn{3}{c}{\textbf{MMSI}} &
\multicolumn{3}{c}{\textbf{CoreCognition}} \\
\cmidrule(lr){2-4}\cmidrule(lr){5-7}\cmidrule(lr){8-10}
&
\makecell{Non} & 
\makecell{Soft} &
\makecell{Hard} & 
\makecell{Non} & 
\makecell{Soft} &
\makecell{Hard} & 
\makecell{Non} & 
\makecell{Soft} &
\makecell{Hard} \\
\midrule
GPT-5-nano-2025-08-07~\cite{openai_gpt5_systemcard} 
& 71.02 & 61.51& 47.96& 31.29 & 29.20& 9.76& 72.28& 72.43 & 66.81 \\
GPT-5-mini-2025-08-07~\cite{openai_gpt5_systemcard} 
& 75.94 &69.07 & 58.30& 33.13 & 32.24& 13.80& 80.20&79.78&71.35 \\
GPT-5-2025-08-07~\cite{openai_gpt5_systemcard} 
& 80.09& 78.30& 72.43& 41.86& 41.37& 26.14 & 86.60 &87.88&83.42\\

\bottomrule
\end{tabular}

\caption{
\textbf{Circular evaluation strategy comparison.} 
\textbf{Non}: Non-circular, standard tests without rotating options. \textbf{Soft}: Soft-circular, questions with rotated options are considered new questions.
\textbf{Hard}: Hard-circular, a question is considered correctly answered if all its rotations are correctly answered.
For \textbf{SITE}, the MultiV subset is excluded, as its large number of questions would make circular testing prohibitively time-consuming.
}


\label{tab:circular}
\end{table*}

%% file: tables/ss_tables/main_results_ss.tex
\begin{table*}[t]
\centering
\setlength{\tabcolsep}{3pt}
\resizebox{1.0\linewidth}{!}{
\begin{tabular}{l *{2}{c} *{8}{c}} 
\toprule
\textbf{Models} &
\textbf{Rank} &
\textbf{Avg.} &
\textbf{VSI}~\cite{yang2025thinking} &
\textbf{SITE}~\cite{wang2025site} &
\textbf{MMSI}~\cite{yang2025mmsi} &
\textbf{OmniSpatial}~\cite{jia2025omnispatial} &
\textbf{MindCube}$^*$~\cite{yin2025spatial} &
\textbf{STARE}~\cite{li2025unfolding} &
\textbf{CoreCognition}~\cite{li2024core} &
\textbf{SpatialViz}~\cite{wang2025spatialviz} \\
\midrule

\rowcolor{line-blue}\textbf{Random Choice} &
- & 0.0 & 0.0 & 0.0 & 0.0 & 0.0 & 0.0 & 0.0 & 0.0 & 0.0 \\
\midrule

\rowcolor{line-blue}\textbf{Proprietary Models} &&&&&&&& \\

Seed-1.6-2025-06-15~\cite{seed2025seed1_5vl} &
4 &
32.05 &
29.31 & 53.78 & 17.73 & 35.18 & 28.32 & 20.46 & 58.39 & 13.22 \\

Gemini-2.5-pro-2025-06~\cite{team2023gemini} &
2 &
\cellcolor{secondcolor}{40.25} &
\cellcolor{secondcolor}{35.79} & \cellcolor{secondcolor}{56.05} &
\cellcolor{secondcolor}{19.33} & 41.35 &
\cellcolor{bestcolor}{\textbf{39.53}} & 26.73 & 73.21 & \cellcolor{secondcolor}{30.05} \\

Grok-4-2025-07-09~\cite{grok4_xai_2025} &
5 &
29.38 &
28.35 & 48.65 & 15.23 & 31.26 & \cellcolor{secondcolor}{38.10} & 10.13 & \cellcolor{secondcolor}{69.93} & -6.64$^{\dagger}$ \\

GPT-5-nano-2025-08-07~\cite{openai_gpt5_systemcard} &
7 &
24.57 &
14.93 & 37.34 & 8.13 & 36.05 & 14.54 & 16.23 & 56.13 & 13.22 \\

GPT-5-mini-2025-08-07~\cite{openai_gpt5_systemcard} &
3 &
38.04 &
34.51 & 53.33 & 13.47 & \cellcolor{secondcolor}{41.79} & 35.80 & \cellcolor{secondcolor}{28.56} & 68.61 & 28.25 \\

GPT-5-2025-08-07~\cite{openai_gpt5_systemcard} &
1 &
\cellcolor{bestcolor}{\textbf{43.06}} &
\cellcolor{bestcolor}{\textbf{36.33}} &
\cellcolor{bestcolor}{\textbf{61.52}} &
\cellcolor{bestcolor}{\textbf{20.13}} &
\cellcolor{bestcolor}{\textbf{44.41}} &
35.37 & \cellcolor{bestcolor}{\textbf{33.46}} &
\cellcolor{bestcolor}{\textbf{78.45}} &
\cellcolor{bestcolor}{\textbf{34.80}} \\
\midrule

\rowcolor{line-blue}\textbf{Open-source Models} &&&&&&&& \\


Qwen2.5-VL-3B-Instruct~\cite{Qwen-VL} &
12 &
12.79 &
12.36 & 28.19 & -0.13 & 20.90 & 9.23 & -1.79 & 36.97 & -3.39 \\

Qwen2.5-VL-7B-Instruct~\cite{Qwen-VL} &
11 &
14.95 &
9.94 & 31.96 & 2.80 & 19.34 & -0.83 & 5.06 & 47.70 & 3.62 \\

Qwen2.5-VL-72B-Instruct~\cite{Qwen-VL} &
8 &
\cellcolor{secondcolor}{22.37} &
12.75 & \cellcolor{secondcolor}{44.06} &
\cellcolor{bestcolor}{\textbf{9.47}} &
\cellcolor{secondcolor}{30.65} &
10.23 & 9.96 &
\cellcolor{secondcolor}{55.41} &
\cellcolor{secondcolor}{6.44} \\

InternVL3-8B~\cite{zhu2025internvl3} &
9 &
20.14 &
13.99 & 38.74 & 6.00 & 28.48 &
\cellcolor{bestcolor}{\textbf{15.98}} & 6.20 & 45.66 & 6.10 \\

InternVL3-78B~\cite{zhu2025internvl3} &
6 &
\cellcolor{bestcolor}{\textbf{26.11}} &
\cellcolor{bestcolor}{\textbf{25.48}} &
\cellcolor{bestcolor}{\textbf{49.08}} &
\cellcolor{secondcolor}{6.27} & 
\cellcolor{bestcolor}{\textbf{35.18}} &
\cellcolor{secondcolor}{13.54} &
\cellcolor{bestcolor}{\textbf{11.79}} &
\cellcolor{bestcolor}{\textbf{58.01}} &
\cellcolor{bestcolor}{\textbf{9.49}} \\


Qwen3-8B-Instruct~\cite{yang2025qwen3technicalreport} & 10 & 18.84 &
\cellcolor{secondcolor}{22.05} & 37.03 &
3.47 & 25.52 & 8.22 & \cellcolor{secondcolor}{11.67} & 53.19 & -10.40$^{\dagger}$ \\

\midrule

\rowcolor{line-blue}\textbf{Human Evaluation} & & & & & & & & & & \\
\footnotesize $\Delta$(Best Model,Human) &
 &
\colornum{-43.35} &
\colornum{-58.75} &
\colornum{-5.62} &
\colornum{-76.14} &
\colornum{-45.77} &
\colornum{-52.41} &
\colornum{-61.17} &
\colornum{-0.65} &
\colornum{-41.79} \\

\textbf{Human} & - & \textbf{86.41} &
\textbf{95.08} & \textbf{67.5} & \textbf{96.27} & \textbf{90.18} &
\textbf{91.94} & \textbf{94.63} & \textbf{79.10} & \textbf{76.59} \\
\bottomrule
\end{tabular}
}

\vspace{-2pt}
\caption{
\textbf{Evaluation on \numbench recent spatial benchmarks (\name Protocol)}.
\textit{For values directly comparable with the official papers, refer to ~\cref{tab:main_results_oo}.}
Note that the reported metric is standardized to \textbf{\textit{Chance-Adjusted Accuracy (CAA)}}~\cite{wang2025site}: all values are calibrated such that random choice is always kept at 0.0. For human evaluations, we converted the original scores to CAA using~\cref{eq:caa}. 
Meanwhile, we use the unified chain-of-thought prompt defined in \cref{appendix:eval_prompt} during evaluation.
MindCube$^*$ denotes MindCube-Tiny. For VSI, only MCQ results are reported here.
$^{\dagger}$ indicates cases where generations were truncated due to overlong chains of thought, yielding no final answer; such instances are counted as incorrect, which depresses the score.
%
%
\textbf{\raisebox{0pt}[0pt][0pt]{\colorbox{bestcolor}{Dark purple}}} marks the best and
\raisebox{0pt}[0pt][0pt]{\colorbox{secondcolor}{light purple}} the second-best within Proprietary / Open-source blocks. 
See~\cref{appendix:detailed_bench_results} for detailed results.
}
\label{tab:main_results_ss}
\end{table*}

%% file: tables/oo_tables/vsi_oo.tex
\begin{table}[ht!]
\centering
\setlength{\tabcolsep}{2pt}
\resizebox{\textwidth}{!}{%
\begin{tabular}{l*{9}{c}} 
\toprule
\multirow{2}{*}{\textbf{Models}} &
\multirow{2}{*}{\textbf{Avg.}} &
\multicolumn{4}{c}{\textbf{Numerical Answer}} &
\multicolumn{4}{c}{\textbf{Multiple-Choice Answer}} \\
\cmidrule(lr){3-6}\cmidrule(lr){7-10}
&
\multicolumn{1}{c}{} &
\makecell{\textbf{Obj. Count}\\\textcolor{sensepurple}{\scriptsize CR}} &
\makecell{\textbf{Abs. Dist}\\\textcolor{sensepurple}{\scriptsize MM}} &
\makecell{\textbf{Obj. Size}\\\textcolor{sensepurple}{\scriptsize MM}} &
\makecell{\textbf{Room. Size}\\\textcolor{sensepurple}{\scriptsize MM}} &
\makecell{\textbf{Rel. Dis}\\\textcolor{sensepurple}{\scriptsize SR,MM}} &
\makecell{\textbf{Rel. Dir}\\\textcolor{sensepurple}{\scriptsize PT}} &
\makecell{\textbf{Route. Plan}\\\textcolor{sensepurple}{\scriptsize CR}} &
\makecell{\textbf{Appr. Order}\\\textcolor{sensepurple}{\scriptsize CR}} \\

\midrule
\rowcolor{line-blue}\textbf{Random Choice} &
- & - & - & - & - & 25.0 & 33.86 & 28.26 & 25.0 \\

\midrule
\rowcolor{line-blue}\multicolumn{10}{l}{\textbf{Proprietary Models}} \\

Seed-1.6-2025-06-15~\cite{seed2025seed1_5vl} &
49.91 & 43.54 & 34.36 & 66.12 & 52.81 & 55.07 & 35.74 & 44.33 & 67.96 \\

Gemini-2.5-pro-2025-06~\cite{team2023gemini} &
\cellcolor{secondcolor}{53.57} & 46.04 & \cellcolor{bestcolor}{\textbf{37.39}} & \cellcolor{secondcolor}{68.72} & \cellcolor{bestcolor}{\textbf{54.37}} & \cellcolor{secondcolor}{61.97} & 43.90 & \cellcolor{secondcolor}{47.42} & \cellcolor{secondcolor}{68.77} \\

Grok-4-2025-07-09~\cite{grok4_xai_2025} &
47.92 & 37.17 & 32.96 & 60.82 & 45.42 & 53.10 & 39.67 & \cellcolor{secondcolor}{47.42} & 66.83 \\

GPT-5-nano-2025-08-07~\cite{openai_gpt5_systemcard} &
43.22 & 44.65 & 29.69 & 63.85 & \cellcolor{secondcolor}{50.28} & 39.01 & 30.97 & 32.46 & 54.82 \\

GPT-5-mini-2025-08-07~\cite{openai_gpt5_systemcard} &
48.67 & \cellcolor{secondcolor}{49.75} & 25.18 & 66.97 & 42.40 & 52.82 & \cellcolor{secondcolor}{44.32} & 44.33 & 63.59 \\

GPT-5-2025-08-07~\cite{openai_gpt5_systemcard} &
\cellcolor{bestcolor}{\textbf{55.03}} & \cellcolor{bestcolor}{\textbf{53.31}} & \cellcolor{secondcolor}{34.47} & \cellcolor{bestcolor}{\textbf{73.31}} & 47.52 & \cellcolor{bestcolor}{\textbf{63.71}} & \cellcolor{bestcolor}{\textbf{48.69}} & \cellcolor{bestcolor}{\textbf{50.26}} & \cellcolor{bestcolor}{\textbf{68.93}} \\

\midrule
\rowcolor{line-blue}\multicolumn{10}{l}{\textbf{Open-source Models}} \\


Qwen2.5-VL-3B-Instruct~\cite{Qwen-VL} &
27.00 & 19.17 & 21.22 & 24.27 & 27.29 & 33.80 & 42.09 & 27.32 & 20.87 \\

Qwen2.5-VL-7B-Instruct~\cite{Qwen-VL} &
32.30 & 32.89 & 18.17 & 43.88 & 31.70 & 38.03 & 37.42 & 28.35 & 27.99 \\

Qwen2.5-VL-72B-Instruct~\cite{Qwen-VL} &
35.77 & 27.43 & 26.27 & 58.52 & 40.07 & 41.97 & 33.26 & 21.13 & 37.54 \\

InternVL3-8B~\cite{zhu2025internvl3} &
42.14 & 66.05 & 34.89 & 43.64 & 47.50 & 48.03 & 39.31 & 26.29 & 31.39 \\

InternVL3-78B~\cite{zhu2025internvl3} &
47.55 & \cellcolor{bestcolor}{\textbf{70.97}} & 38.47 & 53.20 & 43.99 & \cellcolor{secondcolor}{55.77} & 38.43 & 27.32 & 52.27 \\

InternVL3.5-8B~\cite{wang2025internvl3} &
\cellcolor{secondcolor}{56.06} &\cellcolor{secondcolor}{68.58} & \cellcolor{secondcolor}{42.21} & \cellcolor{secondcolor}{68.10} & \cellcolor{secondcolor}{65.21} & 55.49 & \cellcolor{secondcolor}{48.19} & \cellcolor{secondcolor}{39.18} & \cellcolor{secondcolor}{61.49} \\

Qwen3-8B-Instruct~\cite{yang2025qwen3technicalreport} &
\cellcolor{bestcolor}{\textbf{57.90}} & 67.58 & \cellcolor{bestcolor}{\textbf{47.00}} & \cellcolor{bestcolor}{\textbf{76.32}} & \cellcolor{bestcolor}{\textbf{61.94}} & \cellcolor{bestcolor}{\textbf{58.03}} & \cellcolor{bestcolor}{\textbf{50.97}} & \cellcolor{bestcolor}{\textbf{35.05}} & \cellcolor{bestcolor}{\textbf{66.34}} \\

\midrule
\rowcolor{line-blue}\multicolumn{10}{l}{\textbf{Human Evaluation}} \\
\footnotesize $\Delta$(Best Model,Human) & 
\colornum{-21.3} &
\colornum{-23.33} &
\colornum{0.0} &
\colornum{15.92} &
\colornum{16.04} &
\colornum{-30.99} &
\colornum{-44.83} &
\colornum{-45.54} &
\colornum{-31.07} \\

Human & \textbf{79.2} & \textbf{94.3} & \textbf{47.0} & \textbf{60.4} & \textbf{45.9} & \textbf{94.7} & \textbf{95.8} & \textbf{95.8} & \textbf{100.0} \\
\bottomrule
\end{tabular}
}

\vspace{-2pt}
\caption{\textbf{Evaluation on \textbf{VSI-Bench} (Official Protocol)}.
Numerical Answer uses \textbf{\textit{MRA}} score via \cref{eq:mra}; MCQ uses \textbf{\textit{Acc}} score. \textbf{Avg.} is the simple average across these metrics, following the original paper.
\textit{Prompt}: we follow the \textbf{VSI-Bench} prompt: MCQ items are answered by \textit{Direct QA} (choose the option directly), and numerical items require a \textit{single float number} directly.
}
\label{tab:vsi_oo}
\end{table}

%% file: tables/ss_tables/vsi_ss.tex
\begin{table}[ht!]
\centering\small
\setlength{\tabcolsep}{2pt}
\resizebox{\textwidth}{!}{%
\begin{tabular}{l*{9}{c}} 
\toprule
\multirow{2}{*}{\textbf{Models}} &
\multirow{2}{*}{\textbf{Avg.}} &
\multicolumn{4}{c}{\textbf{Numerical Answer}} &
\multicolumn{4}{c}{\textbf{Multiple-Choice Answer}} \\
\cmidrule(lr){3-6}\cmidrule(lr){7-10}
&
\multicolumn{1}{c}{} &
\makecell{\textbf{Obj. Count}\\\textcolor{sensepurple}{\scriptsize CR}} &
\makecell{\textbf{Abs. Dist}\\\textcolor{sensepurple}{\scriptsize MM}} &
\makecell{\textbf{Obj. Size}\\\textcolor{sensepurple}{\scriptsize MM}} &
\makecell{\textbf{Room. Size}\\\textcolor{sensepurple}{\scriptsize MM}} &
\makecell{\textbf{Rel. Dis}\\\textcolor{sensepurple}{\scriptsize SR,MM}} &
\makecell{\textbf{Rel. Dir}\\\textcolor{sensepurple}{\scriptsize PT}} &
\makecell{\textbf{Route. Plan}\\\textcolor{sensepurple}{\scriptsize CR}} &
\makecell{\textbf{Appr. Order}\\\textcolor{sensepurple}{\scriptsize CR}} \\

\midrule
\rowcolor{line-blue}\textbf{Random Choice} &
0.0 & - & - & - & - & 0.0 & 0.0 & 0.0 & 0.0 \\

\midrule

\rowcolor{line-blue}\multicolumn{10}{l}{\textbf{Proprietary Models}} \\

Seed-1.6-2025-06-15~\cite{seed2025seed1_5vl} &
35.07 & 37.36 & 31.89 & 54.52 & 38.09 & 39.72 & 4.72 & 20.24 & 54.05 \\

Gemini-2.5-pro-2025-06~\cite{team2023gemini} &
\cellcolor{secondcolor}{43.03} & 44.94 & \cellcolor{bestcolor}{\textbf{37.91}} & \cellcolor{secondcolor}{70.55} & \cellcolor{bestcolor}{\textbf{51.81}} & \cellcolor{secondcolor}{45.16} & \cellcolor{bestcolor}{\textbf{15.81}} & \cellcolor{secondcolor}{20.96} & \cellcolor{secondcolor}{57.07} \\

Grok-4-2025-07-09~\cite{grok4_xai_2025} &
38.63 & 40.30 & 30.47 & 68.80 & 49.51 & 35.21 & 5.19 & 17.37 & 55.92 \\

GPT-5-nano-2025-08-07~\cite{openai_gpt5_systemcard} &
31.09 & 47.30 & 31.02 & 63.42 & 45.52 & 22.63 & -4.36 & 8.74 & 34.41 \\

GPT-5-mini-2025-08-07~\cite{openai_gpt5_systemcard} &
40.21 & \cellcolor{secondcolor}{51.14} & 24.74 & 66.49 & 39.51 & 42.54 & 14.10 & \cellcolor{bestcolor}{\textbf{25.59}} & \cellcolor{bestcolor}{\textbf{57.54}} \\

GPT-5-2025-08-07~\cite{openai_gpt5_systemcard} &
\cellcolor{bestcolor}{\textbf{43.77}} & \cellcolor{bestcolor}{\textbf{53.61}} & \cellcolor{secondcolor}{33.62} & \cellcolor{bestcolor}{\textbf{73.72}} & \cellcolor{secondcolor}{50.53} & \cellcolor{bestcolor}{\textbf{52.05}} & \cellcolor{secondcolor}{14.25} & 18.08 & 54.26 \\

\midrule
\rowcolor{line-blue}\multicolumn{10}{l}{\textbf{Open-source Models}} \\


Qwen2.5-VL-3B-Instruct~\cite{Qwen-VL} &
20.48 & 29.50 & 23.97 & 34.94 & 31.15 & 14.18 & 12.07 & \cellcolor{secondcolor}{5.15} & 12.84 \\

Qwen2.5-VL-7B-Instruct~\cite{Qwen-VL} &
16.69 & 26.62 & 24.51 & 26.61 & 22.99 & 12.11 & 9.25  & 0.12 & 11.33 \\

Qwen2.5-VL-72B-Instruct~\cite{Qwen-VL} &
21.63 & 19.54 & 25.18 & 43.77 & \cellcolor{secondcolor}{39.76} & 17.56 & 5.35 & 0.84 & 21.04 \\

InternVL3-8B~\cite{zhu2025internvl3} &
\cellcolor{secondcolor}{28.76} & \cellcolor{bestcolor}{\textbf{59.77}} & \cellcolor{secondcolor}{36.45} & \cellcolor{secondcolor}{55.16} & 32.50 & 23.57 & \cellcolor{secondcolor}{13.16} & 1.56 & 7.87 \\

InternVL3-78B~\cite{zhu2025internvl3} &
\cellcolor{bestcolor}{\textbf{34.80}} & \cellcolor{secondcolor}{50.65} & \cellcolor{bestcolor}{\textbf{37.89}} & \cellcolor{bestcolor}{\textbf{56.59}} & \cellcolor{bestcolor}{\textbf{44.13}} & \cellcolor{bestcolor}{\textbf{35.21}} & \cellcolor{bestcolor}{\textbf{15.19}} & 3.71 & \cellcolor{bestcolor}{\textbf{35.06}} \\


Qwen3-8B-Instruct~\cite{yang2025qwen3technicalreport} &
28.42 & 22.50 & 34.29 & 54.18 & 33.16 & \cellcolor{secondcolor}{33.33} & 10.19 & \cellcolor{bestcolor}{\textbf{10.90}} & \cellcolor{secondcolor}{28.80} \\


\midrule
\rowcolor{line-blue}\multicolumn{10}{l}{\textbf{Human Evaluation}} \\
\footnotesize $\Delta$(Best Model,Human) & 
\colornum{-34.77} &
\colornum{-34.53} &
\colornum{-9.09} &
\colornum{13.32} &
\colornum{5.91} &
\colornum{-40.88} &
\colornum{-77.84} &
\colornum{-68.56} &
\colornum{-42.46} \\

Human & \textbf{78.54} & \textbf{94.3} & \textbf{47.0} & \textbf{60.4} & \textbf{45.9} & \textbf{92.93} & \textbf{93.65} & \textbf{94.15} & \textbf{100.00} \\
\bottomrule
\end{tabular}
}

\vspace{-2pt}
\caption{\textbf{Evaluation on \textbf{VSI-Bench} (\name Protocol)}.
Numerical Answer uses \textbf{\textit{MRA}} score via \cref{eq:mra}; MCQ uses \textbf{\textit{CAA}} score. \textbf{Avg.} is the simple average across these metrics.
\textit{Prompt}: we use the unified chain-of-thought prompt during evaluation, defined in \cref{fig:system_prompt}.
}
\label{tab:vsi_ss}
\end{table}


%% file: tables/oo_tables/site_oo.tex
\begin{table*}[ht!]
\centering\small
\setlength{\tabcolsep}{1.2pt}
\begin{adjustbox}{max width=\linewidth} 
\begin{tabular}{l *{7}{c}}
\toprule
\textbf{Models} &
\textbf{Overall} &
\makecell{\textbf{Count}\\\textcolor{sensepurple}{\scriptsize -}} &
\makecell{\textbf{Loc}\\\textcolor{sensepurple}{\scriptsize -}} &
\makecell{\textbf{3D Inf}\\\textcolor{sensepurple}{\scriptsize MM,SR}} &
\makecell{\textbf{MultiV}\\\textcolor{sensepurple}{\scriptsize PT}} &
\makecell{\textbf{Rel}\\\textcolor{sensepurple}{\scriptsize SR}} &
\makecell{\textbf{Mov}\\\textcolor{sensepurple}{\scriptsize CR}} \\

\midrule
\rowcolor{line-blue}\textbf{Random Choice} & 0.0 & 0.0 & 0.0 & 0.0 & 0.0 & 0.0 & 0.0 \\

\midrule
\rowcolor{line-blue}\multicolumn{8}{l}{\textbf{Proprietary Models}} \\

Seed-1.6-2025-06-15~\cite{seed2025seed1_5vl} &
54.61 & \cellcolor{secondcolor}{61.96} & 66.45 & \cellcolor{bestcolor}{\textbf{60.41}} & 37.13 & 70.59 & 32.44 \\

Gemini-2.5-pro-2025-06~\cite{team2023gemini} &
\cellcolor{secondcolor}{57.06} & 61.31 & \cellcolor{bestcolor}{\textbf{69.18}} & 55.17 & \cellcolor{secondcolor}{38.52} & \cellcolor{secondcolor}{71.51} & \cellcolor{secondcolor}{48.62} \\

Grok-4-2025-07-09~\cite{grok4_xai_2025} &
47.01 & 50.44 & 60.33 & 51.57 & 26.21 & 61.22 & 37.42 \\

GPT-5-nano-2025-08-07~\cite{openai_gpt5_systemcard} &
35.81 & 46.03 & 49.16 & 39.67 & 7.27 & 54.42 & 21.45 \\

GPT-5-mini-2025-08-07~\cite{openai_gpt5_systemcard} &
52.47 & 55.54 & 61.16 & 54.43 & 33.58 & 68.02 & 44.62 \\

GPT-5-2025-08-07~\cite{openai_gpt5_systemcard} &
\cellcolor{bestcolor}{\textbf{61.88}} & \cellcolor{bestcolor}{\textbf{63.08}} & \cellcolor{secondcolor}{68.62} & \cellcolor{secondcolor}{55.38} & \cellcolor{bestcolor}{\textbf{51.41}} & \cellcolor{bestcolor}{\textbf{72.50}} & \cellcolor{bestcolor}{\textbf{59.24}} \\

\midrule
\rowcolor{line-blue}\multicolumn{8}{l}{\textbf{Open-source Models}} \\


Qwen2.5-VL-3B-Instruct~\cite{Qwen-VL} &
33.14 & 48.46 & 42.75 & 20.54 & 9.45 & 56.22 & 12.71 \\

Qwen2.5-VL-7B-Instruct~\cite{Qwen-VL} &
37.64 & 54.36 & 44.77 & 23.32 & 12.88 & 63.69 & 15.93 \\

Qwen2.5-VL-72B-Instruct~\cite{Qwen-VL} &
\cellcolor{secondcolor}{47.41} & \cellcolor{secondcolor}{59.80} & \cellcolor{secondcolor}{61.47} & 29.33 & \cellcolor{bestcolor}{\textbf{28.29}} & \cellcolor{secondcolor}{69.77} & 27.59 \\

InternVL3-8B~\cite{zhu2025internvl3} &
41.15 & 57.91 & 51.67 & 33.99 & 10.82 & 61.87 & 26.35 \\

InternVL3-78B~\cite{zhu2025internvl3} &
\cellcolor{bestcolor}{\textbf{52.72}} & \cellcolor{bestcolor}{\textbf{64.48}} & \cellcolor{bestcolor}{\textbf{66.04}} & \cellcolor{bestcolor}{\textbf{61.88}} & 16.15 & \cellcolor{bestcolor}{\textbf{73.99}} & \cellcolor{bestcolor}{\textbf{40.28}} \\

InternVL3.5-8B~\cite{wang2025internvl3} &
43.79 & 59.13 & 55.84 & \cellcolor{secondcolor}{45.58} & 9.36 & 61.17 & \cellcolor{secondcolor}{33.41} \\

Qwen3-8B-Instruct~\cite{yang2025qwen3technicalreport} &
45.83 & 55.81 & 60.30 & 32.29 & \cellcolor{secondcolor}{23.25} & 67.78 & 30.19 \\

\midrule
\rowcolor{line-blue}\multicolumn{8}{l}{\textbf{Human Evaluation}} \\
\footnotesize $\Delta$(Best Model,Human) &
\colornum{-5.62} & \colornum{-1.52} & \colornum{-14.12} & \colornum{7.18} &
\colornum{-36.09} & \colornum{0.99} & \colornum{6.74} \\

Human &
\textbf{67.5} & \textbf{66} & \textbf{83.3} & \textbf{54.7} & \textbf{87.5} & \textbf{73} & \textbf{52.5} \\
\bottomrule
\end{tabular}
\end{adjustbox}
\vspace{-2pt}
\caption{\textbf{Evaluation on \textbf{SITE} (Official Protocol)}.
All scores are \textbf{\textit{CAA}} (computed via \cref{eq:caa}). We follow the \textbf{SITE} paper’s original prompt/inference protocol, where MCQ items are answered by \textit{direct QA}.
Note that the Official Metric of SITE and the \name Metric are identical.
}
\label{tab:site_oo}
\end{table*}

%% file: tables/ss_tables/site_ss.tex
\begin{table*}[ht!]
\centering\small
\setlength{\tabcolsep}{1.2pt}
\begin{adjustbox}{max width=\linewidth} 
\begin{tabular}{l *{7}{c}}
\toprule
\textbf{Models} &
\textbf{Overall} &
\makecell{\textbf{Count}\\\textcolor{sensepurple}{\scriptsize -}} &
\makecell{\textbf{Loc}\\\textcolor{sensepurple}{\scriptsize -}} &
\makecell{\textbf{3D Inf}\\\textcolor{sensepurple}{\scriptsize MM,SR}} &
\makecell{\textbf{MultiV}\\\textcolor{sensepurple}{\scriptsize PT}} &
\makecell{\textbf{Rel}\\\textcolor{sensepurple}{\scriptsize SR}} &
\makecell{\textbf{Mov}\\\textcolor{sensepurple}{\scriptsize CR}} \\

\midrule
\rowcolor{line-blue}\textbf{Random Choice} & 0.0 & 0.0 & 0.0 & 0.0 & 0.0 & 0.0 & 0.0 \\

\midrule
\rowcolor{line-blue}\multicolumn{8}{l}{\textbf{Proprietary Models}} \\

Seed-1.6-2025-06-15~\cite{seed2025seed1_5vl} &
53.78 &
\cellcolor{secondcolor}{61.64} &
65.04 &
\cellcolor{bestcolor}{\textbf{58.28}} &
33.66 &
69.77 &
36.05 \\

Gemini-2.5-pro-2025-06~\cite{team2023gemini} &
\cellcolor{secondcolor}{56.05} &
59.39 &
\cellcolor{secondcolor}{70.32} &
53.37 &
36.70 &
\cellcolor{secondcolor}{72.06} &
\cellcolor{secondcolor}{46.75} \\

Grok-4-2025-07-09~\cite{grok4_xai_2025} &
48.65 &
51.70 &
60.04 &
53.87 &
25.43 &
66.55 &
39.28 \\

GPT-5-nano-2025-08-07~\cite{openai_gpt5_systemcard} &
37.34 &
47.85 &
53.48 &
45.69 &
6.35 &
56.62 &
19.63 \\

GPT-5-mini-2025-08-07~\cite{openai_gpt5_systemcard} &
53.33 &
57.21 &
64.18 &
51.57 &
35.23 &
68.94 &
44.26 \\

GPT-5-2025-08-07~\cite{openai_gpt5_systemcard} &
\cellcolor{bestcolor}{\textbf{61.52}} &
\cellcolor{bestcolor}{\textbf{64.57}} &
\cellcolor{bestcolor}{\textbf{73.03}} &
\cellcolor{secondcolor}{58.12} &
\cellcolor{bestcolor}{\textbf{51.35}} &
\cellcolor{bestcolor}{\textbf{74.08}} &
\cellcolor{bestcolor}{\textbf{47.12}} \\

\midrule
\rowcolor{line-blue}\multicolumn{8}{l}{\textbf{Open-source Models}} \\

Qwen2.5-VL-3B-Instruct~\cite{Qwen-VL} &
28.19 &
43.09 &
34.07 &
15.42 &
5.48 &
47.43 &
17.14 \\

Qwen2.5-VL-7B-Instruct~\cite{Qwen-VL} &
31.96 &
47.44 &
42.06 &
19.02 &
9.13 &
53.41 &
13.65 \\

Qwen2.5-VL-72B-Instruct~\cite{Qwen-VL} &
\cellcolor{secondcolor}{44.06} &
\cellcolor{secondcolor}{54.12} &
\cellcolor{secondcolor}{56.62} &
\cellcolor{secondcolor}{42.90} &
\cellcolor{bestcolor}{\textbf{18.40}} &
\cellcolor{secondcolor}{65.63} &
\cellcolor{secondcolor}{26.59} \\

InternVL3-8B~\cite{zhu2025internvl3} &
38.74 &
53.37 &
49.05 &
38.98 &
9.21 &
58.19 &
23.86 \\

InternVL3-78B~\cite{zhu2025internvl3} &
\cellcolor{bestcolor}{\textbf{49.08}} &
\cellcolor{bestcolor}{\textbf{64.40}} &
\cellcolor{bestcolor}{\textbf{61.76}} &
\cellcolor{bestcolor}{\textbf{56.65}} &
11.64 &
\cellcolor{bestcolor}{\textbf{70.68}} &
\cellcolor{bestcolor}{\textbf{33.93}} \\


Qwen3-8B-Instruct~\cite{yang2025qwen3technicalreport} &
37.03 & 45.35 & 54.91 & 39.63 & \cellcolor{secondcolor}{15.37} & 47.89 & 23.48 \\

\midrule
\rowcolor{line-blue}\multicolumn{8}{l}{\textbf{Human Evaluation}} \\
\footnotesize $\Delta$(Best Model,Human) &
\colornum{-5.98} & 
\colornum{-1.43} & 
\colornum{-10.27} & 
\colornum{3.58} &
\colornum{-36.15} & 
\colornum{1.08} & 
\colornum{-5.38} \\

Human &
\textbf{67.5} & \textbf{66} & \textbf{83.3} & \textbf{54.7} & \textbf{87.5} & \textbf{73} & \textbf{52.5} \\
\bottomrule
\end{tabular}
\end{adjustbox}
\vspace{-2pt}
\caption{\textbf{Evaluation on \textbf{SITE} (\name Protocol)}.
All scores are \textbf{\textit{CAA}} (computed via \cref{eq:caa}). We use the unified chain-of-thought prompt during evaluation, defined in \cref{fig:system_prompt}.
}
\label{tab:site_ss}
\end{table*}


%% file: tables/oo_tables/mmsi_oo.tex
\begin{table*}[ht!]
\centering
\setlength{\tabcolsep}{2pt}
\renewcommand{\arraystretch}{1}
\resizebox{\textwidth}{!}{%
\begin{tabular}{@{}>{\raggedright\arraybackslash}p{\ModelW} *{12}{M}@{}}
\toprule
\multirow{2}{*}{\textbf{Models}} &
\multirow{2}{*}{\textbf{Avg.}} &
\multicolumn{6}{c}{\textbf{Positional Relationship}} &
\multicolumn{2}{c}{\textbf{Attribute}} &
\multicolumn{2}{c}{\textbf{Motion}} &
\multicolumn{1}{c}{\textbf{MSR}} \\
\cmidrule(lr){3-8}\cmidrule(lr){9-10}\cmidrule(lr){11-12}\cmidrule(lr){13-13}
& &
\makecell{\textbf{C–C}\\\textcolor{sensepurple}{\scriptsize PT}} &
\makecell{\textbf{O–O}\\\textcolor{sensepurple}{\scriptsize PT}} &
\makecell{\textbf{R–R}\\\textcolor{sensepurple}{\scriptsize PT}} &
\makecell{\textbf{C–O}\\\textcolor{sensepurple}{\scriptsize PT}} &
\makecell{\textbf{O–R}\\\textcolor{sensepurple}{\scriptsize PT}} &
\makecell{\textbf{C–R}\\\textcolor{sensepurple}{\scriptsize PT}} &
\makecell{\textbf{Meas.}\\\textcolor{sensepurple}{\scriptsize MM}} &
\makecell{\textbf{Appr.}\\\textcolor{sensepurple}{\scriptsize MR}} &
\makecell{\textbf{Cam.}\\\textcolor{sensepurple}{\scriptsize PT}} &
\makecell{\textbf{Obj.}\\\textcolor{sensepurple}{\scriptsize PT}} &
\makecell{\textbf{--}\\\textcolor{sensepurple}{\scriptsize CR}} \\

\midrule
\rowcolor{line-blue}\textbf{Random Choice} &
25.0 & 25.0 & 25.0 & 25.0 & 25.0 & 25.0 & 25.0 & 25.0 & 25.0 & 25.0 & 25.0 & 25.0 \\

\midrule
\rowcolor{line-blue}\multicolumn{13}{l}{\textbf{Proprietary Models}} \\
Seed-1.6-2025-06-15~\cite{seed2025seed1_5vl} &
\cellcolor{secondcolor}{38.30} & 36.56 & \cellcolor{bestcolor}{\textbf{36.17}} & 32.10 & 32.56 & \cellcolor{secondcolor}{42.35} & 46.94 & 48.44 & \cellcolor{secondcolor}{33.00} & 31.08 & \cellcolor{secondcolor}{42.11} & \cellcolor{bestcolor}{\textbf{40.40}} \\

Gemini-2.5-pro-2025-06~\cite{team2023gemini} &
38.00 & \cellcolor{secondcolor}{38.71} & 34.04 & \cellcolor{bestcolor}{\textbf{40.74}} & \cellcolor{secondcolor}{44.19} & 38.82 & 40.96 & \cellcolor{bestcolor}{\textbf{62.50}} & 30.30 & \cellcolor{secondcolor}{39.19} & 25.00 & 33.33 \\

Grok-4-2025-07-09~\cite{grok4_xai_2025} &
37.80 & 36.56 & \cellcolor{secondcolor}{35.11} & \cellcolor{secondcolor}{39.51} & 34.88 & \cellcolor{bestcolor}{\textbf{45.88}} & 50.60 & 21.88 & 22.73 & \cellcolor{bestcolor}{\textbf{40.54}} & \cellcolor{bestcolor}{\textbf{43.42}} & \cellcolor{secondcolor}{38.38} \\

GPT-5-nano-2025-08-07~\cite{openai_gpt5_systemcard} &
28.90 & 31.18 & 29.79 & 25.93 & 26.74 & 31.76 & 33.73 & 45.31 & 22.73 & 16.22 & 25.00 & 29.29 \\

GPT-5-mini-2025-08-07~\cite{openai_gpt5_systemcard} &
34.10 & 34.41 & 29.79 & 28.40 & 29.07 & 34.12 & \cellcolor{secondcolor}{51.81} & 50.00 & 31.82 & 29.73 & 28.95 & 32.32 \\

GPT-5-2025-08-07~\cite{openai_gpt5_systemcard} &
\cellcolor{bestcolor}{\textbf{41.80}} & \cellcolor{bestcolor}{\textbf{41.94}} & 32.98 & 35.80 & \cellcolor{bestcolor}{\textbf{49.84}} & \cellcolor{secondcolor}{42.35} & \cellcolor{bestcolor}{\textbf{68.67}} & \cellcolor{secondcolor}{54.69} & \cellcolor{bestcolor}{\textbf{37.38}} & 28.33 & 40.79 & 36.36 \\

\midrule
\rowcolor{line-blue}\multicolumn{13}{l}{\textbf{Open-source Models}} \\


Qwen2.5-VL-3B-Instruct~\cite{Qwen-VL} &
28.60 & \cellcolor{secondcolor}{36.56} & 30.85 & 28.40 & 26.74 & 28.24 & 31.33 & 31.25 & 16.67 & 16.22 & 35.53 & 28.79 \\

Qwen2.5-VL-7B-Instruct~\cite{Qwen-VL} &
26.80 & 27.96 & 26.60 & 19.75 & \cellcolor{bestcolor}{\textbf{32.56}} & \cellcolor{secondcolor}{38.82} & 28.92 & 23.44 & 21.21 & 20.27 & 30.26 & 24.75 \\

Qwen2.5-VL-72B-Instruct~\cite{Qwen-VL} &
\cellcolor{bestcolor}{\textbf{32.50}} & 23.66 & \cellcolor{secondcolor}{30.85} & \cellcolor{bestcolor}{\textbf{41.98}} & 23.26 & \cellcolor{secondcolor}{38.82} & 28.92 & \cellcolor{bestcolor}{\textbf{42.19}} & \cellcolor{secondcolor}{25.76} & \cellcolor{bestcolor}{\textbf{31.08}} & \cellcolor{bestcolor}{\textbf{40.79}} & \cellcolor{bestcolor}{\textbf{32.83}} \\

InternVL3-8B~\cite{zhu2025internvl3} &
28.00 & 22.58 & 22.34 & 34.57 & \cellcolor{secondcolor}{31.40} & \cellcolor{bestcolor}{\textbf{42.35}} & \cellcolor{secondcolor}{33.73} & 25.00 & 19.70 & 20.27 & 34.21 & 24.75 \\

InternVL3-78B~\cite{zhu2025internvl3} &
30.50 & 35.48 & 23.40 & \cellcolor{secondcolor}{32.10} & 18.60 & 36.47 & 31.33 & \cellcolor{bestcolor}{\textbf{42.19}} & \cellcolor{bestcolor}{\textbf{27.27}} & \cellcolor{secondcolor}{29.73} & 31.58 & \cellcolor{secondcolor}{30.30} \\

InternVL3.5-8B~\cite{wang2025internvl3} &
27.30 & \cellcolor{bestcolor}{\textbf{37.63}} & 24.47 & 24.69 & 26.74 & 27.06 & 25.30 & 34.38 & 16.67 & 12.16 & \cellcolor{secondcolor}{36.84} & 29.29 \\

Qwen3-8B-Instruct~\cite{yang2025qwen3technicalreport} &
\cellcolor{secondcolor}{31.10} & 27.96 & \cellcolor{bestcolor}{\textbf{37.23}} & \cellcolor{secondcolor}{32.10} & \cellcolor{secondcolor}{31.40} & 35.29 & \cellcolor{bestcolor}{\textbf{38.55}} & \cellcolor{secondcolor}{37.50} & 15.15 & 27.03 & 28.95 & 29.80 \\

\midrule
\rowcolor{line-blue}\multicolumn{13}{l}{\textbf{Human Evaluation}} \\
\footnotesize $\Delta$(Best Model,Human) &
\colornum{-55.40} & 
\colornum{-53.76} & 
\colornum{-61.67} & 
\colornum{-55.52} &
\colornum{-44.36} & 
\colornum{-52.92} & 
\colornum{-27.73} & 
\colornum{-32.80} &
\colornum{-61.12} &
\colornum{-58.06} &
\colornum{-55.28} & 
\colornum{-56.60} \\

Human &
\textbf{97.2} & \textbf{95.7} & \textbf{98.9} & \textbf{97.5} & \textbf{94.2} & \textbf{98.8} & \textbf{96.4} & \textbf{95.3} & \textbf{98.5} & \textbf{98.6} & \textbf{98.7} & \textbf{97.0} \\
\bottomrule
\end{tabular}
}
\vspace{-2pt}
\caption{\textbf{Evaluation on \textbf{MMSI} (Official Protocol)}. Scores are \textbf{\textit{Acc}} as in the original paper.
\textit{Prompt}: we follow the \textbf{MMSI} paper’s prompt/inference protocol, answering MCQ via \textit{Direct QA} (choose the option directly). Under Positional Relationship, C: Camera; O: Object; R: Region.}
\label{tab:mmsi_oo}
\end{table*}

%% file: tables/ss_tables/mmsi_ss.tex
\begin{table*}[ht!]
\centering
\setlength{\tabcolsep}{2pt}
\renewcommand{\arraystretch}{1}
\resizebox{\textwidth}{!}{%
\begin{tabular}{@{}>{\raggedright\arraybackslash}p{\ModelW} *{12}{M}@{}}
\toprule
\multirow{2}{*}{\textbf{Models}} &
\multirow{2}{*}{\textbf{Avg.}} &
\multicolumn{6}{c}{\textbf{Positional Relationship}} &
\multicolumn{2}{c}{\textbf{Attribute}} &
\multicolumn{2}{c}{\textbf{Motion}} &
\multicolumn{1}{c}{\textbf{MSR}} \\
\cmidrule(lr){3-8}\cmidrule(lr){9-10}\cmidrule(lr){11-12}\cmidrule(lr){13-13}
& &
\makecell{\textbf{C–C}\\\textcolor{sensepurple}{\scriptsize PT}} &
\makecell{\textbf{O–O}\\\textcolor{sensepurple}{\scriptsize PT}} &
\makecell{\textbf{R–R}\\\textcolor{sensepurple}{\scriptsize PT}} &
\makecell{\textbf{C–O}\\\textcolor{sensepurple}{\scriptsize PT}} &
\makecell{\textbf{O–R}\\\textcolor{sensepurple}{\scriptsize PT}} &
\makecell{\textbf{C–R}\\\textcolor{sensepurple}{\scriptsize PT}} &
\makecell{\textbf{Meas.}\\\textcolor{sensepurple}{\scriptsize MM}} &
\makecell{\textbf{Appr.}\\\textcolor{sensepurple}{\scriptsize MR}} &
\makecell{\textbf{Cam.}\\\textcolor{sensepurple}{\scriptsize PT}} &
\makecell{\textbf{Obj.}\\\textcolor{sensepurple}{\scriptsize PT}} &
\makecell{\textbf{--}\\\textcolor{sensepurple}{\scriptsize CR}} \\

\midrule
\rowcolor{line-blue}\textbf{Random Choice} &
0.0 & 0.0 & 0.0 & 0.0 & 0.0 & 0.0 & 0.0 & 0.0 & 0.0 & 0.0 & 0.0 & 0.0 \\

\midrule

\rowcolor{line-blue}\multicolumn{13}{l}{\textbf{Proprietary Models}} \\
Seed-1.6-2025-06-15~\cite{seed2025seed1_5vl} &
17.73 & \cellcolor{secondcolor}{21.15} & \cellcolor{secondcolor}{10.64} & 12.76 & 14.73 & \cellcolor{bestcolor}{\textbf{23.14}} & 35.74 & \cellcolor{secondcolor}{41.67} & \cellcolor{secondcolor}{9.09} & 6.31 & \cellcolor{bestcolor}{\textbf{21.05}} & 11.11 \\

Gemini-2.5-pro-2025-06~\cite{team2023gemini} &
\cellcolor{secondcolor}{19.33} & 19.71 & 9.22 & \cellcolor{secondcolor}{19.34} & \cellcolor{bestcolor}{\textbf{30.23}} & 15.29 & 24.50 & \cellcolor{bestcolor}{\textbf{50.00}} & 3.03 & 8.11 & 12.28 & \cellcolor{bestcolor}{\textbf{21.21}} \\

Grok-4-2025-07-09~\cite{grok4_xai_2025} &
14.40 & 6.81 & \cellcolor{bestcolor}{\textbf{16.31}} & 12.76 & 17.83 & \cellcolor{secondcolor}{20.00} & 30.92 & 25.00 & 1.01 & \cellcolor{secondcolor}{13.51} & 14.04 & 8.42 \\

GPT-5-nano-2025-08-07~\cite{openai_gpt5_systemcard} &
6.67 & 5.38 & 4.96 & 14.40 & 6.98 & 7.45 & 13.25 & 20.83 & -3.03 & -13.51 & 10.53 & 6.40 \\

GPT-5-mini-2025-08-07~\cite{openai_gpt5_systemcard} &
15.20 & 18.28 & 9.22 & 6.17 & 22.48 & 5.88 & \cellcolor{secondcolor}{37.35} & 39.58 & \cellcolor{bestcolor}{\textbf{17.17}} & 4.50 & -1.75 & \cellcolor{secondcolor}{13.80} \\

GPT-5-2025-08-07~\cite{openai_gpt5_systemcard} &
\cellcolor{bestcolor}{\textbf{20.13}} & \cellcolor{bestcolor}{\textbf{26.88}} & 4.96 & \cellcolor{bestcolor}{\textbf{25.93}} & \cellcolor{secondcolor}{24.03} & 13.73 & \cellcolor{bestcolor}{\textbf{48.59}} & \cellcolor{secondcolor}{41.67} & -1.01 & \cellcolor{bestcolor}{\textbf{22.52}} & \cellcolor{secondcolor}{19.30} & 10.44 \\

\midrule
\rowcolor{line-blue}\multicolumn{13}{l}{\textbf{Open-source Models}} \\


Qwen2.5-VL-3B-Instruct~\cite{Qwen-VL} &
0.80 & 2.51 & 0.71 & -0.41 & -2.33 & 7.45 & 2.01 & 2.08 & \cellcolor{secondcolor}{-3.03} & -6.31 & \cellcolor{secondcolor}{14.04} & -3.03 \\

Qwen2.5-VL-7B-Instruct~\cite{Qwen-VL} &
3.47 & 6.81 & 2.13 & \cellcolor{bestcolor}{\textbf{11.11}} & 2.33 & \cellcolor{bestcolor}{\textbf{15.29}} & 6.83 & 10.42 & -5.05 & -4.50 & -1.75 & -1.01 \\

Qwen2.5-VL-72B-Instruct~\cite{Qwen-VL} &
\cellcolor{bestcolor}{\textbf{7.20}} & \cellcolor{secondcolor}{8.24} & -3.55 & \cellcolor{secondcolor}{9.47} & 3.88 & \cellcolor{secondcolor}{12.16} & 8.43 & \cellcolor{secondcolor}{25.00} & \cellcolor{bestcolor}{\textbf{11.11}} & \cellcolor{secondcolor}{0.90} & 10.53 & 3.70 \\

InternVL3-8B~\cite{zhu2025internvl3} &
3.87 & \cellcolor{secondcolor}{8.24} & 2.13 & 4.53 & \cellcolor{secondcolor}{13.18} & -3.53 & \cellcolor{secondcolor}{14.86} & 14.58 & \cellcolor{secondcolor}{-3.03} & -6.31 & 8.77 & -2.36 \\

InternVL3-78B~\cite{zhu2025internvl3} &
4.67 & 5.38 & \cellcolor{bestcolor}{\textbf{7.80}} & 6.17 & -2.33 & 1.18 & 6.83 & \cellcolor{bestcolor}{\textbf{31.25}} & -5.05 & \cellcolor{bestcolor}{\textbf{6.31}} & -7.02 & \cellcolor{secondcolor}{4.38} \\

InternVL3.5-8B~\cite{wang2025internvl3} & \cellcolor{secondcolor}{6.67} & \cellcolor{bestcolor}{\textbf{11.11}} & \cellcolor{secondcolor}{6.38} & 6.17 & 0.78 & 7.45 & 13.25 & 20.83 & -5.05 & -13.51 & \cellcolor{bestcolor}{\textbf{15.79}} &  \cellcolor{bestcolor}{\textbf{7.74}} \\

Qwen3-8B-Instruct~\cite{yang2025qwen3technicalreport} &
3.47 & 1.08 & 3.55 & 2.88 & \cellcolor{bestcolor}{\textbf{16.28}} & 5.88 & \cellcolor{bestcolor}{\textbf{19.68}} & 22.92 & -11.11 & -4.50 & 1.75 & -6.40 \\

\midrule
\rowcolor{line-blue}\multicolumn{13}{l}{\textbf{Human Evaluation}} \\
\footnotesize $\Delta$(Best Model,Human) &
\colornum{-76.14} & \colornum{-71.65} & \colornum{-80.36} & \colornum{-66.34} &
\colornum{-68.17} & \colornum{-72.06} & \colornum{-45.14} & \colornum{-48.00} &
\colornum{-80.96} & \colornum{-75.75} & \colornum{-74.95} & \colornum{-75.06} \\

Human &
\textbf{96.27} & \textbf{98.53} & \textbf{96.67} & \textbf{92.27} & \textbf{98.4} & \textbf{95.2} & \textbf{93.73} & \textbf{98.0} & \textbf{98.13} & \textbf{98.27} & \textbf{96.0} & \textbf{96.27} \\

\bottomrule
\end{tabular}
}
\vspace{-2pt}
\caption{\textbf{Evaluation on \textbf{MMSI}  (\name Protocol)}.
All scores are \textbf{\textit{CAA}} (computed via \cref{eq:caa}). We use the unified chain-of-thought prompt during evaluation, defined in \cref{appendix:eval_prompt}.
Under Positional Relationship, C: Camera; O: Object; R: Region.
}
\label{tab:mmsi_ss}
\end{table*}



%% file: tables/oo_tables/omni_oo.tex
\begin{table*}[h]
\centering
\setlength{\tabcolsep}{1.2pt}
\resizebox{1.0\linewidth}{!}{

\begin{tabular}{l c *{10}{c}}
\toprule
\multirow{4}{*}{\textbf{Models}} &
\multirow{4}{*}{\textbf{Avg.}} &
\multicolumn{2}{c}{\textbf{Dynamic Reasoning}} &
\multicolumn{3}{c}{\textbf{Spatial Interaction}} &
\multicolumn{2}{c}{\textbf{Complex Logic}} &
\multicolumn{3}{c}{\textbf{Perspective Taking}} \\
\cmidrule(lr){3-4}\cmidrule(lr){5-7}\cmidrule(lr){8-9}\cmidrule(lr){10-12}
& &
\makecell{\textbf{Manipulate}\\\textbf{}\\\textcolor{sensepurple}{\scriptsize -}} &
\makecell{\textbf{Motion}\\\textbf{Analysis}\\\textcolor{sensepurple}{\scriptsize MM,CR}} &
\makecell{\textbf{Traffic}\\\textbf{Analysis}\\\textcolor{sensepurple}{\scriptsize CR}} &
\makecell{\textbf{Locate}\\\textbf{}\\\textcolor{sensepurple}{\scriptsize -}} &
\makecell{\textbf{Geospatial}\\\textbf{Strategy}\\\textcolor{sensepurple}{\scriptsize -}} &
\makecell{\textbf{Pattern}\\\textbf{Recognition}\\\textcolor{sensepurple}{\scriptsize CR}} &
\makecell{\textbf{Geometric}\\\textbf{Reasoning}\\\textcolor{sensepurple}{\scriptsize CR}} &
\makecell{\textbf{Ego}\\\textbf{Centric}\\\textcolor{sensepurple}{\scriptsize -}} &
\makecell{\textbf{Allo}\\\textbf{Centric}\\\textcolor{sensepurple}{\scriptsize PT}} &
\makecell{\textbf{Hypothetical}\\\textbf{}\\\textcolor{sensepurple}{\scriptsize PT}} \\

\midrule
\rowcolor{line-blue}\textbf{Random Choice} &
24.98 & 24.86 & 26.3 & 35.88 & 23.43 & 27.27 & 21.44 & 24.77 & 22.55 & 24.84 & 25.78 \\

\midrule
\rowcolor{line-blue}\textbf{Proprietary Models} &&&&&&&&&& \\

Seed-1.6-2025-06-15~\cite{seed2025seed1_5vl}
& 49.32 & \cellcolor{secondcolor}{67.57} & 59.54 & 54.12 & 75.24 & 60.91 & 30.93 & 27.74 & 77.45 & 32.98 & 38.55 \\

Gemini-2.5-pro-2025-06~\cite{team2023gemini}
& 55.38 & 62.16 & \cellcolor{secondcolor}{65.90} & 62.35 & \cellcolor{bestcolor}{\textbf{78.10}} & \cellcolor{secondcolor}{68.99} & \cellcolor{bestcolor}{\textbf{44.33}} & \cellcolor{bestcolor}{\textbf{31.61}} & 83.33 & 40.43 & 42.17 \\

Grok-4-2025-07-09~\cite{grok4_xai_2025}
& 46.84 & 59.46 & 57.51 & 47.06 & 46.67 & 50.00 & 26.80$^{\dagger}$ & \cellcolor{secondcolor}{30.97} & 73.53 & 37.23 & \cellcolor{bestcolor}{\textbf{50.60}} \\

GPT-5-nano-2025-08-07~\cite{openai_gpt5_systemcard}
& 47.81 & 50.00 & 61.27 & 49.40 & 65.71 & 50.00 & 32.99 & 27.74 & 70.59 & 37.50 & 36.14 \\

GPT-5-mini-2025-08-07~\cite{openai_gpt5_systemcard}
& \cellcolor{secondcolor}{55.52} & 66.22 & 60.40 & \cellcolor{secondcolor}{65.06} & \cellcolor{secondcolor}{77.14} & 68.18 & 39.18 & 30.32 & \cellcolor{bestcolor}{\textbf{85.29}} & \cellcolor{secondcolor}{45.21} & \cellcolor{secondcolor}{48.19} \\

GPT-5-2025-08-07~\cite{openai_gpt5_systemcard}
& \cellcolor{bestcolor}{\textbf{59.90}} & \cellcolor{bestcolor}{\textbf{72.97}} & \cellcolor{bestcolor}{\textbf{70.52}} & \cellcolor{bestcolor}{\textbf{68.67}} & \cellcolor{bestcolor}{\textbf{78.10}} & \cellcolor{bestcolor}{\textbf{69.09}} & \cellcolor{secondcolor}{41.24} & \cellcolor{bestcolor}{\textbf{31.61}} & \cellcolor{secondcolor}{84.31} & \cellcolor{bestcolor}{\textbf{50.37}} & \cellcolor{secondcolor}{48.19} \\

\midrule
\rowcolor{line-blue}\textbf{Open-source Models} &&&&&&&&&& \\


Qwen2.5-VL-3B-Instruct~\cite{Qwen-VL}
& 42.47 & 58.11 & 47.40 & 47.06 & 46.67 & 49.09 & \cellcolor{secondcolor}{29.90} & 21.29 & 62.75 & 37.50 & 40.96 \\

Qwen2.5-VL-7B-Instruct~\cite{Qwen-VL}
& 39.07 & 50.00 & 32.66 & 49.41 & 49.52 & 45.45 & 26.80 & \cellcolor{bestcolor}{\textbf{32.26}} & 66.67 & 31.61 & \cellcolor{bestcolor}{\textbf{50.60}} \\

Qwen2.5-VL-72B-Instruct~\cite{Qwen-VL}
& \cellcolor{secondcolor}{47.81} & 60.81 & \cellcolor{secondcolor}{60.69} & \cellcolor{secondcolor}{55.29} & \cellcolor{bestcolor}{\textbf{66.67}} & 53.64 & 24.74 & 25.16 & \cellcolor{secondcolor}{76.47} & 34.07 & 38.55 \\

InternVL3-8B~\cite{zhu2025internvl3}
& 46.25 & \cellcolor{bestcolor}{\textbf{64.86}} & 56.65 & 49.41 & 47.62 & 47.21 & 22.68 & 21.94 & 73.53 & \cellcolor{secondcolor}{39.36} & \cellcolor{bestcolor}{\textbf{50.60}} \\

InternVL3-78B~\cite{zhu2025internvl3}
& \cellcolor{bestcolor}{\textbf{50.95}} & \cellcolor{secondcolor}{62.16} & \cellcolor{bestcolor}{\textbf{65.32}} & \cellcolor{bestcolor}{\textbf{56.47}} & 59.05 & \cellcolor{secondcolor}{54.55} & 28.87 & 28.39 & \cellcolor{bestcolor}{\textbf{77.45}} & \cellcolor{bestcolor}{\textbf{40.69}} & 42.17 \\

InternVL3.5-8B~\cite{wang2025internvl3}
& 46.71 & 60.81 & 60.12 & 42.35 & 58.10 & 51.82 & \cellcolor{bestcolor}{\textbf{32.99}} & \cellcolor{secondcolor}{29.68} & 67.65 & 33.78 & 42.17 \\

Qwen3-8B-Instruct~\cite{yang2025qwen3technicalreport} &
45.73 & \cellcolor{secondcolor}{62.16} & 58.09 & 54.12 & \cellcolor{secondcolor}{63.81} & \cellcolor{bestcolor}{\textbf{55.45}} & 14.43 & 25.81 & 68.63 & 31.91 & \cellcolor{secondcolor}{43.37} \\

\midrule
\rowcolor{line-blue}\textbf{Human Evaluation} &&&&&&&&&& \\
\footnotesize $\Delta$(Best Model,Human)
& \colornum{-32.40} & \colornum{-23.56} & \colornum{-26.78} & \colornum{-24.27}
& \colornum{-19.04} & \colornum{-25.46} & \colornum{-46.97} & \colornum{-55.37}
& \colornum{-13.73} & \colornum{-45.37} & \colornum{-43.38} \\

Human
& \textbf{92.63} & \textbf{96.53} & \textbf{97.30} & \textbf{92.94}
& \textbf{97.14} & \textbf{94.55} & \textbf{91.30} & \textbf{87.63}
& \textbf{99.02} & \textbf{95.74} & \textbf{93.98} \\

\bottomrule
\end{tabular}
}
\vspace{-2pt}
\caption{
\textbf{Evaluation on \textbf{OmniSpatial} (Official Protocol)}.
Metrics are \textbf{\textit{Acc}} following the original paper.
\textit{Prompt}: we follow \textbf{OmniSpatial} paper’s original prompt/inference protocol, which adpot \textit{Manual-CoT} (see \cref{tab:metric_prompt}). $^{\dagger}$ indicates cases where some generations were truncated without a final answer, these are counted as incorrect and reduce the score.
}
\label{tab:omnispatial_oo}
\end{table*}

%% file: tables/ss_tables/omni_ss.tex
\begin{table*}[h]
\centering
\setlength{\tabcolsep}{1.2pt}
\resizebox{1.0\linewidth}{!}{

\begin{tabular}{l c *{10}{c}}
\toprule
\multirow{4}{*}{\textbf{Models}} &
\multirow{4}{*}{\textbf{Avg.}} &
\multicolumn{2}{c}{\textbf{Dynamic Reasoning}} &
\multicolumn{3}{c}{\textbf{Spatial Interaction}} &
\multicolumn{2}{c}{\textbf{Complex Logic}} &
\multicolumn{3}{c}{\textbf{Perspective Taking}} \\
\cmidrule(lr){3-4}\cmidrule(lr){5-7}\cmidrule(lr){8-9}\cmidrule(lr){10-12}
& &
\makecell{\textbf{Manipulate}\\\textbf{}\\\textcolor{sensepurple}{\scriptsize -}} &
\makecell{\textbf{Motion}\\\textbf{Analysis}\\\textcolor{sensepurple}{\scriptsize MM,CR}} &
\makecell{\textbf{Traffic}\\\textbf{Analysis}\\\textcolor{sensepurple}{\scriptsize CR}} &
\makecell{\textbf{Locate}\\\textbf{}\\\textcolor{sensepurple}{\scriptsize -}} &
\makecell{\textbf{Geospatial}\\\textbf{Strategy}\\\textcolor{sensepurple}{\scriptsize -}} &
\makecell{\textbf{Pattern}\\\textbf{Recognition}\\\textcolor{sensepurple}{\scriptsize CR}} &
\makecell{\textbf{Geometric}\\\textbf{Reasoning}\\\textcolor{sensepurple}{\scriptsize CR}} &
\makecell{\textbf{Ego}\\\textbf{Centric}\\\textcolor{sensepurple}{\scriptsize -}} &
\makecell{\textbf{Allo}\\\textbf{Centric}\\\textcolor{sensepurple}{\scriptsize PT}} &
\makecell{\textbf{Hypothetical}\\\textbf{}\\\textcolor{sensepurple}{\scriptsize PT}} \\

\midrule
\rowcolor{line-blue}\textbf{Random Choice} &
0.0 & 0.0 & 0.0 & 0.0 & 0.0 & 0.0 & 0.0 & 0.0 & 0.0 & 0.0 & 0.0 \\

\midrule
\rowcolor{line-blue}\textbf{Proprietary Models} &&&&&&&&&& \\

Seed-1.6-2025-06-15~\cite{seed2025seed1_5vl}
& 35.18 & 54.95 & 51.83 & 43.53 & 60.63 & 47.88 & 12.03 & 9.29 & 68.63 & 15.60 & 13.25 \\

Gemini-2.5-pro-2025-06~\cite{team2023gemini}
& 41.35 & 49.55 & \cellcolor{bestcolor}{\textbf{59.92}} & \cellcolor{bestcolor}{\textbf{59.22}} & 64.44 & \cellcolor{secondcolor}{58.79} & \cellcolor{secondcolor}{17.53} & \cellcolor{bestcolor}{\textbf{17.93}} & \cellcolor{bestcolor}{\textbf{79.08}} & 18.09 & 16.47 \\

Grok-4-2025-07-09~\cite{grok4_xai_2025}
& 31.26 & 49.55 & 44.51 & 38.82 & 35.24 & 27.27 & -4.47$^{\dagger}$ & 6.70 & 67.32 & 21.63 & \cellcolor{bestcolor}{\textbf{38.96}} \\

GPT-5-nano-2025-08-07~\cite{openai_gpt5_systemcard}
& 36.05 & 40.54 & 52.22 & 41.96 & 63.17 & 41.82 & 7.90 & 10.15 & 72.55 & 21.28 & 19.68 \\

GPT-5-mini-2025-08-07~\cite{openai_gpt5_systemcard}
& \cellcolor{secondcolor}{41.79} & \cellcolor{secondcolor}{62.16} & 54.14 & \cellcolor{secondcolor}{49.80} & \cellcolor{bestcolor}{\textbf{69.52}} & \cellcolor{secondcolor}{58.79} & \cellcolor{bestcolor}{\textbf{24.40}} & 4.10 & \cellcolor{secondcolor}{75.16} & \cellcolor{secondcolor}{26.95} & 22.89 \\

GPT-5-2025-08-07~\cite{openai_gpt5_systemcard}
& \cellcolor{bestcolor}{\textbf{44.41}} & \cellcolor{bestcolor}{\textbf{65.77}} & \cellcolor{secondcolor}{58.38} & \cellcolor{bestcolor}{\textbf{59.22}} & \cellcolor{secondcolor}{65.71} & \cellcolor{bestcolor}{\textbf{60.00}} & 12.03 & \cellcolor{secondcolor}{16.20} & 71.24 & \cellcolor{bestcolor}{\textbf{28.37}} & \cellcolor{secondcolor}{32.53} \\

\midrule
\rowcolor{line-blue}\textbf{Open-source Models} &&&&&&&&&& \\


Qwen2.5-VL-3B-Instruct~\cite{Qwen-VL}
& 20.90 & 47.75 & 18.30 & 30.98 & 27.62 & 34.55 & -9.97 & 10.15 & 41.18 & \cellcolor{secondcolor}{15.96} & 24.50 \\

Qwen2.5-VL-7B-Instruct~\cite{Qwen-VL}
& 19.34 & 40.54 & 8.29 & \cellcolor{secondcolor}{41.96} & 35.24 & 28.48 & -4.47 & 4.10 & 55.56 & 13.48 & \cellcolor{bestcolor}{\textbf{29.32}} \\

Qwen2.5-VL-72B-Instruct~\cite{Qwen-VL}
& \cellcolor{secondcolor}{30.65} & 54.95 & \cellcolor{secondcolor}{48.36} & 37.25 & 45.40 & 39.39 & \cellcolor{secondcolor}{3.78} & 4.97 & \cellcolor{bestcolor}{\textbf{71.24}} & 10.64 & 18.07 \\

InternVL3-8B~\cite{zhu2025internvl3}
& 28.48 & \cellcolor{bestcolor}{\textbf{58.56}} & 40.66 & \cellcolor{bestcolor}{\textbf{45.10}} & 27.62 & \cellcolor{bestcolor}{\textbf{47.88}} & -0.34 & 4.10 & 56.86 & 12.06 & \cellcolor{secondcolor}{27.71} \\

InternVL3-78B~\cite{zhu2025internvl3}
& \cellcolor{bestcolor}{\textbf{35.18}} & \cellcolor{secondcolor}{56.76} & \cellcolor{bestcolor}{\textbf{49.52}} & 38.82 & 40.32 & \cellcolor{secondcolor}{44.24} & \cellcolor{bestcolor}{\textbf{17.53}} & \cellcolor{bestcolor}{\textbf{17.06}} & \cellcolor{bestcolor}{\textbf{71.24}} & \cellcolor{bestcolor}{\textbf{18.09}} & 21.29 \\

InternVL3.5-8B~\cite{wang2025internvl3} & 30.22 & 42.34 & 44.89 & 30.98 & \cellcolor{bestcolor}{\textbf{55.56}} & 43.03 & -0.34 & \cellcolor{secondcolor}{12.74} & 59.48 & 12.41 & 21.29 \\

Qwen3-8B-Instruct~\cite{yang2025qwen3technicalreport} &
25.52 & 51.35 & 32.95 & 37.25 & \cellcolor{secondcolor}{46.67} & 30.91 & -4.47 & 2.38 & \cellcolor{secondcolor}{58.17} & 12.77 & 21.29 \\

\midrule
\rowcolor{line-blue}\textbf{Human Evaluation} &&&&&&&&&& \\
\footnotesize $\Delta$(Best Model,Human)
& \colornum{-45.87} & \colornum{-29.61} & \colornum{-36.42} & \colornum{-29.77} & \colornum{-26.74} & \colornum{-32.51} & \colornum{-64.53} & \colornum{-65.63} & \colornum{-19.65} & \colornum{-65.96} & \colornum{-52.93} \\

Human
& \textbf{90.18} & \textbf{95.38} & \textbf{96.34} & \textbf{88.99} & \textbf{96.26} & \textbf{92.51} & \textbf{88.93} & \textbf{83.56} & \textbf{98.73} & \textbf{94.33} & \textbf{91.89} \\

\bottomrule
\end{tabular}
}
\vspace{-2pt}
\caption{
\textbf{Evaluation on \textbf{OmniSpatial} (\name Protocol)}.
All scores are \textbf{\textit{CAA}} (computed via \cref{eq:caa}).
\textit{Prompt}: we use the unified chain-of-thought prompt during evaluation defined in \cref{fig:system_prompt}. $^{\dagger}$ indicates cases where some responses were truncated without a final answer, these are counted as incorrect and reduce the score.
}
\label{tab:omnispatial_ss}
\end{table*}



%% file: tables/oo_tables/mindcube_oo.tex
\begin{table*}[ht!]
\centering
\setlength{\tabcolsep}{1.2pt}
\begin{tabular}{l *{4}{c}}
\toprule
\textbf{Models} &
\textbf{Avg.} &
\makecell{\textbf{Rotation}\\\textcolor{sensepurple}{\scriptsize PT}} &
\makecell{\textbf{Among}\\\textcolor{sensepurple}{\scriptsize PT}} &
\makecell{\textbf{Around}\\\textcolor{sensepurple}{\scriptsize PT}} \\

\midrule
\rowcolor{line-blue}\textbf{Random Choice} &
33.05 & 33.33 & 31.82 & 35.73 \\

\midrule
\rowcolor{line-blue}\textbf{Proprietary Models} & & & & \\

Seed-1.6-2025-06-15~\cite{seed2025seed1_5vl} &
48.75 & 89.00 & 36.44 & 45.60 \\

Gemini-2.5-pro-2025-06~\cite{team2023gemini} &
\cellcolor{secondcolor}{57.60} & 88.00 & \cellcolor{secondcolor}{44.92} & \cellcolor{secondcolor}{63.20} \\

Grok-4-2025-07-09~\cite{grok4_xai_2025} &
\cellcolor{bestcolor}{\textbf{63.56}} & \cellcolor{secondcolor}{93.00} & \cellcolor{bestcolor}{\textbf{54.41}} & 61.60 \\

GPT-5-nano-2025-08-07~\cite{openai_gpt5_systemcard} &
41.48 & 43.50 & 35.99 & 52.80 \\

GPT-5-mini-2025-08-07~\cite{openai_gpt5_systemcard} &
56.69 & 86.50 & 44.65 & 61.20 \\

GPT-5-2025-08-07~\cite{openai_gpt5_systemcard} &
56.30 & \cellcolor{bestcolor}{\textbf{94.50}} & 38.20 & \cellcolor{bestcolor}{\textbf{68.40}} \\

\midrule
\rowcolor{line-blue}\textbf{Open-source Models} & & & & \\


Qwen2.5-VL-3B-Instruct~\cite{Qwen-VL} &
37.60 & 33.50 & 35.93 & 44.80 \\

Qwen2.5-VL-7B-Instruct~\cite{Qwen-VL} &
36.05 & 37.00 & 32.37 & 44.00 \\

Qwen2.5-VL-72B-Instruct~\cite{Qwen-VL} &
\cellcolor{secondcolor}{42.40} & \cellcolor{bestcolor}{\textbf{44.00}} & \cellcolor{secondcolor}{39.32} & 48.40 \\

InternVL3-8B~\cite{zhu2025internvl3} &
41.54 & 36.50 & 38.14 & \cellcolor{secondcolor}{53.60} \\

InternVL3-78B~\cite{zhu2025internvl3} &
\cellcolor{bestcolor}{\textbf{49.52}} & \cellcolor{secondcolor}{38.50} & \cellcolor{bestcolor}{\textbf{48.31}} & \cellcolor{bestcolor}{\textbf{61.20}} \\

InternVL3.5-8B~\cite{wang2025internvl3} &
41.54 & 36.50 & 38.14 & \cellcolor{secondcolor}{53.60} \\

Qwen3-8B-Instruct~\cite{yang2025qwen3technicalreport}  &
29.42 & 29.50 & 28.64 & 31.20 \\

\midrule
\rowcolor{line-blue}\textbf{Human Evaluation} & & & & \\%
\footnotesize $\Delta$(Best Model,Human) & \colornum{-30.99} & - & - & - \\
Human & 94.55 & - & - & - \\

\bottomrule
\end{tabular}
\vspace{-2pt}
\caption{
\textbf{Evaluation on \textbf{MindCube-Tiny} (Official Protocol)}.
All scores are \textbf{\textit{Acc}}. We follow the MindCube paper’s original prompt/inference protocol with \textit{Raw-QA} format.
}
\label{tab:mindcube_oo}
\end{table*}

%% file: tables/ss_tables/mindcube_ss.tex
\begin{table*}[ht!]
\centering
\setlength{\tabcolsep}{1.2pt}
\begin{tabular}{l *{4}{c}}
\toprule
\textbf{Models} &
\textbf{Avg.} &
\makecell{\textbf{Rotation}\\\textcolor{sensepurple}{\scriptsize PT}} &
\makecell{\textbf{Among}\\\textcolor{sensepurple}{\scriptsize PT}} &
\makecell{\textbf{Around}\\\textcolor{sensepurple}{\scriptsize PT}} \\

\midrule
\rowcolor{line-blue}\textbf{Random Choice} &
0.0 & 0.0 & 0.0 & 0.0 \\

\midrule
\rowcolor{line-blue}\textbf{Proprietary Models} & & & & \\

Seed-1.6-2025-06-15~\cite{seed2025seed1_5vl} &
26.32 & \cellcolor{bestcolor}{\textbf{89.50}} & 10.25 & 14.11 \\

Gemini-2.5-pro-2025-06~\cite{team2023gemini} &
\cellcolor{bestcolor}{\textbf{39.53}} & 85.50 & \cellcolor{bestcolor}{\textbf{47.46}} & \cellcolor{bestcolor}{\textbf{67.20}} \\

Grok-4-2025-07-09~\cite{grok4_xai_2025} &
\cellcolor{secondcolor}{38.10} & \cellcolor{secondcolor}{88.00} & \cellcolor{secondcolor}{18.46} & 45.85 \\

GPT-5-nano-2025-08-07~\cite{openai_gpt5_systemcard} &
14.54 & 10.75 & 11.75 & 24.69 \\

GPT-5-mini-2025-08-07~\cite{openai_gpt5_systemcard} &
35.80 & 86.50 & \cellcolor{secondcolor}{18.46} & 39.63 \\

GPT-5-2025-08-07~\cite{openai_gpt5_systemcard} &
35.37 & \cellcolor{bestcolor}{\textbf{89.50}} & 9.01 & \cellcolor{secondcolor}{56.43} \\

\midrule
\rowcolor{line-blue}\textbf{Open-source Models}  & & & & \\


Qwen2.5-VL-3B-Instruct~\cite{Qwen-VL} &
9.23 & -5.75 & 8.02 & 24.69 \\

Qwen2.5-VL-7B-Instruct~\cite{Qwen-VL} &
-0.83 & -2.00 & -0.68 & -0.21 \\

Qwen2.5-VL-72B-Instruct~\cite{Qwen-VL} &
10.23 & 13.00 & 7.77 & 14.11 \\

InternVL3-8B~\cite{zhu2025internvl3} &
\cellcolor{bestcolor}{\textbf{15.98}} & 7.75 & \cellcolor{bestcolor}{\textbf{11.25}} & \cellcolor{bestcolor}{\textbf{34.65}} \\

InternVL3-78B~\cite{zhu2025internvl3} &
\cellcolor{secondcolor}{13.54} & \cellcolor{secondcolor}{16.00} & \cellcolor{bestcolor}{\textbf{11.25}} & 17.22 \\

InternVL3.5-8B~\cite{wang2025internvl3} & 12.96 & \cellcolor{bestcolor}{\textbf{22.00}} & \cellcolor{secondcolor}{10.75} & 11.00 \\

Qwen3-8B-Instruct~\cite{yang2025qwen3technicalreport}  &
8.22 & 7.0 & 1.31 & \cellcolor{secondcolor}{26.56} \\

\midrule
\rowcolor{line-blue}\textbf{Human Evaluation} & & & & \\%
\footnotesize $\Delta$(Best Model,Human) & \colornum{-52.41} & - & - & - \\
Human & 91.94 & - & - & - \\

\bottomrule
\end{tabular}
\vspace{-2pt}
\caption{
\textbf{Evaluation on \textbf{MindCube-Tiny} (\name Protocol)}.
All scores are \textbf{\textit{CAA}} (computed via \cref{eq:caa}).
\textit{Prompt}: we use the unified chain-of-thought prompt during evaluation, defined in \cref{fig:system_prompt}.
}
\label{tab:mindcube_ss}
\end{table*}


%% file: tables/oo_tables/stare_oo.tex
\begin{table*}[t]
\centering
\setlength{\tabcolsep}{1.2pt}
\resizebox{1.0\linewidth}{!}{

\begin{tabular}{l c *{8}{c} *{2}{c}}
\toprule
\multirow{3}{*}{\textbf{Models}} &
\multirow{3}{*}{\textbf{Overall}} &
\multicolumn{2}{c}{\makecell{\textbf{2D Trans.}\\\textcolor{sensepurple}{\scriptsize -}}} &
\multicolumn{2}{c}{\makecell{\textbf{3D Trans.}\\\textcolor{sensepurple}{\scriptsize CR}}} &
\multicolumn{2}{c}{\makecell{\textbf{Cube Net}\\\textcolor{sensepurple}{\scriptsize DA}}} &
\multicolumn{2}{c}{\makecell{\textbf{Tangram}\\\textcolor{sensepurple}{\scriptsize -}}} &
\multirow{2}{*}{\makecell{\textbf{Temp-}\\\textbf{oral}\\\textcolor{sensepurple}{\scriptsize PT}}} &
\multirow{2}{*}{\makecell{\textbf{Pers-}\\\textbf{pective}\\\textcolor{sensepurple}{\scriptsize PT}}} \\
\cmidrule(lr){3-4}\cmidrule(lr){5-6}\cmidrule(lr){7-8}\cmidrule(lr){9-10}
& & $\times$VSim & $\checkmark$VSim
  & $\times$VSim & $\checkmark$VSim
  & $\times$VSim & $\checkmark$VSim
  & $\times$VSim & $\checkmark$VSim
  & & \\

\midrule
\rowcolor{line-blue}\textbf{Random Choice} &
34.80 & 25.00 & 25.00 & 25.00 & 25.00 & 50.00 & 50.00 & 50.00 & 50.00 & 33.33 & 25.00 \\
  
\midrule
\rowcolor{line-blue}\textbf{Proprietary Models} & & & & & & & & & & & \\
Seed-1.6-2025-06-15~\cite{seed2025seed1_5vl}
& 46.06 & 50.08 & 54.37 & \cellcolor{secondcolor}{36.76} & \cellcolor{secondcolor}{36.27} & \cellcolor{secondcolor}{66.31} & 66.24 & 53.85 & 61.54 & 35.67 & 28.00 \\

Gemini-2.5-pro-2025-06~\cite{team2023gemini}
& 49.14 & 44.76 & 52.96 & 35.13 & 30.15 & 46.91 & \cellcolor{secondcolor}{71.07} & 61.10 & 65.19 & \cellcolor{secondcolor}{51.59} & \cellcolor{secondcolor}{39.60} \\

Grok-4-2025-07-09~\cite{grok4_xai_2025}
& 26.90 & 39.59 & 39.95 & 30.56 & 28.43 & 8.51 & 24.24 & 14.76 & 6.78 & 34.18 & 30.80 \\

GPT-5-nano-2025-08-07~\cite{openai_gpt5_systemcard}
& 46.05 & 41.47 & 42.79 & 34.31 & 25.49 & 63.16 & \cellcolor{bestcolor}{\textbf{73.29}} & \cellcolor{secondcolor}{69.96} & 63.06 & 38.64 & 30.92 \\

GPT-5-mini-2025-08-07~\cite{openai_gpt5_systemcard}
& \cellcolor{secondcolor}{52.51} & \cellcolor{secondcolor}{54.46} & \cellcolor{bestcolor}{\textbf{57.68}} & 35.95 & \cellcolor{bestcolor}{\textbf{38.97}} & \cellcolor{bestcolor}{\textbf{70.19}} & 70.59 & \cellcolor{bestcolor}{\textbf{72.21}} & \cellcolor{bestcolor}{\textbf{79.05}} & 45.01 & 30.52 \\

GPT-5-2025-08-07~\cite{openai_gpt5_systemcard}
& \cellcolor{bestcolor}{\textbf{54.59}} & \cellcolor{bestcolor}{\textbf{56.65}} & \cellcolor{secondcolor}{55.56} & \cellcolor{bestcolor}{\textbf{38.24}} & 31.62 & 63.95 & 69.92 & 61.11 & \cellcolor{secondcolor}{78.15} & \cellcolor{bestcolor}{\textbf{54.99}} & \cellcolor{bestcolor}{\textbf{44.98}} \\

\midrule
\rowcolor{line-blue}\textbf{Open-source Models} & & & & & & & & & & & \\


Qwen2.5-VL-3B-Instruct~\cite{Qwen-VL}
& 37.83 & 23.00 & 27.90 & 23.37 & 24.51 & 61.71 & \cellcolor{bestcolor}{\textbf{69.82}} & 54.19 & \cellcolor{secondcolor}{56.20} & 31.42 & 25.20 \\

Qwen2.5-VL-7B-Instruct~\cite{Qwen-VL}
& 35.03 & 28.17 & 30.73 & 30.23 & 27.45 & 53.66 & 54.24 & 54.81 & 32.11 & 28.87 & 25.60 \\

Qwen2.5-VL-72B-Instruct~\cite{Qwen-VL}
& 38.37 & 28.64 & \cellcolor{secondcolor}{36.64} & \cellcolor{secondcolor}{31.37} & 30.15 & \cellcolor{secondcolor}{66.67} & 64.77 & 47.67 & 41.11 & 32.48 & 26.40 \\

InternVL3-8B~\cite{zhu2025internvl3}
& \cellcolor{secondcolor}{41.36} & 30.99 & 34.52 & 30.07 & 26.47 & 64.73 & 62.02 & 45.95 & \cellcolor{bestcolor}{\textbf{58.72}} & \cellcolor{bestcolor}{\textbf{42.25}} & 29.20 \\

InternVL3-78B~\cite{zhu2025internvl3}
& \cellcolor{bestcolor}{\textbf{42.00}} & \cellcolor{secondcolor}{35.05} & 32.39 & \cellcolor{bestcolor}{\textbf{32.84}} & \cellcolor{secondcolor}{31.37} & \cellcolor{secondcolor}{66.67} & 65.81 & \cellcolor{secondcolor}{57.43} & 54.09 & 33.76 & \cellcolor{secondcolor}{30.40} \\

InternVL3.5-8B~\cite{wang2025internvl3}
& 40.18 & 24.41 & 34.28 & 28.76 & 28.19 & \cellcolor{bestcolor}{\textbf{67.83}} & \cellcolor{secondcolor}{65.90} & \cellcolor{bestcolor}{\textbf{57.49}} & 53.68 & \cellcolor{secondcolor}{34.82} & 26.00 \\

Qwen3-8B-Instruct~\cite{yang2025qwen3technicalreport}
& 39.76 & \cellcolor{bestcolor}{\textbf{38.03}} & \cellcolor{bestcolor}{\textbf{46.57}} & 29.90 & \cellcolor{bestcolor}{\textbf{32.60}} & 38.03 & 46.57 & 29.90 & 32.60 & 31.21 & \cellcolor{bestcolor}{\textbf{35.20}} \\

\midrule
\rowcolor{line-blue}\textbf{Human Evaluation} & & & & & & & & & & & \\
\footnotesize $\Delta$(Best Model,Human)
& \colornum{-41.91} & \colornum{-38.35} & \colornum{-39.32} & \colornum{-57.76}
& \colornum{-58.53} & \colornum{-28.81} & \colornum{-25.71} & \colornum{-15.29}
& \colornum{-14.95} & \colornum{-43.11} & \colornum{-53.42} \\

Human
& \textbf{96.50} & \textbf{95.00} & \textbf{97.00} & \textbf{96.00}
& \textbf{97.50} & \textbf{99.00} & \textbf{99.00} & \textbf{87.50}
& \textbf{94.00} & \textbf{98.10} & \textbf{98.40} \\
\bottomrule
\end{tabular}
}
\vspace{-2pt}

\caption{\textbf{Evaluation on \textbf{STARE} (Official Protocol).}
\textit{Metrics}: for the Cube Net and Tangram tasks we follow the original paper and report \textit{F1} score. For other tasks, scores are reported as \textit{Acc}; The Overall score is the \textbf{macro average} across all subsets. We follow STARE’s original prompt/inference protocol (paper or released code), which uses Zero-shot CoT (see \cref{tab:metric_prompt}).}

\label{tab:stare_oo}
\end{table*}

%% file: tables/ss_tables/stare_ss.tex
\begin{table*}[h!]
\centering
\setlength{\tabcolsep}{1.2pt}
\resizebox{\linewidth}{!}{
\begin{tabular}{@{\extracolsep{\fill}} l c *{8}{c} *{2}{c}}
\toprule
\multirow{3}{*}{\textbf{Models}} &
\multirow{3}{*}{\textbf{Overall}} &
\multicolumn{2}{c}{\makecell{\textbf{2D Trans.}\\\textcolor{sensepurple}{\scriptsize -}}} &
\multicolumn{2}{c}{\makecell{\textbf{3D Trans.}\\\textcolor{sensepurple}{\scriptsize CR}}} &
\multicolumn{2}{c}{\makecell{\textbf{Cube Net}\\\textcolor{sensepurple}{\scriptsize DA}}} &
\multicolumn{2}{c}{\makecell{\textbf{Tangram}\\\textcolor{sensepurple}{\scriptsize -}}} &
\multirow{2}{*}{\makecell{\textbf{Temp-}\\\textbf{oral}\\\textcolor{sensepurple}{\scriptsize PT}}} &
\multirow{2}{*}{\makecell{\textbf{Pers-}\\\textbf{pective}\\\textcolor{sensepurple}{\scriptsize PT}}} \\
\cmidrule(lr){3-4}\cmidrule(lr){5-6}\cmidrule(lr){7-8}\cmidrule(lr){9-10}
& & $\times$VSim & $\checkmark$VSim
  & $\times$VSim & $\checkmark$VSim
  & $\times$VSim & $\checkmark$VSim
  & $\times$VSim & $\checkmark$VSim
  & & \\

\midrule
\rowcolor{line-blue}\textbf{Random Choice} &
0.0 & 0.0 & 0.0 & 0.0 & 0.0 & 0.0 & 0.0 & 0.0 & 0.0 & 0.0 & 0.0 \\

\midrule
\rowcolor{line-blue}\textbf{Proprietary Models} & & & & & & & & & & & \\

Seed-1.6-2025-06-15~\cite{seed2025seed1_5vl}
& 20.46 & 26.34 & 34.12 & 13.29 & 12.42 & 27.46 & 5 & 29.32 & 43.25 & 6.69 & 7.2 \\

Gemini-2.5-pro-2025-06~\cite{team2023gemini}
& 26.73 & 31.35 & 31.6 & 13.73 & 14.38 & 24.35 & \cellcolor{secondcolor}{43.33} & 43.61 & \cellcolor{secondcolor}{61.94} & \cellcolor{secondcolor}{21.34} & \cellcolor{secondcolor}{12.53} \\

Grok-4-2025-07-09~\cite{grok4_xai_2025}
& 10.13 & 12.51 & 22.72 & 8.71 & 5.74 & -6.74$^{\dagger}$ & 23.33 & 13.91 & 25.26 & -4.78$^{\dagger}$ & 5.6 \\

GPT-5-nano-2025-08-07~\cite{openai_gpt5_systemcard}
& 16.23 & 19.46 & 19.94 & 8.32 & 6.54 & \cellcolor{bestcolor}{\textbf{36.79}} & 18.33 & \cellcolor{secondcolor}{46.99} & 8.65 & 6.05 & 4.69 \\

GPT-5-mini-2025-08-07~\cite{openai_gpt5_systemcard}
& \cellcolor{secondcolor}{28.56} & \cellcolor{bestcolor}{\textbf{40.11}} & \cellcolor{secondcolor}{41.69} & \cellcolor{secondcolor}{13.94} & \cellcolor{secondcolor}{16.67} & \cellcolor{secondcolor}{31.61} & 21.67 & \cellcolor{bestcolor}{\textbf{53.01}} & 34.26 & 19.75 & 8.27 \\

GPT-5-2025-08-07~\cite{openai_gpt5_systemcard}
& \cellcolor{bestcolor}{\textbf{33.46}} & \cellcolor{secondcolor}{33.44} & \cellcolor{bestcolor}{\textbf{47.99}} & \cellcolor{bestcolor}{\textbf{14.38}} & \cellcolor{bestcolor}{\textbf{18.63}} & 30.57 & \cellcolor{bestcolor}{\textbf{46.67}} & 44.36 & \cellcolor{bestcolor}{\textbf{73.7}} & \cellcolor{bestcolor}{\textbf{33.44}} & \cellcolor{bestcolor}{\textbf{30.67}} \\

\midrule
\rowcolor{line-blue}\textbf{Open-source Models} & & & & & & & & & & & \\

Qwen2.5-VL-3B-Instruct~\cite{Qwen-VL}
& -1.79 & -4.33 & 1.02 & 3.49 & -1.63 & -7.77 & -15 & -3.76 & -18.34 & -1.27 & -1.87 \\

Qwen2.5-VL-7B-Instruct~\cite{Qwen-VL}
& 5.06 & 2.76 & 4.49 & 2.40 & \cellcolor{secondcolor}{7.52} & -4.66 & \cellcolor{secondcolor}{8.33} & 16.54 & 14.88 & -0.96 & 4.53 \\

Qwen2.5-VL-72B-Instruct~\cite{Qwen-VL}
& 9.96 & \cellcolor{secondcolor}{8.40} & \cellcolor{secondcolor}{13.00} & \cellcolor{bestcolor}{\textbf{9.59}} & 3.59 & 0.52 & -1.67 & 22.18 & \cellcolor{secondcolor}{16.96} & \cellcolor{bestcolor}{\textbf{8.60}} & 8.27 \\

InternVL3-8B~\cite{zhu2025internvl3}
& 6.20 & 3.81 & 9.54 & \cellcolor{secondcolor}{8.71} & 5.88 & -2.59 & 0 & 16.54 & -5.19 & 6.37 & 1.33 \\

InternVL3-78B~\cite{zhu2025internvl3}
& \cellcolor{bestcolor}{\textbf{11.79}} & 6.73 & \cellcolor{secondcolor}{17.73} & 4.58 & \cellcolor{bestcolor}{\textbf{10.46}} & \cellcolor{bestcolor}{\textbf{11.92}} & 0 & \cellcolor{secondcolor}{36.47} & \cellcolor{bestcolor}{\textbf{17.65}} & 4.14 & \cellcolor{secondcolor}{11.47} \\

InternVL3.5-8B~\cite{wang2025internvl3} & 7.64 & -1.20 & 7.64 & 4.79 & 1.96 & \cellcolor{secondcolor}{9.84} & 0 & \cellcolor{bestcolor}{\textbf{37.22}} & \cellcolor{secondcolor}{16.96} & \cellcolor{secondcolor}{6.69} & 0.27 \\

Qwen3-8B-Instruct~\cite{yang2025qwen3technicalreport} &
\cellcolor{secondcolor}{11.67} & \cellcolor{bestcolor}{\textbf{19.25}} & \cellcolor{bestcolor}{\textbf{34.12}} & 8.28 & \cellcolor{secondcolor}{7.52} & -13.99 & \cellcolor{bestcolor}{\textbf{25.00}} & 10.90 & -8.65 & 0.96 & \cellcolor{bestcolor}{\textbf{13.07}} \\

\midrule
\rowcolor{line-blue}\textbf{Human Evaluation} & & & & & & & & & & & \\
\footnotesize $\Delta$(Best Model,Human)
& \colornum{-61.17} & \colornum{-53.22} & \colornum{-48.01} & \colornum{-80.29} & \colornum{-78.04} & \colornum{-61.21} & \colornum{-51.33} & \colornum{-21.99} & \colornum{-14.30} & \colornum{-63.71} & \colornum{-67.20} \\

Human &
\textbf{94.63} & \textbf{93.33} & \textbf{96.00} & \textbf{94.67} & \textbf{96.67} & \textbf{98.00} & \textbf{98.00} & \textbf{75.00} & \textbf{88.00} & \textbf{97.15} & \textbf{97.87} \\

\bottomrule
\end{tabular}
}
\vspace{-2pt}
\caption{\textbf{Evaluation on \textbf{STARE}  (\name Protocol)}.
All scores are \textbf{\textit{CAA}} (computed via \cref{eq:caa}).
\textit{Prompt}: we use the unified chain-of-thought prompt during evaluation defined in \cref{fig:system_prompt}. $^{\dagger}$ indicates cases where some generations were truncated without a final answer, these are counted as incorrect and reduce the score.
}
\label{tab:stare_ss}
\end{table*}

%% file: tables/oo_tables/core_oo.tex
\begin{table*}[ht!]
\centering
\setlength{\tabcolsep}{1.2pt}
\resizebox{1.0\linewidth}{!}{

\begin{tabular}{ l c *{5}{c} *{4}{c} *{3}{c}}
\toprule
\multirow{2}{*}{\textbf{Models}} &
\multirow{2}{*}{\textbf{Avg.}} &
\multicolumn{5}{c}{\textbf{Sensorimotor}} &
\multicolumn{3}{c}{\textbf{Concrete Operation}} &
\multicolumn{4}{c}{\textbf{Formal Operation}} \\
\cmidrule(lr){3-7}\cmidrule(lr){8-10}\cmidrule(lr){11-14}
& &
\makecell{Boundary\\\textcolor{sensepurple}{\scriptsize -}} & 
\makecell{Continuity\\\textcolor{sensepurple}{\scriptsize -}} & \makecell{Permanence\\\textcolor{sensepurple}{\scriptsize -}} & 
\makecell{Spatiality\\\textcolor{sensepurple}{\scriptsize SR}} & \makecell{Perceptual\\Constancy\\\textcolor{sensepurple}{\scriptsize -}} &
\makecell{Intuitive\\Physics\\\textcolor{sensepurple}{\scriptsize -}} & \makecell{Perspective\\Taking\\\textcolor{sensepurple}{\scriptsize PT}} & \makecell{Conservation\\\textcolor{sensepurple}{\scriptsize -}} &
\makecell{Hierarchical\\Relation\\\textcolor{sensepurple}{\scriptsize -}} & 
\makecell{Intentionality\\Understanding\\\textcolor{sensepurple}{\scriptsize -}} &
\makecell{Mechanical\\Reasoning\\\textcolor{sensepurple}{\scriptsize -}} & \makecell{Tool\\Using\\\textcolor{sensepurple}{\scriptsize -}} \\

\midrule
\rowcolor{line-blue}\textbf{Random Choice} &
37.70 & 38.86 & 26.45 & 40.00 & 29.36 & 49.65 & 48.33 & 40.13 & 48.39 & 30.29 & 26.72 & 36.51 & 22.40 \\

\midrule
\rowcolor{line-blue}\multicolumn{14}{l}{\textbf{Proprietary Models}} \\

Seed-1.6-2025-06-15~\cite{seed2025seed1_5vl}
& 77.17 & 81.66 & \cellcolor{bestcolor}{\textbf{72.73}} & 47.50 & 56.65 & \cellcolor{secondcolor}{91.32} & 73.33 & 49.89 & 93.67 & 80.29 & 82.35 & \cellcolor{bestcolor}{\textbf{89.63}} & \cellcolor{secondcolor}{97.81} \\

Gemini-2.5-pro-2025-06~\cite{team2023gemini}
& 76.70 & \cellcolor{secondcolor}{84.28} & 65.29 & \cellcolor{secondcolor}{65.00} & 65.63 & 81.60 & \cellcolor{bestcolor}{\textbf{78.33}} & 56.83 & 91.24 & \cellcolor{bestcolor}{\textbf{86.76}} & \cellcolor{bestcolor}{\textbf{86.03}} & \cellcolor{secondcolor}{84.23} & 79.42 \\

Grok-4-2025-07-09~\cite{grok4_xai_2025}
& \cellcolor{secondcolor}{79.27} & 75.55 & 60.74 & 56.83 & 56.83 & \cellcolor{secondcolor}{91.32} & 71.67 & \cellcolor{secondcolor}{67.25} & \cellcolor{bestcolor}{\textbf{96.77}} & 81.47 & \cellcolor{bestcolor}{\textbf{86.03}} & 83.44 & 96.72 \\

GPT-5-nano-2025-08-07~\cite{openai_gpt5_systemcard}
& 67.92 & 79.48 & 64.46 & 42.50 & 49.40 & 89.93 & 68.33 & 47.29 & 72.81 & 73.24 & 63.37 & 74.69 & 81.42 \\

GPT-5-mini-2025-08-07~\cite{openai_gpt5_systemcard}
& 77.77 & \cellcolor{bestcolor}{\textbf{86.46}} & 61.16 & 52.50 & \cellcolor{bestcolor}{\textbf{70.88}} & 90.62 & \cellcolor{secondcolor}{75.83} & 61.17 & 91.24 & 73.24 & 78.22 & 79.67 & 92.53 \\

GPT-5-2025-08-07~\cite{openai_gpt5_systemcard}
& \cellcolor{bestcolor}{\textbf{84.37}} & \cellcolor{secondcolor}{84.28} & \cellcolor{secondcolor}{69.42} & \cellcolor{bestcolor}{\textbf{70.00}} & \cellcolor{secondcolor}{69.93} & \cellcolor{bestcolor}{\textbf{92.71}} & \cellcolor{secondcolor}{75.83} & \cellcolor{bestcolor}{\textbf{80.04}} & \cellcolor{secondcolor}{96.31} & \cellcolor{secondcolor}{83.82} & \cellcolor{secondcolor}{85.15} & 83.44 & \cellcolor{bestcolor}{\textbf{99.64}} \\

\midrule
\rowcolor{line-blue}\multicolumn{14}{l}{\textbf{Open-source Models}} \\

Qwen2.5-VL-3B-Instruct~\cite{Qwen-VL}
& 60.19 & 68.56 & 57.02 & \cellcolor{secondcolor}{52.50} & 37.95 & 68.06 & 48.33 & \cellcolor{secondcolor}{31.45} & 69.12 & 55.00 & 68.07 & 61.41 & 90.35 \\

Qwen2.5-VL-7B-Instruct~\cite{Qwen-VL}
& 62.16 & 83.84 & 50.83 & \cellcolor{secondcolor}{52.50} & 41.05 & 69.79 & \cellcolor{secondcolor}{55.83} & 28.63 & \cellcolor{secondcolor}{89.40} & 60.00 & 72.30 & 58.51 & 85.06 \\

Qwen2.5-VL-72B-Instruct~\cite{Qwen-VL}
& 69.22 & \cellcolor{bestcolor}{\textbf{87.34}} & 62.40 & \cellcolor{secondcolor}{52.50} & \cellcolor{bestcolor}{\textbf{51.79}} & \cellcolor{bestcolor}{\textbf{90.28}} & 55.00 & 28.42 & \cellcolor{bestcolor}{\textbf{91.71}} & 70.29 & 74.75 & 77.18 & 88.34 \\

InternVL3-8B~\cite{zhu2025internvl3}
& 60.92 & 67.69 & 68.18 & 40.00 & 40.57 & 77.43 & 48.33 & 22.99 & 76.04 & 66.18 & 68.87 & 61.83 & 82.33 \\

InternVL3-78B~\cite{zhu2025internvl3}
& \cellcolor{bestcolor}{\textbf{71.16}} & \cellcolor{secondcolor}{86.90} & \cellcolor{bestcolor}{\textbf{72.31}} & 45.00 & 47.26 & \cellcolor{secondcolor}{88.54} & 50.00 & \cellcolor{bestcolor}{\textbf{31.89}} & 85.71 & 71.76 & \cellcolor{bestcolor}{\textbf{82.60}} & \cellcolor{secondcolor}{78.84} & \cellcolor{secondcolor}{94.72} \\

InternVL3.5-8B~\cite{wang2025internvl3}
& 66.40 & 79.91 & \cellcolor{secondcolor}{69.42} & 40.00 & 41.05 & 87.50 & \cellcolor{secondcolor}{55.83} & 23.64 & 81.11 & \cellcolor{bestcolor}{\textbf{76.18}} & 75.49 & 73.86 & 85.97 \\

Qwen3-8B--Instruct~\cite{yang2025qwen3technicalreport}
& \cellcolor{secondcolor}{69.67} & 79.04 & 55.37 & \cellcolor{bestcolor}{\textbf{57.50}} & \cellcolor{secondcolor}{48.21} & 85.76 & \cellcolor{bestcolor}{\textbf{59.17}} & 25.16 & 84.33 & \cellcolor{secondcolor}{75.00} & \cellcolor{secondcolor}{81.13} & \cellcolor{bestcolor}{\textbf{81.74}} & \cellcolor{bestcolor}{\textbf{97.63}} \\

\midrule
\rowcolor{line-blue}\multicolumn{14}{l}{\textbf{Human Evaluation}} \\
\scriptsize $\Delta$(Best Model,Human)
& \colornum{-2.61} & \colornum{1.63} & \colornum{-6.16} & \colornum{-18.1} & \colornum{-4.69} & \colornum{2.01}
& \colornum{-13.19} & \colornum{-11.95} & \colornum{7.88}
& \colornum{14.88} & \colornum{4.08} & \colornum{1.91} & \colornum{7.77} \\
Human
& \textbf{86.98} & \textbf{85.71} & \textbf{78.89} & \textbf{88.10} & \textbf{75.57} & \textbf{90.70}
& \textbf{91.52} & \textbf{91.99} & \textbf{88.89}
& \textbf{71.88} & \textbf{81.98} & \textbf{87.72} & \textbf{91.87} \\
\bottomrule
\end{tabular}}
\vspace{-2pt}
\caption{
\textbf{Evaluation on \textbf{CoreCognition} (Official Protocol)}.
All scores are \textbf{\textit{Acc}}. Results use the \textbf{soft circular strategy} defined in Section~\ref{sec:bench:eval_protocols}, which is aligned with original paper. A potential misalignment may occur as human scores are measured by non-circular accuracy, while model scores are based on soft-circular accuracy.
\textit{Prompt}: we follow the CoreCognition paper’s prompt protocol with CoreCognition instruction format.
}
\label{tab:corecognition_oo}
\end{table*}

%% file: tables/ss_tables/core_ss.tex
\begin{table*}[ht!]
\centering
\setlength{\tabcolsep}{1.2pt}
\resizebox{1.0\linewidth}{!}{

\begin{tabular}{ l c *{5}{c} *{4}{c} *{3}{c}}
\toprule
\multirow{2}{*}{\textbf{Models}} &
\multirow{2}{*}{\textbf{Avg.}} &
\multicolumn{5}{c}{\textbf{Sensorimotor}} &
\multicolumn{3}{c}{\textbf{Concrete Operation}} &
\multicolumn{4}{c}{\textbf{Formal Operation}} \\
\cmidrule(lr){3-7}\cmidrule(lr){8-10}\cmidrule(lr){11-14}
& &
\makecell{Boundary\\\textcolor{sensepurple}{\scriptsize -}} & 
\makecell{Continuity\\\textcolor{sensepurple}{\scriptsize -}} & \makecell{Permanence\\\textcolor{sensepurple}{\scriptsize -}} & 
\makecell{Spatiality\\\textcolor{sensepurple}{\scriptsize SR}} & \makecell{Perceptual\\Constancy\\\textcolor{sensepurple}{\scriptsize -}} &
\makecell{Intuitive\\Physics\\\textcolor{sensepurple}{\scriptsize -}} & \makecell{Perspective\\Taking\\\textcolor{sensepurple}{\scriptsize PT}} & \makecell{Conservation\\\textcolor{sensepurple}{\scriptsize -}} &
\makecell{Hierarchical\\Relation\\\textcolor{sensepurple}{\scriptsize -}} & 
\makecell{Intentionality\\Understanding\\\textcolor{sensepurple}{\scriptsize -}} &
\makecell{Mechanical\\Reasoning\\\textcolor{sensepurple}{\scriptsize -}} & \makecell{Tool\\Using\\\textcolor{sensepurple}{\scriptsize -}} \\

\midrule
\rowcolor{line-blue}\textbf{Random Choice} &
0.0 & 0.0 & 0.0 & 0.0 & 0.0 & 0.0 & 0.0 & 0.0 & 0.0 & 0.0 & 0.0 & 0.0 & 0.0 \\

\midrule

\rowcolor{line-blue}\multicolumn{14}{l}{\textbf{Proprietary Models}} \\
Seed-1.6-2025-06-15~\cite{seed2025seed1_5vl} &
58.39 & 72.14 & \cellcolor{secondcolor}{62.36} & 25.00 & 33.45 & 80.00 & 35.48 & 16.67 & 87.34 & 70.89 & 74.25 & 77.12 & 95.54 \\

Gemini-2.5-pro-2025-06~\cite{team2023gemini} &
\cellcolor{secondcolor}{73.21} & \cellcolor{bestcolor}{\textbf{79.29}} & \cellcolor{bestcolor}{\textbf{62.92}} & 25.00 & \cellcolor{secondcolor}{57.09} & \cellcolor{secondcolor}{85.52} & \cellcolor{secondcolor}{58.06} & 13.04 & \cellcolor{bestcolor}{\textbf{94.64}} & \cellcolor{bestcolor}{\textbf{81.43}} & \cellcolor{bestcolor}{\textbf{87.63}} & \cellcolor{bestcolor}{\textbf{90.85}} & \cellcolor{bestcolor}{\textbf{99.77}} \\

Grok-4-2025-07-09~\cite{grok4_xai_2025} &
69.93 & 75.00 & 49.44 & 29.17 & 36.15 & \cellcolor{bestcolor}{\textbf{86.21}} & 46.77 & \cellcolor{secondcolor}{49.28} & 83.93 & 73.00 & 81.27 & 75.82 & 98.36 \\

GPT-5-nano-2025-08-07~\cite{openai_gpt5_systemcard} &
56.13 & 70.71 & 57.87 & -8.33 & 31.42 & 76.55 & 40.32 & 5.80 & 43.75 & 56.96 & 55.85 & 76.47 & 95.07 \\

GPT-5-mini-2025-08-07~\cite{openai_gpt5_systemcard} &
68.61 & \cellcolor{secondcolor}{76.43} & 49.44 & \cellcolor{secondcolor}{33.33} & 56.42 & \cellcolor{secondcolor}{85.52} & 46.77 & 31.88 & 72.32 & 62.45 & 74.92 & \cellcolor{secondcolor}{83.01} & 98.59 \\

GPT-5-2025-08-07~\cite{openai_gpt5_systemcard} &
\cellcolor{bestcolor}{\textbf{78.45}} & \cellcolor{secondcolor}{76.43} & 61.80 & \cellcolor{bestcolor}{\textbf{37.50}} & \cellcolor{bestcolor}{\textbf{63.18}} & \cellcolor{bestcolor}{\textbf{86.21}} & \cellcolor{bestcolor}{\textbf{75.81}} & \cellcolor{bestcolor}{\textbf{63.41}} & \cellcolor{secondcolor}{89.29} & \cellcolor{secondcolor}{77.64} & \cellcolor{secondcolor}{82.94} & 82.35 & \cellcolor{secondcolor}{99.53} \\

\midrule
\rowcolor{line-blue}\multicolumn{14}{l}{\textbf{Open-source Models}} \\


Qwen2.5-VL-3B-Instruct~\cite{Qwen-VL} &
36.97 & 62.14 & 38.76 & -16.67 & 7.09 & 20.69 & -3.23 & -20.29 & 37.50 & 37.97 & 52.17 & 28.76 & 91.78 \\

Qwen2.5-VL-7B-Instruct~\cite{Qwen-VL} &
47.70 & \cellcolor{secondcolor}{75.71} & 43.26 & \cellcolor{bestcolor}{\textbf{16.67}} & 10.47 & 51.72 & 0.00 & \cellcolor{bestcolor}{\textbf{-17.39}} & \cellcolor{bestcolor}{\textbf{79.46}} & 48.95 & 63.88 & 45.75 & 96.01 \\

Qwen2.5-VL-72B-Instruct~\cite{Qwen-VL} &
\cellcolor{secondcolor}{55.41} & 73.57 & 52.81 & \cellcolor{bestcolor}{\textbf{16.67}} & \cellcolor{secondcolor}{27.03} & \cellcolor{secondcolor}{78.62} & \cellcolor{secondcolor}{19.35} & -22.46 & \cellcolor{secondcolor}{70.54} & 56.12 & \cellcolor{secondcolor}{70.23} & \cellcolor{secondcolor}{74.51} & \cellcolor{bestcolor}{\textbf{98.59}} \\

InternVL3-8B~\cite{zhu2025internvl3} &
45.66 & 61.43 & 51.69 & \cellcolor{secondcolor}{8.33} & 13.18 & 53.79 & 8.06 & -21.74 & 34.82 & 51.48 & 69.90 & 47.06 & 91.08 \\

InternVL3-78B~\cite{zhu2025internvl3} &
\cellcolor{bestcolor}{\textbf{58.01}} & \cellcolor{bestcolor}{\textbf{81.43}} & \cellcolor{bestcolor}{\textbf{67.42}} & -8.33 & \cellcolor{bestcolor}{\textbf{31.42}} & \cellcolor{bestcolor}{\textbf{82.76}} & \cellcolor{secondcolor}{19.35} & \cellcolor{secondcolor}{-19.57} & 55.36 & \cellcolor{secondcolor}{64.98} & \cellcolor{bestcolor}{\textbf{74.58}} & 67.97 & \cellcolor{secondcolor}{97.65} \\

InternVL3.5-8B~\cite{wang2025internvl3} & 49.45 & 67.86 & \cellcolor{secondcolor}{55.06} & \cellcolor{secondcolor}{8.33} & 18.58 & 68.28 & 9.68 & -28.26 & 51.79 & 63.71 & 60.54 & 63.40 & 93.19 \\

Qwen3-8B--Instruct~\cite{yang2025qwen3technicalreport} & 
47.30 & 65.22 & 45.95 & -11.11 & 12.84 & 70.77 & \cellcolor{bestcolor}{\textbf{22.47}} & -33.85 & 60.87 & \cellcolor{bestcolor}{\textbf{66.28}} & 69.68 & \cellcolor{bestcolor}{\textbf{77.46}} & \cellcolor{secondcolor}{97.90} \\

\midrule
\rowcolor{line-blue}\multicolumn{14}{l}{\textbf{Human Evaluation}} \\
\scriptsize $\Delta$(Best Model,Human) &
\colornum{-0.65} & \colornum{2.66} & \colornum{-8.38} & \colornum{-42.67} & \colornum{-2.24} & \colornum{4.68} & \colornum{-7.78} & \colornum{-23.21} & \colornum{16.17} & \colornum{21.77} & \colornum{12.22} & \colornum{10.19} & \colornum{10.25} \\
Human &
\textbf{79.10} & \textbf{76.63} & \textbf{71.30} & \textbf{80.17} & \textbf{65.42} & \textbf{81.53} & \textbf{83.59} & \textbf{86.62} & \textbf{78.47} & \textbf{59.66} & \textbf{75.41} & \textbf{80.66} & \textbf{89.52} \\

\bottomrule
\end{tabular}}
\vspace{-2pt}
\caption{
\textbf{Evaluation on \textbf{CoreCognition} (\name Protocol)}.
All scores are \textbf{\textit{CAA}} (computed via \cref{eq:caa}). Results use the \textbf{soft circular scoring} defined in Section~\ref{sec:bench:eval_protocols}. 
\textit{Prompt}: we use the unified chain-of-thought prompt during evaluation defined in \cref{fig:system_prompt}.
}
\label{tab:corecognition_ss}
\end{table*}





%% file: tables/oo_tables/spatialviz_oo.tex
\begin{table*}[ht!]
\centering
\setlength{\tabcolsep}{1.2pt}
\resizebox{1.0\linewidth}{!}{
\begin{tabular}{l c *{4}{c} *{4}{c} *{4}{c} *{4}{c}}
\toprule
\multirow{3}{*}{\textbf{Models}} &
\multirow{3}{*}{\textbf{Avg.}} &
\multicolumn{4}{c}{\textbf{Mental Rotation}} &
\multicolumn{4}{c}{\textbf{Mental Folding}} &
\multicolumn{4}{c}{\textbf{Visual Penetration}} &
\multicolumn{4}{c}{\textbf{Mental Animation}} \\
\cmidrule(lr){3-6}\cmidrule(lr){7-10}\cmidrule(lr){11-14}\cmidrule(lr){15-18}
& 
& \makecell{2DR\\\textcolor{sensepurple}{\scriptsize -}}
& \makecell{3DR\\\textcolor{sensepurple}{\scriptsize MR}}
& \makecell{3VP\\\textcolor{sensepurple}{\scriptsize MR}}
& \makecell{Avg\\\textcolor{sensepurple}{\scriptsize -}}
& \makecell{PF\\\textcolor{sensepurple}{\scriptsize DA}}
& \makecell{CU\\\textcolor{sensepurple}{\scriptsize DA}}
& \makecell{CR\\\textcolor{sensepurple}{\scriptsize DA}}
& \makecell{Avg\\\textcolor{sensepurple}{\scriptsize -}}
& \makecell{CS\\\textcolor{sensepurple}{\scriptsize DA}}
& \makecell{CC\\\textcolor{sensepurple}{\scriptsize -}}
& \makecell{CA\\\textcolor{sensepurple}{\scriptsize DA}}
& \makecell{Avg\\\textcolor{sensepurple}{\scriptsize -}}
& \makecell{AM\\\textcolor{sensepurple}{\scriptsize -}}
& \makecell{BM\\\textcolor{sensepurple}{\scriptsize SR}}
& \makecell{MS\\\textcolor{sensepurple}{\scriptsize CR}}
& \makecell{Avg\\\textcolor{sensepurple}{\scriptsize -}} \\

\midrule
\rowcolor{line-blue}\textbf{Random Choice} &
25.0 & 25.0 & 25.0 & 25.0 & 25.0 & 25.0 & 25.0 & 25.0 & 25.0 & 25.0 & 25.0 & 25.0 & 25.0 & 25.0 & 25.0 & 25.0 & 25.0\\

\midrule
\rowcolor{line-blue}\textbf{Proprietary Models} &
 &  &  &  &  &  &  &  &  &  &  &  &  &  &  &  & \\

Seed-1.6-2025-06-15~\cite{seed2025seed1_5vl}
& 34.58 & 30.00 & 15.00 & 32.00 & 26.15 & \cellcolor{secondcolor}{28.33} & \cellcolor{bestcolor}{\textbf{37.50}} & 28.33 & 31.39 & 43.33 & 50.83 & 6.25 & 36.88 & 68.75 & 18.75 & 48.75 & 45.42 \\

Gemini-2.5-pro-2025-06~\cite{team2023gemini}
& 42.71 & 53.75 & 28.75 & \cellcolor{secondcolor}{47.00} & 43.46 & 26.67 & 25.83 & 31.39 & \cellcolor{bestcolor}{\textbf{37.50}} & \cellcolor{bestcolor}{\textbf{61.67}} & \cellcolor{secondcolor}{58.33} & 32.50 & 45.31 & 83.75 & 30.00 & 52.50 & 55.42 \\

Grok-4-2025-07-09$^{\dagger}$~\cite{grok4_xai_2025}
& 19.40 & 17.50 & 22.50 & 12.12 & 16.99 & 7.63 & 24.17 & 21.67 & 17.88 & 15.97 & 8.70 & 5.00 & 10.51 & 46.25 & 5.06 & \cellcolor{bestcolor}{\textbf{56.25}} & 35.98 \\

GPT-5-nano-2025-08-07~\cite{openai_gpt5_systemcard}
& 35.59 & 52.50 & 28.75 & 41.00 & 40.77 & 12.50 & 26.67 & \cellcolor{secondcolor}{31.67} & 23.61 & 23.33 & 39.17 & \cellcolor{secondcolor}{37.50} & 32.81 & 62.50 & \cellcolor{secondcolor}{41.25} & 51.25 & 51.67 \\

GPT-5-mini-2025-08-07~\cite{openai_gpt5_systemcard}
& \cellcolor{secondcolor}{44.66} & \cellcolor{secondcolor}{82.50} & \cellcolor{bestcolor}{\textbf{33.75}} & 44.00 & \cellcolor{secondcolor}{52.69} & 15.83 & 27.50 & \cellcolor{bestcolor}{\textbf{37.50}} & 26.94 & 40.83 & 55.00 & \cellcolor{bestcolor}{\textbf{40.00}} & \cellcolor{secondcolor}{45.94} & \cellcolor{secondcolor}{91.25} & 36.25 & \cellcolor{secondcolor}{55.00} & \cellcolor{secondcolor}{60.83} \\

GPT-5-2025-08-07~\cite{openai_gpt5_systemcard}
& \cellcolor{bestcolor}{\textbf{51.27}} & \cellcolor{bestcolor}{\textbf{91.25}} & \cellcolor{secondcolor}{30.00} & \cellcolor{bestcolor}{\textbf{57.00}} & \cellcolor{bestcolor}{\textbf{59.23}} & \cellcolor{bestcolor}{\textbf{52.50}} & \cellcolor{secondcolor}{32.50} & 26.67 & \cellcolor{secondcolor}{37.22} & \cellcolor{secondcolor}{45.83} & \cellcolor{bestcolor}{\textbf{66.67}} & 36.25 & \cellcolor{bestcolor}{\textbf{51.25}} & \cellcolor{bestcolor}{\textbf{96.25}} & \cellcolor{bestcolor}{\textbf{43.75}} & 51.25 & \cellcolor{bestcolor}{\textbf{63.75}} \\

\midrule
\rowcolor{line-blue}\textbf{Open-source Models} &
 &  &  &  &  &  &  &  &  &  &  &  &  &  &  &  & \\


Qwen2.5-VL-3B-Instruct~\cite{Qwen-VL}
& 21.86 & \cellcolor{secondcolor}{27.50} & 13.75 & 30.00 & 24.23 & 18.33 & 17.50 & 25.00 & 20.28 & 15.00 & 20.00 & 7.50 & 15.00 & 16.25 & 35.00 & 41.25 & 30.83 \\

Qwen2.5-VL-7B-Instruct~\cite{Qwen-VL}
& 26.78 & \cellcolor{bestcolor}{\textbf{28.75}} & 17.50 & 31.00 & 26.15 & \cellcolor{bestcolor}{\textbf{38.33}} & \cellcolor{bestcolor}{\textbf{25.83}} & 21.67 & \cellcolor{bestcolor}{\textbf{28.61}} & 10.00 & 32.50 & 27.50 & 22.81 & 16.25 & 28.75 & \cellcolor{secondcolor}{45.00} & 30.00 \\

Qwen2.5-VL-72B-Instruct~\cite{Qwen-VL}
& \cellcolor{bestcolor}{\textbf{32.54}} & 17.50 & \cellcolor{secondcolor}{30.00} & \cellcolor{secondcolor}{38.00} & \cellcolor{bestcolor}{\textbf{29.23}} & 22.50 & 18.33 & 30.00 & 23.61 & \cellcolor{bestcolor}{\textbf{28.33}} & \cellcolor{bestcolor}{\textbf{45.83}} & \cellcolor{bestcolor}{\textbf{43.75}} & \cellcolor{bestcolor}{\textbf{38.75}} & \cellcolor{secondcolor}{32.50} & \cellcolor{secondcolor}{36.25} & \cellcolor{bestcolor}{\textbf{55.00}} & \cellcolor{bestcolor}{\textbf{41.25}} \\

InternVL3-8B~\cite{zhu2025internvl3}
& 30.00 & 26.25 & \cellcolor{bestcolor}{\textbf{31.25}} & 29.00 & \cellcolor{secondcolor}{28.85} & 20.00 & \cellcolor{secondcolor}{21.67} & \cellcolor{secondcolor}{30.83} & 24.17 & 20.83 & \cellcolor{secondcolor}{35.00} & \cellcolor{secondcolor}{40.00} & 28.75 & \cellcolor{bestcolor}{\textbf{35.00}} & \cellcolor{bestcolor}{\textbf{52.50}} & 38.75 & \cellcolor{secondcolor}{38.75} \\

InternVL3-78B~\cite{zhu2025internvl3}
& \cellcolor{secondcolor}{31.10} & 26.25 & 20.00 & \cellcolor{bestcolor}{\textbf{39.00}} & \cellcolor{bestcolor}{\textbf{29.23}} & \cellcolor{secondcolor}{23.33} & \cellcolor{secondcolor}{21.67} & \cellcolor{bestcolor}{\textbf{35.00}} & \cellcolor{secondcolor}{26.67} & \cellcolor{secondcolor}{26.67} & \cellcolor{bestcolor}{\textbf{45.83}} & 37.50 & \cellcolor{secondcolor}{34.38} & 28.75 & 32.50 & \cellcolor{secondcolor}{45.00} & 35.42 \\

InternVL3.5-8B~\cite{wang2025internvl3}
& 23.98 & \cellcolor{bestcolor}{\textbf{28.75}} & 27.50 & 25.00 & 26.92 & 10.83 & \cellcolor{secondcolor}{21.67} & 19.17 & 17.22 & 20.83 & 20.83 & 27.50 & 22.50 & 27.50 & 28.75 & 42.50 & 32.92 \\

Qwen3-8B-Instruct$^{\dagger}$~\cite{yang2025qwen3technicalreport}
& 17.54 & 16.25 & 15.00 & 29.00 & 20.77 & 4.17 & 14.17 & 19.17 & 12.50 & 24.17 & 22.50 & 1.25 & 17.81 & 16.25 & 6.25 & 41.25 & 21.25 \\

\midrule
\rowcolor{line-blue}\textbf{Human Evaluation} &
 &  &  &  &  &  &  &  &  &  &  &  &  &  &  &  & \\
\scriptsize $\Delta$(Best Model,Human)
& \colornum{-31.19} & \colornum{1.25} & \colornum{-45.41} & \colornum{-30.5} & \colornum{-26.33}
& \colornum{-41.25} & \colornum{-37.5} & \colornum{-35.42} & \colornum{-43.06}
& \colornum{-11.25} & \colornum{-4.16} & \colornum{-38.75} & \colornum{-24.17}
& \colornum{6.25} & \colornum{-35.00} & \colornum{-31.25} & \colornum{-24.58} \\
Human
& \textbf{82.46} & \textbf{90.00} & \textbf{79.16} & \textbf{87.50} & \textbf{85.56}
& \textbf{93.75} & \textbf{75.00} & \textbf{72.92} & \textbf{80.56}
& \textbf{72.92} & \textbf{70.83} & \textbf{82.50} & \textbf{75.42}
& \textbf{90.00} & \textbf{87.50} & \textbf{87.50} & \textbf{88.33} \\
\bottomrule
\end{tabular}
}
\vspace{-2pt}
\caption{
\textbf{Evaluation on \textbf{SpatialViz} (Official Protocol)}.
All values are \textbf{\textit{Acc}}.
$^{\dagger}$ indicates cases where generations were truncated due to overlong chains of thought, yielding no final answer; such instances are counted as incorrect, which depresses the score.
\textit{Prompt}: we follow the SpatialViz paper's original CoT-style prompt and inference protocol. 
}
\label{tab:spatialviz_oo}
\end{table*}

%% file: tables/ss_tables/spatialviz_ss.tex
\begin{table*}[ht!]
\centering
\setlength{\tabcolsep}{1.2pt}
\resizebox{1.0\linewidth}{!}{
\begin{tabular}{l c *{4}{c} *{4}{c} *{4}{c} *{4}{c}}
\toprule
\multirow{3}{*}{\textbf{Models}} &
\multirow{3}{*}{\textbf{Avg.}} &
\multicolumn{4}{c}{\textbf{Mental Rotation}} &
\multicolumn{4}{c}{\textbf{Mental Folding}} &
\multicolumn{4}{c}{\textbf{Visual Penetration}} &
\multicolumn{4}{c}{\textbf{Mental Animation}} \\
\cmidrule(lr){3-6}\cmidrule(lr){7-10}\cmidrule(lr){11-14}\cmidrule(lr){15-18}
& 
& \makecell{2DR\\\textcolor{sensepurple}{\scriptsize -}}
& \makecell{3DR\\\textcolor{sensepurple}{\scriptsize MR}}
& \makecell{3VP\\\textcolor{sensepurple}{\scriptsize MR}}
& \makecell{Avg\\\textcolor{sensepurple}{\scriptsize -}}
& \makecell{PF\\\textcolor{sensepurple}{\scriptsize DA}}
& \makecell{CU\\\textcolor{sensepurple}{\scriptsize DA}}
& \makecell{CR\\\textcolor{sensepurple}{\scriptsize DA}}
& \makecell{Avg\\\textcolor{sensepurple}{\scriptsize -}}
& \makecell{CS\\\textcolor{sensepurple}{\scriptsize DA}}
& \makecell{CC\\\textcolor{sensepurple}{\scriptsize -}}
& \makecell{CA\\\textcolor{sensepurple}{\scriptsize DA}}
& \makecell{Avg\\\textcolor{sensepurple}{\scriptsize -}}
& \makecell{AM\\\textcolor{sensepurple}{\scriptsize -}}
& \makecell{BM\\\textcolor{sensepurple}{\scriptsize SR}}
& \makecell{MS\\\textcolor{sensepurple}{\scriptsize CR}}
& \makecell{Avg\\\textcolor{sensepurple}{\scriptsize -}} \\

\midrule
\rowcolor{line-blue}\textbf{Random Choice} &
0.0 & 0.0 & 0.0 & 0.0 & 0.0 & 0.0 & 0.0 & 0.0 & 0.0 & 0.0 & 0.0 & 0.0 & 0.0 & 0.0 & 0.0 & 0.0 & 0.0\\

\midrule
\rowcolor{line-blue}\textbf{Proprietary Models} &
 &  &  &  &  &  &  &  &  &  &  &  &  &  &  &  & \\

Seed-1.6-2025-06-15~\cite{seed2025seed1_5vl} &
13.22 & 0.00 & -11.67 & 29.33 & 7.69 & 11.11 & \cellcolor{secondcolor}{6.67} & 7.78 & 8.52 & 16.67 & 26.67 & -11.67 & 13.33 & 55.00 & -8.33 & 31.67 & 26.11 \\

Gemini-2.5-pro-2025-06~\cite{team2023gemini} &
\cellcolor{secondcolor}{30.05} & 58.33 & \cellcolor{secondcolor}{8.33} & \cellcolor{secondcolor}{33.33} & 33.33 & \cellcolor{secondcolor}{24.44} & 2.22 & \cellcolor{secondcolor}{11.11} & \cellcolor{secondcolor}{12.59} & \cellcolor{bestcolor}{\textbf{27.78}} & \cellcolor{secondcolor}{52.22} & 20.00 & \cellcolor{bestcolor}{\textbf{35.00}} & 80.00 & \cellcolor{secondcolor}{16.67} & \cellcolor{bestcolor}{\textbf{41.67}} & \cellcolor{secondcolor}{46.11} \\

Grok-4-2025-07-09$^{\dagger}$~\cite{grok4_xai_2025} &
-6.64 & -8.33 & -3.33 & -6.40 & -6.05 & -27.78 & -3.33 & -1.11 & -10.74 & -12.22 & -24.44 & -15.00 & -17.50 & 20.00 & -21.67 & \cellcolor{bestcolor}{\textbf{41.67}} & 13.33 \\

GPT-5-nano-2025-08-07~\cite{openai_gpt5_systemcard} &
13.22 & 30.00 & -6.67 & 16.00 & 13.33 & -13.33 & 3.33 & 2.22 & -2.59 & -1.11 & 21.11 & \cellcolor{secondcolor}{26.67} & 14.17 & 66.67 & 5.00 & \cellcolor{secondcolor}{35.00} & 35.56 \\

GPT-5-mini-2025-08-07~\cite{openai_gpt5_systemcard} &
28.25 & \cellcolor{secondcolor}{75.00} & \cellcolor{bestcolor}{\textbf{11.67}} & 30.67 & \cellcolor{bestcolor}{\textbf{38.46}} & -7.78 & 5.56 & \cellcolor{bestcolor}{\textbf{20.00}} & 5.93 & 23.33 & 42.22 & \cellcolor{bestcolor}{\textbf{35.00}} & 33.33 & \cellcolor{secondcolor}{90.00} & 10.00 & 31.67 & 43.89 \\

GPT-5-2025-08-07~\cite{openai_gpt5_systemcard} &
\cellcolor{bestcolor}{\textbf{34.80}} & \cellcolor{bestcolor}{\textbf{83.33}} & -8.33 & \cellcolor{bestcolor}{\textbf{37.33}} & \cellcolor{secondcolor}{37.44} & \cellcolor{bestcolor}{\textbf{43.33}} & \cellcolor{bestcolor}{\textbf{15.56}} & -1.11 & \cellcolor{bestcolor}{\textbf{19.26}} & \cellcolor{secondcolor}{26.67} & \cellcolor{bestcolor}{\textbf{60.00}} & 8.33 & \cellcolor{secondcolor}{34.58} & \cellcolor{bestcolor}{\textbf{98.33}} & \cellcolor{bestcolor}{\textbf{33.33}} & \cellcolor{secondcolor}{35.00} & \cellcolor{bestcolor}{\textbf{55.56}} \\

\midrule
\rowcolor{line-blue}\textbf{Open-source Models} &
 &  &  &  &  &  &  &  &  &  &  &  &  &  &  &  & \\


Qwen2.5-VL-3B-Instruct~\cite{Qwen-VL}
& -3.39 & -10.00 & \cellcolor{secondcolor}{-3.33} & -2.67 & -5.13 & -1.11 & -14.44 & -14.44 & -10.00 & -4.44 & -3.33 & 5.00 & -1.67 & -10.00 & 11.67 & 16.67 & 6.11 \\

Qwen2.5-VL-7B-Instruct~\cite{Qwen-VL}
& 3.62 & -6.67 & -20.00 & \cellcolor{bestcolor}{\textbf{16.00}} & -2.05 & \cellcolor{bestcolor}{\textbf{12.22}} & \cellcolor{bestcolor}{\textbf{4.44}} & \cellcolor{secondcolor}{4.44} & \cellcolor{bestcolor}{\textbf{7.04}} & -12.22 & 6.67 & 10.00 & 0.42 & -3.33 & 5.00 & 25.00 & 8.89 \\

Qwen2.5-VL-72B-Instruct~\cite{Qwen-VL}
& \cellcolor{secondcolor}{6.44} & 3.33 & \cellcolor{bestcolor}{\textbf{10.00}} & 9.33 & \cellcolor{secondcolor}{7.69} & \cellcolor{secondcolor}{1.11} & -10.00 & -3.33 & -4.07 & -6.67 & 18.89 & \cellcolor{secondcolor}{28.33} & 11.67 & \cellcolor{secondcolor}{6.67} & 6.67 & 28.33 & 13.89 \\

InternVL3-8B~\cite{zhu2025internvl3}
& 6.10 & -6.67 & \cellcolor{bestcolor}{\textbf{10.00}} & 2.67 & 2.05 & -11.11 & -12.22 & \cellcolor{bestcolor}{\textbf{7.78}} & -5.19 & -4.44 & \cellcolor{bestcolor}{\textbf{26.67}} & 20.00 & \cellcolor{secondcolor}{13.33} & 0.00 & \cellcolor{bestcolor}{\textbf{18.33}} & \cellcolor{bestcolor}{\textbf{35.00}} & \cellcolor{bestcolor}{\textbf{17.78}} \\

InternVL3-78B~\cite{zhu2025internvl3}
& \cellcolor{bestcolor}{\textbf{9.49}} & \cellcolor{bestcolor}{\textbf{11.67}} & \cellcolor{secondcolor}{-3.33} & \cellcolor{secondcolor}{14.67} & \cellcolor{bestcolor}{\textbf{8.21}} & -4.44 & -13.33 & \cellcolor{bestcolor}{\textbf{7.78}} & \cellcolor{secondcolor}{-3.33} & \cellcolor{secondcolor}{2.22} & \cellcolor{secondcolor}{24.44} & \cellcolor{bestcolor}{\textbf{40.00}} & \cellcolor{bestcolor}{\textbf{20.00}} & \cellcolor{bestcolor}{\textbf{8.33}} & \cellcolor{secondcolor}{13.33} & 26.67 & \cellcolor{secondcolor}{16.11} \\

InternVL3.5-8B~\cite{wang2025internvl3} & 2.37 & \cellcolor{secondcolor}{6.67} & \cellcolor{bestcolor}{\textbf{10.00}} & 0.00 & 5.13 & -8.89 & \cellcolor{secondcolor}{-7.78} & 0.00 & -5.56 & \cellcolor{bestcolor}{\textbf{11.11}} & -4.44 & 10.00 & 5.00 & -6.67 & 0.00 & \cellcolor{secondcolor}{30.00} & 7.78 \\

Qwen3-8B-Instruct$^{\dagger}$~\cite{yang2025qwen3technicalreport}
& -10.40 & -15.00 & -13.33 & 2.67 & -7.69 & -27.78 & -15.56 & -15.56 & -19.63 & -1.11 & -2.22 & -26.67 & -7.92 & -6.67 & -18.33 & 16.67 & -2.78 \\

\midrule
\rowcolor{line-blue}\textbf{Human Evaluation} &
 &  &  &  &  &  &  &  &  &  &  &  &  &  &  &  & \\
\scriptsize $\Delta$(Best Model,Human)
& \colornum{-41.79} & \colornum{-3.32} & \colornum{-60.51} & \colornum{-45.99} & \colornum{-42.27}
& \colornum{-48.33} & \colornum{-51.07} & \colornum{-43.86} & \colornum{-54.79}
& \colornum{-36.07} & \colornum{-1.07} & \colornum{-41.64} & \colornum{-32.19}
& \colornum{11.68} & \colornum{-49.99} & \colornum{-41.65} & \colornum{-28.86} \\
Human
& \textbf{76.59} & \textbf{86.65} & \textbf{72.18} & \textbf{83.32} & \textbf{80.73}
& \textbf{91.66} & \textbf{66.63} & \textbf{63.86} & \textbf{74.05}
& \textbf{63.85} & \textbf{61.07} & \textbf{76.64} & \textbf{67.19}
& \textbf{86.65} & \textbf{83.32} & \textbf{83.32} & \textbf{84.42} \\

\bottomrule
\end{tabular}
}
\vspace{-2pt}
\caption{
\textbf{Evaluation on \textbf{SpatialViz}  (\name Protocol)}.
All scores are \textbf{\textit{CAA}} (computed via \cref{eq:caa}).
\textit{Prompt}: we use the unified chain-of-thought prompt during evaluation defined in \cref{fig:system_prompt}. $^{\dagger}$ indicates cases where some responses were truncated without a final answer, these are counted as incorrect and reduce the score. 
}
\label{tab:spatialviz_ss}
\end{table*}

%% file: tables/token.tex
\begin{table}[t]
\centering\small
\setlength{\tabcolsep}{6pt}
\renewcommand{\arraystretch}{1.15}

\begin{tabular}{lccccccc}
\toprule
\multirow{2}{*}{\textbf{Models}} & \multicolumn{3}{c}{\textbf{Internal Tokens}} & \multicolumn{3}{c}{\textbf{External Tokens}} & \multirow{2}{*}{\textbf{Exposure Ratio}}\\
\cmidrule(lr){2-4} \cmidrule(lr){5-7}
& \textbf{Mean} & \textbf{Median} & \boldmath$p_{95}$ & \textbf{Mean} & \textbf{Median} & \boldmath$p_{95}$ \\
\midrule

\rowcolor{line-blue}
\multicolumn{8}{l}{\textbf{Proprietary Models}} \\
\rowcolor{white}
Seed-1.6-2025-06-15~\cite{seed2025seed1_5vl} & 
    1,321 &   948 & 3,730 & 327 & 311 & 618 & 24.7\%\\
Gemini-2.5-pro-2025-06~\cite{team2023gemini} & 
    3,013 & 1,314 & 12,167 & 622 & 538 & 1,272 & 20.6\% \\
Grok-4-2025-07-09~\cite{grok4_xai_2025} & 
    7,666 & 7,158 & 16,384 & 173 & 173 & 291 & 2.3\%\\
GPT-5-2025-08-07~\cite{openai_gpt5_systemcard} & 
    1,359 & 1,024 & 4,160 & 104 & 95  & 198 & 7.6\%\\
\midrule

\rowcolor{line-blue}
\multicolumn{8}{l}{\textbf{Open-source Models}}\\
\rowcolor{white}
Qwen2.5-VL-72B-Instruct~\cite{Qwen-VL} & 
    - & - & - & 205 & 187 & 361 & - \\
InternVL3-78B~\cite{zhu2025internvl3} & 
    - & - & - & 207 & 147 & 416 & - \\
\bottomrule
\end{tabular}
\caption{Token consumption key statistics comparison. \boldmath$p_{95}$ represents the 95th percentile. Open-source models do not have an internalized reasoning process and the exposure ratio does not apply.}
\label{tab:token}
\end{table}






%% file: tables/thinking_mode.tex
\begin{table}[h]
    \centering\small
    \begin{tabular}{lcccc}
    \toprule
        \textbf{Thinking Mode} & 
        \textbf{Accuracy} &
        \makecell{\textbf{Reasoning tokens}\\(Average)} &
        \makecell{\textbf{Reasoning tokens}\\(Max)} &
        \makecell{\textbf{Runtime(s)}\\(Average)} \\
    \midrule
        Minimal & 48.31 & 0 & 0 & 11.69 \\
        Low & 54.24 & 1,899 & 6,636 & 53.89 \\
        Medium & 56.78 & 5,860 & 13,760 & 140.3 \\
        High & 52.54 & 8,567 & 16,064 & 305.2 \\
    \bottomrule
    \end{tabular}
    \caption{
    Ablation on thinking mode of GPT-5~\cite{openai_gpt5_systemcard} on the SpatialViz-Tiny set 
    (sampled at one-tenth per task from full SpatialViz set), 
    with \texttt{max\_completion\_tokens}=16{,}384. 
    In \textit{High} mode, 28 questions exceeded the 15-minute time limit or hit token limit, and were counted as incorrect, 
    resulting in an accuracy of 52.54\%; excluding these cases yields 68.89\%. 
    }
    \label{tab:thinking_mode}

\end{table}

%% file: main.bbl
\begin{thebibliography}{81}
\providecommand{\natexlab}[1]{#1}
\providecommand{\url}[1]{\texttt{#1}}
\expandafter\ifx\csname urlstyle\endcsname\relax
  \providecommand{\doi}[1]{doi: #1}\else
  \providecommand{\doi}{doi: \begingroup \urlstyle{rm}\Url}\fi

\bibitem[Bai et~al.(2023)Bai, Bai, Yang, Wang, Tan, Wang, Lin, Zhou, and Zhou]{Qwen-VL}
Jinze Bai, Shuai Bai, Shusheng Yang, Shijie Wang, Sinan Tan, Peng Wang, Junyang Lin, Chang Zhou, and Jingren Zhou.
\newblock Qwen-vl: A versatile vision-language model for understanding, localization, text reading, and beyond.
\newblock \emph{arXiv preprint arXiv:2308.12966}, 2023.

\bibitem[Cai et~al.(2024)Cai, Ponomarenko, Yuan, Li, Yang, Dong, and Zhao]{cai2024spatialbot}
Wenxiao Cai, Iaroslav Ponomarenko, Jianhao Yuan, Xiaoqi Li, Wankou Yang, Hao Dong, and Bo~Zhao.
\newblock Spatialbot: Precise spatial understanding with vision language models.
\newblock \emph{arXiv preprint arXiv:2406.13642}, 2024.

\bibitem[Chen et~al.(2024{\natexlab{a}})Chen, Xu, Kirmani, Ichter, Sadigh, Guibas, and Xia]{chen2024spatialvlm}
Boyuan Chen, Zhuo Xu, Sean Kirmani, Brain Ichter, Dorsa Sadigh, Leonidas Guibas, and Fei Xia.
\newblock Spatialvlm: Endowing vision-language models with spatial reasoning capabilities.
\newblock In \emph{Proceedings of the IEEE/CVF Conference on Computer Vision and Pattern Recognition}, pages 14455--14465, 2024{\natexlab{a}}.

\bibitem[Chen et~al.(2024{\natexlab{b}})Chen, Liang, Siu, Wang, Wang, Wang, Ni, Zhu, Jiang, Lyu, et~al.]{chen2024mega}
Jiacheng Chen, Tianhao Liang, Sherman Siu, Zhengqing Wang, Kai Wang, Yubo Wang, Yuansheng Ni, Wang Zhu, Ziyan Jiang, Bohan Lyu, et~al.
\newblock Mega-bench: Scaling multimodal evaluation to over 500 real-world tasks.
\newblock \emph{arXiv preprint arXiv:2410.10563}, 2024{\natexlab{b}}.

\bibitem[Chen et~al.(2024{\natexlab{c}})Chen, Li, Dong, Zhang, Zang, Chen, Duan, Wang, Qiao, Lin, et~al.]{chen2024we}
Lin Chen, Jinsong Li, Xiaoyi Dong, Pan Zhang, Yuhang Zang, Zehui Chen, Haodong Duan, Jiaqi Wang, Yu~Qiao, Dahua Lin, et~al.
\newblock Are we on the right way for evaluating large vision-language models?
\newblock \emph{Advances in Neural Information Processing Systems}, 37:\penalty0 27056--27087, 2024{\natexlab{c}}.

\bibitem[Chen et~al.(2024{\natexlab{d}})Chen, Wu, Wang, Su, Chen, Xing, Zhong, Zhang, Zhu, Lu, et~al.]{chen2024internvl}
Zhe Chen, Jiannan Wu, Wenhai Wang, Weijie Su, Guo Chen, Sen Xing, Muyan Zhong, Qinglong Zhang, Xizhou Zhu, Lewei Lu, et~al.
\newblock Internvl: Scaling up vision foundation models and aligning for generic visual-linguistic tasks.
\newblock In \emph{Proceedings of the IEEE/CVF Conference on Computer Vision and Pattern Recognition,}, pages 24185--24198, 2024{\natexlab{d}}.

\bibitem[Cheng et~al.(2024)Cheng, Yin, Fu, Guo, Yang, Kautz, Wang, and Liu]{cheng2024spatialrgpt}
An-Chieh Cheng, Hongxu Yin, Yang Fu, Qiushan Guo, Ruihan Yang, Jan Kautz, Xiaolong Wang, and Sifei Liu.
\newblock Spatialrgpt: Grounded spatial reasoning in vision-language models.
\newblock \emph{Advances in Neural Information Processing Systems}, 37:\penalty0 135062--135093, 2024.

\bibitem[Cheng et~al.(2025)Cheng, Ge, Wang, Ge, Liao, and Shan]{cheng2025video}
Junhao Cheng, Yuying Ge, Teng Wang, Yixiao Ge, Jing Liao, and Ying Shan.
\newblock Video-holmes: Can mllm think like holmes for complex video reasoning?
\newblock \emph{arXiv preprint arXiv:2505.21374}, 2025.

\bibitem[Deng et~al.(2025{\natexlab{a}})Deng, Zhu, Li, Gou, Li, Wang, Zhong, Yu, Nie, Song, Shi, and Fan]{deng2025bagel}
Chaorui Deng, Deyao Zhu, Kunchang Li, Chenhui Gou, Feng Li, Zeyu Wang, Shu Zhong, Weihao Yu, Xiaonan Nie, Ziang Song, Guang Shi, and Haoqi Fan.
\newblock Emerging properties in unified multimodal pretraining.
\newblock \emph{arXiv preprint arXiv:2505.14683}, 2025{\natexlab{a}}.

\bibitem[Deng et~al.(2025{\natexlab{b}})Deng, Gu, Ye, He, Chen, Li, Wang, Wei, Yang, Dou, et~al.]{deng2025internspatial}
Nianchen Deng, Lixin Gu, Shenglong Ye, Yinan He, Zhe Chen, Songze Li, Haomin Wang, Xingguang Wei, Tianshuo Yang, Min Dou, et~al.
\newblock Internspatial: A comprehensive dataset for spatial reasoning in vision-language models.
\newblock \emph{arXiv preprint arXiv:2506.18385}, 2025{\natexlab{b}}.

\bibitem[Du et~al.(2024)Du, Wu, Li, Huang, and Wei]{du2024embspatial}
Mengfei Du, Binhao Wu, Zejun Li, Xuanjing Huang, and Zhongyu Wei.
\newblock Embspatial-bench: Benchmarking spatial understanding for embodied tasks with large vision-language models.
\newblock \emph{arXiv preprint arXiv:2406.05756}, 2024.

\bibitem[Duan et~al.(2024)Duan, Yang, Qiao, Fang, Chen, Liu, Dong, Zang, Zhang, Wang, et~al.]{duan2024vlmevalkit}
Haodong Duan, Junming Yang, Yuxuan Qiao, Xinyu Fang, Lin Chen, Yuan Liu, Xiaoyi Dong, Yuhang Zang, Pan Zhang, Jiaqi Wang, et~al.
\newblock Vlmevalkit: An open-source toolkit for evaluating large multi-modality models.
\newblock In \emph{Proceedings of the 32nd ACM International Conference on Multimedia}, pages 11198--11201, 2024.

\bibitem[Feng(2025)]{feng2025towards}
Qi~Feng.
\newblock Towards visuospatial cognition via hierarchical fusion of visual experts.
\newblock \emph{arXiv preprint arXiv:2505.12363}, 2025.

\bibitem[Fu et~al.(2024{\natexlab{a}})Fu, Kuang, Song, Huang, Yang, Li, Zhu, Luo, Wang, Lu, et~al.]{fu2024ocrbench}
Ling Fu, Zhebin Kuang, Jiajun Song, Mingxin Huang, Biao Yang, Yuzhe Li, Linghao Zhu, Qidi Luo, Xinyu Wang, Hao Lu, et~al.
\newblock Ocrbench v2: An improved benchmark for evaluating large multimodal models on visual text localization and reasoning.
\newblock \emph{arXiv preprint arXiv:2501.00321}, 2024{\natexlab{a}}.

\bibitem[Fu et~al.(2024{\natexlab{b}})Fu, Hu, Li, Feng, Wang, Lin, Roth, Smith, Ma, and Krishna]{fu2024blink}
Xingyu Fu, Yushi Hu, Bangzheng Li, Yu~Feng, Haoyu Wang, Xudong Lin, Dan Roth, Noah~A Smith, Wei-Chiu Ma, and Ranjay Krishna.
\newblock Blink: Multimodal large language models can see but not perceive.
\newblock In \emph{Proceedings of the European Conference on Computer Vision}, pages 148--166. Springer, 2024{\natexlab{b}}.

\bibitem[Gong et~al.(2025)Gong, Li, Ma, Li, Ji, Yang, Luo, Yan, and Ji]{gong2025space}
Ziyang Gong, Wenhao Li, Oliver Ma, Songyuan Li, Jiayi Ji, Xue Yang, Gen Luo, Junchi Yan, and Rongrong Ji.
\newblock Space-10: A comprehensive benchmark for multimodal large language models in compositional spatial intelligence.
\newblock \emph{arXiv preprint arXiv:2506.07966}, 2025.

\bibitem[Guan et~al.(2024)Guan, Liu, Wu, Xian, Li, Liu, Wang, Chen, Huang, Yacoob, et~al.]{guan2024hallusionbench}
Tianrui Guan, Fuxiao Liu, Xiyang Wu, Ruiqi Xian, Zongxia Li, Xiaoyu Liu, Xijun Wang, Lichang Chen, Furong Huang, Yaser Yacoob, et~al.
\newblock Hallusionbench: an advanced diagnostic suite for entangled language hallucination and visual illusion in large vision-language models.
\newblock In \emph{Proceedings of the IEEE/CVF Conference on Computer Vision and Pattern Recognition}, pages 14375--14385, 2024.

\bibitem[He et~al.(2025)He, Huang, Chen, Pei, Xu, Lu, and Pang]{he2025egoexobench}
Yuping He, Yifei Huang, Guo Chen, Baoqi Pei, Jilan Xu, Tong Lu, and Jiangmiao Pang.
\newblock Egoexobench: A benchmark for first-and third-person view video understanding in mllms.
\newblock \emph{arXiv preprint arXiv:2507.18342}, 2025.

\bibitem[Hou et~al.(2025)Hou, Zhao, Xu, Fan, and Jiang]{hou2025tdbench}
Kaiyuan Hou, Minghui Zhao, Lilin Xu, Yuang Fan, and Xiaofan Jiang.
\newblock Tdbench: Benchmarking vision-language models in understanding top-down images.
\newblock \emph{arXiv preprint arXiv:2504.03748}, 2025.

\bibitem[Huang et~al.(2025)Huang, Wu, Xie, and Han]{huang2025mllms}
Xiaohu Huang, Jingjing Wu, Qunyi Xie, and Kai Han.
\newblock Mllms need 3d-aware representation supervision for scene understanding.
\newblock \emph{arXiv preprint arXiv:2506.01946}, 2025.

\bibitem[Hurst et~al.(2024)Hurst, Lerer, Goucher, Perelman, Ramesh, Clark, Ostrow, Welihinda, Hayes, Radford, et~al.]{hurst2024gpt}
Aaron Hurst, Adam Lerer, Adam~P Goucher, Adam Perelman, Aditya Ramesh, Aidan Clark, AJ~Ostrow, Akila Welihinda, Alan Hayes, Alec Radford, et~al.
\newblock Gpt-4o system card.
\newblock \emph{arXiv preprint arXiv:2410.21276}, 2024.

\bibitem[Ji et~al.(2025)Ji, Wang, Liu, Hao, Liu, Zhao, Lyu, and Zheng]{ji2025visualtrans}
Yuheng Ji, Yipu Wang, Yuyang Liu, Xiaoshuai Hao, Yue Liu, Yuting Zhao, Huaihai Lyu, and Xiaolong Zheng.
\newblock Visualtrans: A benchmark for real-world visual transformation reasoning.
\newblock \emph{arXiv preprint arXiv:2508.04043}, 2025.

\bibitem[Jia et~al.(2025)Jia, Qi, Zhang, Zhang, Yu, He, Wang, and Yi]{jia2025omnispatial}
Mengdi Jia, Zekun Qi, Shaochen Zhang, Wenyao Zhang, Xinqiang Yu, Jiawei He, He~Wang, and Li~Yi.
\newblock Omnispatial: Towards comprehensive spatial reasoning benchmark for vision language models.
\newblock \emph{arXiv preprint arXiv:2506.03135}, 2025.

\bibitem[Kim et~al.(2024)Kim, Pertsch, Karamcheti, Xiao, Balakrishna, Nair, Rafailov, Foster, Lam, Sanketi, et~al.]{kim2024openvla}
Moo~Jin Kim, Karl Pertsch, Siddharth Karamcheti, Ted Xiao, Ashwin Balakrishna, Suraj Nair, Rafael Rafailov, Ethan Foster, Grace Lam, Pannag Sanketi, et~al.
\newblock Openvla: An open-source vision-language-action model.
\newblock \emph{arXiv preprint arXiv:2406.09246}, 2024.

\bibitem[Kong et~al.(2025)Kong, Duan, Xu, Guo, Zhu, and Shi]{kong2025lrr}
Fei Kong, Jinhao Duan, Kaidi Xu, Zhenhua Guo, Xiaofeng Zhu, and Xiaoshuang Shi.
\newblock Lrr-bench: Left, right or rotate? vision-language models still struggle with spatial understanding tasks.
\newblock \emph{arXiv preprint arXiv:2507.20174}, 2025.

\bibitem[Li et~al.(2025{\natexlab{a}})Li, Wang, Yue, Cai, Liu, Fu, Guo, Zhu, Sharan, Jia, et~al.]{li2025zebra}
Ang Li, Charles Wang, Kaiyu Yue, Zikui Cai, Ollie Liu, Deqing Fu, Peng Guo, Wang~Bill Zhu, Vatsal Sharan, Robin Jia, et~al.
\newblock Zebra-cot: A dataset for interleaved vision language reasoning.
\newblock \emph{arXiv preprint arXiv:2507.16746}, 2025{\natexlab{a}}.

\bibitem[Li et~al.(2024{\natexlab{a}})Li, Zhang, Zhang, Pu, Du, Dong, Liu, Zhang, Zhang, Li, and Liu]{lmms_eval2024}
Bo~Li, Peiyuan Zhang, Kaichen Zhang, Fanyi Pu, Xinrun Du, Yuhao Dong, Haotian Liu, Yuanhan Zhang, Ge~Zhang, Chunyuan Li, and Ziwei Liu.
\newblock Lmms-eval: Accelerating the development of large multimoal models, March 2024{\natexlab{a}}.
\newblock URL \url{https://github.com/EvolvingLMMs-Lab/lmms-eval}.

\bibitem[Li et~al.(2024{\natexlab{b}})Li, Zhang, Guo, Zhang, Li, Zhang, Zhang, Zhang, Li, Liu, et~al.]{li2024llava}
Bo~Li, Yuanhan Zhang, Dong Guo, Renrui Zhang, Feng Li, Hao Zhang, Kaichen Zhang, Peiyuan Zhang, Yanwei Li, Ziwei Liu, et~al.
\newblock Llava-onevision: Easy visual task transfer.
\newblock \emph{arXiv preprint arXiv:2408.03326}, 2024{\natexlab{b}}.

\bibitem[Li et~al.(2025{\natexlab{b}})Li, Zhang, Chen, Wang, Pu, Cahyono, Yang, Li, and Liu]{li2025otter}
Bo~Li, Yuanhan Zhang, Liangyu Chen, Jinghao Wang, Fanyi Pu, Joshua~Adrian Cahyono, Jingkang Yang, Chunyuan Li, and Ziwei Liu.
\newblock Otter: A multi-modal model with in-context instruction tuning.
\newblock \emph{IEEE Transactions on Pattern Analysis and Machine Intelligence}, 2025{\natexlab{b}}.

\bibitem[Li et~al.(2024{\natexlab{c}})Li, Ge, Ge, Wang, Wang, Zhang, and Shan]{li2024seed}
Bohao Li, Yuying Ge, Yixiao Ge, Guangzhi Wang, Rui Wang, Ruimao Zhang, and Ying Shan.
\newblock Seed-bench: Benchmarking multimodal large language models.
\newblock In \emph{Proceedings of the IEEE/CVF Conference on Computer Vision and Pattern Recognition}, pages 13299--13308, 2024{\natexlab{c}}.

\bibitem[Li et~al.(2024{\natexlab{d}})Li, Nan, Lu, Du, and Zhang]{li2024proximity}
Jianing Li, Xi~Nan, Ming Lu, Li~Du, and Shanghang Zhang.
\newblock Proximity qa: Unleashing the power of multi-modal large language models for spatial proximity analysis.
\newblock \emph{arXiv preprint arXiv:2401.17862}, 2024{\natexlab{d}}.

\bibitem[Li et~al.(2025{\natexlab{c}})Li, Bigverdi, Gu, Ma, Yang, Li, Choi, and Krishna]{li2025unfolding}
Linjie Li, Mahtab Bigverdi, Jiawei Gu, Zixian Ma, Yinuo Yang, Ziang Li, Yejin Choi, and Ranjay Krishna.
\newblock Unfolding spatial cognition: Evaluating multimodal models on visual simulations.
\newblock \emph{arXiv preprint arXiv:2506.04633}, 2025{\natexlab{c}}.

\bibitem[Li et~al.(2024{\natexlab{e}})Li, Gao, Zhao, Wang, Sun, Lyu, Hawkins, Vasconcelos, Golan, Luo, et~al.]{li2024core}
Yijiang Li, Qingying Gao, Tianwei Zhao, Bingyang Wang, Haoran Sun, Haiyun Lyu, Robert~D Hawkins, Nuno Vasconcelos, Tal Golan, Dezhi Luo, et~al.
\newblock Core knowledge deficits in multi-modal language models.
\newblock \emph{arXiv preprint arXiv:2410.10855}, 2024{\natexlab{e}}.

\bibitem[Li et~al.(2024{\natexlab{f}})Li, Yang, Choi, Zhu, Hsieh, Kim, Lim, Ji, Lee, Yan, et~al.]{li2024mmsci}
Zekun Li, Xianjun Yang, Kyuri Choi, Wanrong Zhu, Ryan Hsieh, HyeonJung Kim, Jin~Hyuk Lim, Sungyoung Ji, Byungju Lee, Xifeng Yan, et~al.
\newblock Mmsci: A multimodal multi-discipline dataset for phd-level scientific comprehension.
\newblock In \emph{AI for Accelerated Materials Design-Vienna 2024}, 2024{\natexlab{f}}.

\bibitem[Liu et~al.(2023)Liu, Li, Wu, and Lee]{liu2023llava}
Haotian Liu, Chunyuan Li, Qingyang Wu, and Yong~Jae Lee.
\newblock Visual instruction tuning, 2023.

\bibitem[Liu et~al.(2024{\natexlab{a}})Liu, Li, Li, Li, Zhang, Shen, and Lee]{liu2024llavanext}
Haotian Liu, Chunyuan Li, Yuheng Li, Bo~Li, Yuanhan Zhang, Sheng Shen, and Yong~Jae Lee.
\newblock Llava-next: Improved reasoning, ocr, and world knowledge, January 2024{\natexlab{a}}.
\newblock URL \url{https://llava-vl.github.io/blog/2024-01-30-llava-next/}.

\bibitem[Liu et~al.(2024{\natexlab{b}})Liu, Duan, Zhang, Li, Zhang, Zhao, Yuan, Wang, He, Liu, et~al.]{liu2024mmbench}
Yuan Liu, Haodong Duan, Yuanhan Zhang, Bo~Li, Songyang Zhang, Wangbo Zhao, Yike Yuan, Jiaqi Wang, Conghui He, Ziwei Liu, et~al.
\newblock Mmbench: Is your multi-modal model an all-around player?
\newblock In \emph{Proceedings of the European Conference on Computer Vision}, pages 216--233. Springer, 2024{\natexlab{b}}.

\bibitem[Lu et~al.(2022)Lu, Mishra, Xia, Qiu, Chang, Zhu, Tafjord, Clark, and Kalyan]{lu2022learn}
Pan Lu, Swaroop Mishra, Tanglin Xia, Liang Qiu, Kai-Wei Chang, Song-Chun Zhu, Oyvind Tafjord, Peter Clark, and Ashwin Kalyan.
\newblock Learn to explain: Multimodal reasoning via thought chains for science question answering.
\newblock \emph{Advances in Neural Information Processing Systems}, 35:\penalty0 2507--2521, 2022.

\bibitem[Lu et~al.(2023)Lu, Bansal, Xia, Liu, Li, Hajishirzi, Cheng, Chang, Galley, and Gao]{lu2023mathvista}
Pan Lu, Hritik Bansal, Tony Xia, Jiacheng Liu, Chunyuan Li, Hannaneh Hajishirzi, Hao Cheng, Kai-Wei Chang, Michel Galley, and Jianfeng Gao.
\newblock Mathvista: Evaluating mathematical reasoning of foundation models in visual contexts.
\newblock \emph{arXiv preprint arXiv:2310.02255}, 2023.

\bibitem[Ma et~al.(2024{\natexlab{a}})Ma, Chen, Zhang, Chou, de~Melo, and Yuille]{ma20243dsrbench}
Wufei Ma, Haoyu Chen, Guofeng Zhang, Yu-Cheng Chou, Celso~M de~Melo, and Alan Yuille.
\newblock 3dsrbench: A comprehensive 3d spatial reasoning benchmark.
\newblock \emph{arXiv preprint arXiv:2412.07825}, 2024{\natexlab{a}}.

\bibitem[Ma et~al.(2024{\natexlab{b}})Ma, Zang, Chen, Chen, Jiao, Li, Lu, Liu, Ma, Dong, et~al.]{ma2024mmlongbench}
Yubo Ma, Yuhang Zang, Liangyu Chen, Meiqi Chen, Yizhu Jiao, Xinze Li, Xinyuan Lu, Ziyu Liu, Yan Ma, Xiaoyi Dong, et~al.
\newblock Mmlongbench-doc: Benchmarking long-context document understanding with visualizations.
\newblock \emph{Advances in Neural Information Processing Systems}, 37:\penalty0 95963--96010, 2024{\natexlab{b}}.

\bibitem[Masry et~al.(2022)Masry, Long, Tan, Joty, and Hoque]{masry2022chartqa}
Ahmed Masry, Do~Xuan Long, Jia~Qing Tan, Shafiq Joty, and Enamul Hoque.
\newblock Chartqa: A benchmark for question answering about charts with visual and logical reasoning.
\newblock \emph{arXiv preprint arXiv:2203.10244}, 2022.

\bibitem[Mishra et~al.(2019)Mishra, Shekhar, Singh, and Chakraborty]{mishra2019ocr}
Anand Mishra, Shashank Shekhar, Ajeet~Kumar Singh, and Anirban Chakraborty.
\newblock Ocr-vqa: Visual question answering by reading text in images.
\newblock In \emph{Proceedings of the International Conference on Document Analysis and Recognition}, pages 947--952. IEEE, 2019.

\bibitem[OpenAI(2025)]{gpt_o3}
OpenAI.
\newblock Openai o3 and o4-mini system card, 2025.
\newblock URL \url{https://openai.com/research/o3-o4-mini-system-card}.

\bibitem[{OpenAI}(2025)]{openai_gpt5_systemcard}
{OpenAI}.
\newblock {GPT-5 System Card}.
\newblock Technical report, OpenAI, August 2025.
\newblock Accessed: 2025-08-10.

\bibitem[Rizvi et~al.(2025)Rizvi, Zhu, and Gurevych]{rizvi2025spare}
Md~Imbesat~Hassan Rizvi, Xiaodan Zhu, and Iryna Gurevych.
\newblock Spare: Single-pass annotation with reference-guided evaluation for automatic process supervision and reward modelling.
\newblock \emph{arXiv preprint arXiv:2506.15498}, 2025.

\bibitem[Song et~al.(2025)Song, Lin, Huang, Wang, and Lin]{song2025siri}
Zijian Song, Xiaoxin Lin, Qiuming Huang, Guangrun Wang, and Liang Lin.
\newblock Siri-bench: Challenging vlms' spatial intelligence through complex reasoning tasks.
\newblock \emph{arXiv preprint arXiv:2506.14512}, 2025.

\bibitem[Sun et~al.(2024{\natexlab{a}})Sun, Zhou, Li, Lu, Yi, Chen, Xu, Luo, Zhang, Zhan, et~al.]{sun2024parrot}
Hai-Long Sun, Da-Wei Zhou, Yang Li, Shiyin Lu, Chao Yi, Qing-Guo Chen, Zhao Xu, Weihua Luo, Kaifu Zhang, De-Chuan Zhan, et~al.
\newblock Parrot: Multilingual visual instruction tuning.
\newblock \emph{arXiv preprint arXiv:2406.02539}, 2024{\natexlab{a}}.

\bibitem[Sun et~al.(2024{\natexlab{b}})Sun, Wu, Zhu, Zheng, Chen, Zhang, Zhang, Wan, Lan, Zheng, et~al.]{sun2024pathmmu}
Yuxuan Sun, Hao Wu, Chenglu Zhu, Sunyi Zheng, Qizi Chen, Kai Zhang, Yunlong Zhang, Dan Wan, Xiaoxiao Lan, Mengyue Zheng, et~al.
\newblock Pathmmu: A massive multimodal expert-level benchmark for understanding and reasoning in pathology.
\newblock In \emph{Proceedings of the European Conference on Computer Vision}, pages 56--73. Springer, 2024{\natexlab{b}}.

\bibitem[Szyma{\'n}ska et~al.(2024)Szyma{\'n}ska, Dusmanu, Buurlage, Rad, and Pollefeys]{szymanska2024space3d}
Emilia Szyma{\'n}ska, Mihai Dusmanu, Jan-Willem Buurlage, Mahdi Rad, and Marc Pollefeys.
\newblock Space3d-bench: Spatial 3d question answering benchmark.
\newblock In \emph{Proceedings of the European Conference on Computer Vision}, pages 68--85. Springer, 2024.

\bibitem[Team(2025)]{seed2025seed1_5vl}
ByteDance~Seed Team.
\newblock Seed1.5-vl technical report.
\newblock \emph{arXiv preprint arXiv:2505.07062}, 2025.

\bibitem[Team et~al.(2023)Team, Anil, Borgeaud, Alayrac, Yu, Soricut, Schalkwyk, Dai, Hauth, Millican, et~al.]{team2023gemini}
Gemini Team, Rohan Anil, Sebastian Borgeaud, Jean-Baptiste Alayrac, Jiahui Yu, Radu Soricut, Johan Schalkwyk, Andrew~M Dai, Anja Hauth, Katie Millican, et~al.
\newblock Gemini: a family of highly capable multimodal models.
\newblock \emph{arXiv preprint arXiv:2312.11805}, 2023.

\bibitem[Tong et~al.(2024)Tong, Brown, Wu, Woo, IYER, Akula, Yang, Yang, Middepogu, Wang, et~al.]{tong2024cambrian}
Peter Tong, Ellis Brown, Penghao Wu, Sanghyun Woo, Adithya Jairam~Vedagiri IYER, Sai~Charitha Akula, Shusheng Yang, Jihan Yang, Manoj Middepogu, Ziteng Wang, et~al.
\newblock Cambrian-1: A fully open, vision-centric exploration of multimodal llms.
\newblock \emph{Advances in Neural Information Processing Systems}, 37:\penalty0 87310--87356, 2024.

\bibitem[Wang et~al.(2025{\natexlab{a}})Wang, Zhao, Wang, Fan, Zhang, and Zhang]{wang2025ross3d}
Haochen Wang, Yucheng Zhao, Tiancai Wang, Haoqiang Fan, Xiangyu Zhang, and Zhaoxiang Zhang.
\newblock Ross3d: Reconstructive visual instruction tuning with 3d-awareness.
\newblock \emph{arXiv preprint arXiv:2504.01901}, 2025{\natexlab{a}}.

\bibitem[Wang et~al.(2025{\natexlab{b}})Wang, Sun, Deng, Shao, Pei, Tian, Zhang, and Wang]{wang2025spatialviz}
Siting Wang, Luoyang Sun, Cheng Deng, Kun Shao, Minnan Pei, Zheng Tian, Haifeng Zhang, and Jun Wang.
\newblock Spatialviz-bench: Automatically generated spatial visualization reasoning tasks for mllms.
\newblock \emph{arXiv preprint arXiv:2507.07610}, 2025{\natexlab{b}}.

\bibitem[Wang et~al.(2025{\natexlab{c}})Wang, Gao, Gu, Pu, Cui, Wei, Liu, Jing, Ye, Shao, et~al.]{wang2025internvl3}
Weiyun Wang, Zhangwei Gao, Lixin Gu, Hengjun Pu, Long Cui, Xingguang Wei, Zhaoyang Liu, Linglin Jing, Shenglong Ye, Jie Shao, et~al.
\newblock Internvl3. 5: Advancing open-source multimodal models in versatility, reasoning, and efficiency.
\newblock \emph{arXiv preprint arXiv:2508.18265}, 2025{\natexlab{c}}.

\bibitem[Wang et~al.(2025{\natexlab{d}})Wang, Tan, Zhu, Yang, Yang, Wang, Kolobov, Gao, and Gong]{wang2025site}
Wenqi Wang, Reuben Tan, Pengyue Zhu, Jianwei Yang, Zhengyuan Yang, Lijuan Wang, Andrey Kolobov, Jianfeng Gao, and Boqing Gong.
\newblock Site: towards spatial intelligence thorough evaluation.
\newblock \emph{arXiv preprint arXiv:2505.05456}, 2025{\natexlab{d}}.

\bibitem[Wang et~al.(2025{\natexlab{e}})Wang, Ma, Zhang, de~Melo, Chen, and Yuille]{wang2025spatial457}
Xingrui Wang, Wufei Ma, Tiezheng Zhang, Celso~M de~Melo, Jieneng Chen, and Alan Yuille.
\newblock Spatial457: A diagnostic benchmark for 6d spatial reasoning of large mutimodal models.
\newblock In \emph{Proceedings of the IEEE/CVF Conference on Computer Vision and Pattern Recognition,}, pages 24669--24679, 2025{\natexlab{e}}.

\bibitem[Wei et~al.(2022)Wei, Wang, Schuurmans, Bosma, Xia, Chi, Le, Zhou, et~al.]{wei2022chain}
Jason Wei, Xuezhi Wang, Dale Schuurmans, Maarten Bosma, Fei Xia, Ed~Chi, Quoc~V Le, Denny Zhou, et~al.
\newblock Chain-of-thought prompting elicits reasoning in large language models.
\newblock \emph{Advances in neural information processing systems}, 35:\penalty0 24824--24837, 2022.

\bibitem[Wu et~al.(2025{\natexlab{a}})Wu, Huang, Chen, Zhang, Wang, and Xie]{wu2025spatialscore}
Haoning Wu, Xiao Huang, Yaohui Chen, Ya~Zhang, Yanfeng Wang, and Weidi Xie.
\newblock Spatialscore: Towards unified evaluation for multimodal spatial understanding.
\newblock \emph{arXiv preprint arXiv:2505.17012}, 2025{\natexlab{a}}.

\bibitem[Wu et~al.(2025{\natexlab{b}})Wu, Zhang, Diao, Li, Lu, and Liu]{wu2025visual}
Penghao Wu, Yushan Zhang, Haiwen Diao, Bo~Li, Lewei Lu, and Ziwei Liu.
\newblock Visual jigsaw post-training improves mllms.
\newblock \emph{arXiv preprint arXiv:2509.25190}, 2025{\natexlab{b}}.

\bibitem[{xAI}(2025)]{grok4_xai_2025}
{xAI}.
\newblock Grok 4, 7 2025.
\newblock URL \url{https://x.ai/news/grok-4}.
\newblock Model announcement.

\bibitem[Xu et~al.(2025)Xu, Wang, Tang, Chen, Wang, Chu, Lin, Feiszli, and Liang]{xu2025multi}
Runsen Xu, Weiyao Wang, Hao Tang, Xingyu Chen, Xiaodong Wang, Fu-Jen Chu, Dahua Lin, Matt Feiszli, and Kevin~J Liang.
\newblock Multi-spatialmllm: Multi-frame spatial understanding with multi-modal large language models.
\newblock \emph{arXiv preprint arXiv:2505.17015}, 2025.

\bibitem[Yan et~al.(2025)Yan, Li, Chen, Luo, Wang, Zhang, and Shen]{yan2025mmcr}
Dawei Yan, Yang Li, Qing-Guo Chen, Weihua Luo, Peng Wang, Haokui Zhang, and Chunhua Shen.
\newblock Mmcr: Advancing visual language model in multimodal multi-turn contextual reasoning.
\newblock \emph{arXiv preprint arXiv:2503.18533}, 2025.

\bibitem[Yang et~al.(2025{\natexlab{a}})Yang, Li, Yang, Zhang, Hui, Zheng, Yu, Gao, Huang, Lv, Zheng, Liu, Zhou, Huang, Hu, Ge, Wei, Lin, Tang, Yang, Tu, Zhang, Yang, Yang, Zhou, Zhou, Lin, Dang, Bao, Yang, Yu, Deng, Li, Xue, Li, Zhang, Wang, Zhu, Men, Gao, Liu, Luo, Li, Tang, Yin, Ren, Wang, Zhang, Ren, Fan, Su, Zhang, Zhang, Wan, Liu, Wang, Cui, Zhang, Zhou, and Qiu]{yang2025qwen3technicalreport}
An~Yang, Anfeng Li, Baosong Yang, Beichen Zhang, Binyuan Hui, Bo~Zheng, Bowen Yu, Chang Gao, Chengen Huang, Chenxu Lv, Chujie Zheng, Dayiheng Liu, Fan Zhou, Fei Huang, Feng Hu, Hao Ge, Haoran Wei, Huan Lin, Jialong Tang, Jian Yang, Jianhong Tu, Jianwei Zhang, Jianxin Yang, Jiaxi Yang, Jing Zhou, Jingren Zhou, Junyang Lin, Kai Dang, Keqin Bao, Kexin Yang, Le~Yu, Lianghao Deng, Mei Li, Mingfeng Xue, Mingze Li, Pei Zhang, Peng Wang, Qin Zhu, Rui Men, Ruize Gao, Shixuan Liu, Shuang Luo, Tianhao Li, Tianyi Tang, Wenbiao Yin, Xingzhang Ren, Xinyu Wang, Xinyu Zhang, Xuancheng Ren, Yang Fan, Yang Su, Yichang Zhang, Yinger Zhang, Yu~Wan, Yuqiong Liu, Zekun Wang, Zeyu Cui, Zhenru Zhang, Zhipeng Zhou, and Zihan Qiu.
\newblock Qwen3 technical report, 2025{\natexlab{a}}.
\newblock URL \url{https://arxiv.org/abs/2505.09388}.

\bibitem[Yang et~al.(2025{\natexlab{b}})Yang, Yang, Gupta, Han, Fei-Fei, and Xie]{yang2025thinking}
Jihan Yang, Shusheng Yang, Anjali~W Gupta, Rilyn Han, Li~Fei-Fei, and Saining Xie.
\newblock Thinking in space: How multimodal large language models see, remember, and recall spaces.
\newblock In \emph{Proceedings of the IEEE/CVF Conference on Computer Vision and Pattern Recognition,}, pages 10632--10643, 2025{\natexlab{b}}.

\bibitem[Yang et~al.(2025{\natexlab{c}})Yang, Liu, Guo, Dong, Zhang, Zhang, Wang, Zhou, Xie, Wang, et~al.]{yang2025egolife}
Jingkang Yang, Shuai Liu, Hongming Guo, Yuhao Dong, Xiamengwei Zhang, Sicheng Zhang, Pengyun Wang, Zitang Zhou, Binzhu Xie, Ziyue Wang, et~al.
\newblock Egolife: Towards egocentric life assistant.
\newblock In \emph{Proceedings of the IEEE/CVF Conference on Computer Vision and Pattern Recognition,}, pages 28885--28900, 2025{\natexlab{c}}.

\bibitem[Yang et~al.(2025{\natexlab{d}})Yang, Xu, Xie, Yang, Li, Lin, Zhu, Chen, Duan, Yue, et~al.]{yang2025mmsi}
Sihan Yang, Runsen Xu, Yiman Xie, Sizhe Yang, Mo~Li, Jingli Lin, Chenming Zhu, Xiaochen Chen, Haodong Duan, Xiangyu Yue, et~al.
\newblock Mmsi-bench: A benchmark for multi-image spatial intelligence.
\newblock \emph{arXiv preprint arXiv:2505.23764}, 2025{\natexlab{d}}.

\bibitem[Yin et~al.(2025)Yin, Wang, Zhang, Zhang, Wang, Wang, Zhang, Chandrasegaran, Liu, Krishna, et~al.]{yin2025spatial}
Baiqiao Yin, Qineng Wang, Pingyue Zhang, Jianshu Zhang, Kangrui Wang, Zihan Wang, Jieyu Zhang, Keshigeyan Chandrasegaran, Han Liu, Ranjay Krishna, et~al.
\newblock Spatial mental modeling from limited views.
\newblock \emph{arXiv preprint arXiv:2506.21458}, 2025.

\bibitem[Yu et~al.(2025)Yu, Chen, Ju, Jia, Zhang, Huang, Wu, Cui, Ran, Zhang, et~al.]{yu2025far}
Songsong Yu, Yuxin Chen, Hao Ju, Lianjie Jia, Fuxi Zhang, Shaofei Huang, Yuhan Wu, Rundi Cui, Binghao Ran, Zaibin Zhang, et~al.
\newblock How far are vlms from visual spatial intelligence? a benchmark-driven perspective.
\newblock \emph{arXiv preprint arXiv:2509.18905}, 2025.

\bibitem[Yu et~al.(2023)Yu, Yang, Li, Wang, Lin, Liu, Wang, and Wang]{yu2023mm}
Weihao Yu, Zhengyuan Yang, Linjie Li, Jianfeng Wang, Kevin Lin, Zicheng Liu, Xinchao Wang, and Lijuan Wang.
\newblock Mm-vet: Evaluating large multimodal models for integrated capabilities.
\newblock \emph{arXiv preprint arXiv:2308.02490}, 2023.

\bibitem[Yue et~al.(2024)Yue, Ni, Zhang, Zheng, Liu, Zhang, Stevens, Jiang, Ren, Sun, et~al.]{yue2024mmmu}
Xiang Yue, Yuansheng Ni, Kai Zhang, Tianyu Zheng, Ruoqi Liu, Ge~Zhang, Samuel Stevens, Dongfu Jiang, Weiming Ren, Yuxuan Sun, et~al.
\newblock Mmmu: A massive multi-discipline multimodal understanding and reasoning benchmark for expert agi.
\newblock In \emph{Proceedings of the IEEE/CVF Conference on Computer Vision and Pattern Recognition}, pages 9556--9567, 2024.

\bibitem[Zhang et~al.(2024{\natexlab{a}})Zhang, Li, Zhang, Pu, Cahyono, Hu, Liu, Zhang, Yang, Li, and Liu]{zhang2024lmmsevalrealitycheckevaluation}
Kaichen Zhang, Bo~Li, Peiyuan Zhang, Fanyi Pu, Joshua~Adrian Cahyono, Kairui Hu, Shuai Liu, Yuanhan Zhang, Jingkang Yang, Chunyuan Li, and Ziwei Liu.
\newblock Lmms-eval: Reality check on the evaluation of large multimodal models, 2024{\natexlab{a}}.
\newblock URL \url{https://arxiv.org/abs/2407.12772}.

\bibitem[Zhang et~al.(2021)Zhang, Zheng, Wu, Fu, and Chen]{zhang2021mme}
Yunhang Shen Yulei Qin~Mengdan Zhang, Xu~Lin Jinrui Yang~Xiawu Zheng, Ke~Li Xing Sun~Yunsheng Wu, Rongrong Ji~Chaoyou Fu, and Peixian Chen.
\newblock Mme: A comprehensive evaluation benchmark for multimodal large language models.
\newblock \emph{arXiv preprint arXiv:2306.13394}, 2021.

\bibitem[Zhang et~al.(2024{\natexlab{b}})Zhang, Wu, Li, Zhou, Sun, Min, Chen, Liu, Lin, and Zhai]{zhang2024bench}
Zicheng Zhang, Haoning Wu, Chunyi Li, Yingjie Zhou, Wei Sun, Xiongkuo Min, Zijian Chen, Xiaohong Liu, Weisi Lin, and Guangtao Zhai.
\newblock A-bench: Are lmms masters at evaluating ai-generated images?
\newblock \emph{arXiv preprint arXiv:2406.03070}, 2024{\natexlab{b}}.

\bibitem[Zheng et~al.(2025)Zheng, Huang, Li, and Wang]{zheng2025learning}
Duo Zheng, Shijia Huang, Yanyang Li, and Liwei Wang.
\newblock Learning from videos for 3d world: Enhancing mllms with 3d vision geometry priors.
\newblock \emph{arXiv preprint arXiv:2505.24625}, 2025.

\bibitem[Zhou et~al.(2025{\natexlab{a}})Zhou, An, Chi, Han, Rong, Zhang, Wang, Wang, Huang, Sheng, et~al.]{zhou2025roborefer}
Enshen Zhou, Jingkun An, Cheng Chi, Yi~Han, Shanyu Rong, Chi Zhang, Pengwei Wang, Zhongyuan Wang, Tiejun Huang, Lu~Sheng, et~al.
\newblock Roborefer: Towards spatial referring with reasoning in vision-language models for robotics.
\newblock \emph{arXiv preprint arXiv:2506.04308}, 2025{\natexlab{a}}.

\bibitem[Zhou et~al.(2025{\natexlab{b}})Zhou, Vilesov, He, Wan, Zhang, Nagachandra, Chang, Chen, Wang, and Kadambi]{zhou2025vlm4d}
Shijie Zhou, Alexander Vilesov, Xuehai He, Ziyu Wan, Shuwang Zhang, Aditya Nagachandra, Di~Chang, Dongdong Chen, Xin~Eric Wang, and Achuta Kadambi.
\newblock Vlm4d: Towards spatiotemporal awareness in vision language models.
\newblock \emph{arXiv preprint arXiv:2508.02095}, 2025{\natexlab{b}}.

\bibitem[Zhu et~al.(2025{\natexlab{a}})Zhu, Wang, Chen, Liu, Ye, Gu, Tian, Duan, Su, Shao, et~al.]{zhu2025internvl3}
Jinguo Zhu, Weiyun Wang, Zhe Chen, Zhaoyang Liu, Shenglong Ye, Lixin Gu, Hao Tian, Yuchen Duan, Weijie Su, Jie Shao, et~al.
\newblock Internvl3: Exploring advanced training and test-time recipes for open-source multimodal models.
\newblock \emph{arXiv preprint arXiv:2504.10479}, 2025{\natexlab{a}}.

\bibitem[Zhu et~al.(2025{\natexlab{b}})Zhu, Jia, Wang, Zhao, Duan, Min, Wang, Zhang, and Zhai]{zhu2025gobench}
Xiaorong Zhu, Ziheng Jia, Jiarui Wang, Xiangyu Zhao, Haodong Duan, Xiongkuo Min, Jia Wang, Zicheng Zhang, and Guangtao Zhai.
\newblock Gobench: Benchmarking geometric optics generation and understanding of mllms.
\newblock \emph{arXiv preprint arXiv:2506.00991}, 2025{\natexlab{b}}.

\bibitem[Zitkovich et~al.(2023)Zitkovich, Yu, Xu, Xu, Xiao, Xia, Wu, Wohlhart, Welker, Wahid, et~al.]{zitkovich2023rt}
Brianna Zitkovich, Tianhe Yu, Sichun Xu, Peng Xu, Ted Xiao, Fei Xia, Jialin Wu, Paul Wohlhart, Stefan Welker, Ayzaan Wahid, et~al.
\newblock Rt-2: Vision-language-action models transfer web knowledge to robotic control.
\newblock In \emph{Conference on Robot Learning}, pages 2165--2183. PMLR, 2023.

\end{thebibliography}
